\documentclass[10pt,journal,compsoc]{IEEEtran}
\ifCLASSOPTIONcompsoc
  \usepackage[nocompress]{cite}
\else
  \usepackage{cite}
\fi

\usepackage{amsmath,amsthm}
\usepackage{amsfonts}
\usepackage{graphicx}
\usepackage{multirow}
\usepackage{epsfig}
\usepackage{setspace}
\usepackage{graphicx}
\usepackage{caption}
\usepackage{booktabs}
\usepackage{subfigure}
\usepackage{adjustbox}
\usepackage{algorithm}
\usepackage{algorithmicx}
\usepackage{algpseudocode}
\usepackage[dvipsnames]{xcolor}
\usepackage[shortlabels]{enumitem}
\theoremstyle{definition}
\newtheorem{definition}{Definition}
\usepackage{float}

\ifCLASSINFOpdf
\else
\fi
\usepackage{url}

\begin{document}
%

\title{Learning Single/Multi-Attribute of Object with Symmetry and Group}

%

\author{Yong-Lu~Li,~Yue~Xu,~Xinyu~Xu,~Xiaohan~Mao,~Cewu~Lu,~\IEEEmembership{Member,~IEEE}
\IEEEcompsocitemizethanks{
\IEEEcompsocthanksitem Yong-Lu Li, Yue Xu, Xinyu Xu, Xiaohan Mao are with the Department of Electrical and Computer Engineering, Shanghai Jiao Tong University, Shanghai, 200240, China.\protect\\
E-mail: \{yonglu\_li, silicxuyue, xuxinyu2000, mxh1999\}@sjtu.edu.cn.
\IEEEcompsocthanksitem Cewu Lu is the corresponding author, member of Qing Yuan Research Institute and MoE Key Lab of Artificial Intelligence, AI Institute, Shanghai Jiao Tong University, China. E-mail: lucewu@sjtu.edu.cn.
}}

\IEEEtitleabstractindextext{%
\begin{abstract}
Attributes and objects can compose diverse compositions. To model the compositional nature of these concepts, it is a good choice to learn them as transformations, e.g., coupling and decoupling. However, complex transformations need to satisfy specific principles to guarantee rationality. Here, we first propose a previously ignored principle of attribute-object transformation: \textbf{Symmetry}. For example, coupling \texttt{peeled-apple} with attribute \texttt{peeled} should result in \texttt{peeled-apple}, and decoupling \texttt{peeled} from \texttt{apple} should still output \texttt{apple}. Incorporating the symmetry, we propose a transformation framework inspired by group theory, i.e., SymNet. It consists of two modules: Coupling Network and Decoupling Network. We adopt deep neural networks to implement SymNet and train it in an end-to-end paradigm with the group axioms and symmetry as objectives. Then, we propose a Relative Moving Distance (RMD) based method to utilize the attribute change instead of the attribute pattern itself to classify attributes.
Besides the compositions of single-attribute and object, our RMD is also suitable for complex compositions of multiple attributes and objects when incorporating attribute correlations.
SymNet can be utilized for attribute learning, compositional zero-shot learning and outperforms the state-of-the-art on four widely-used benchmarks. 
Code is at \url{https://github.com/DirtyHarryLYL/SymNet}.
\end{abstract}

}

\maketitle

\IEEEdisplaynontitleabstractindextext

%
\IEEEpeerreviewmaketitle

\IEEEraisesectionheading{
\section{Introduction}}
\IEEEPARstart{A}{ttributes} describe the properties of generic objects, e.g., material, color, weight, etc. Understanding the attributes would directly facilitate many tasks that require deep semantics, such as scene graph generation~\cite{li2017scene}, object perception~\cite{faster}, human-object interaction detection~\cite{hicodet,li2019hake,li2019transferable,idn,djrn}. 
As side information, attributes can also be employed in zero-shot learning~\cite{apy,sun,awa2,cub,mit,ut}.

\begin{figure}[!ht]
	\begin{center}
		\includegraphics[width=0.45\textwidth]{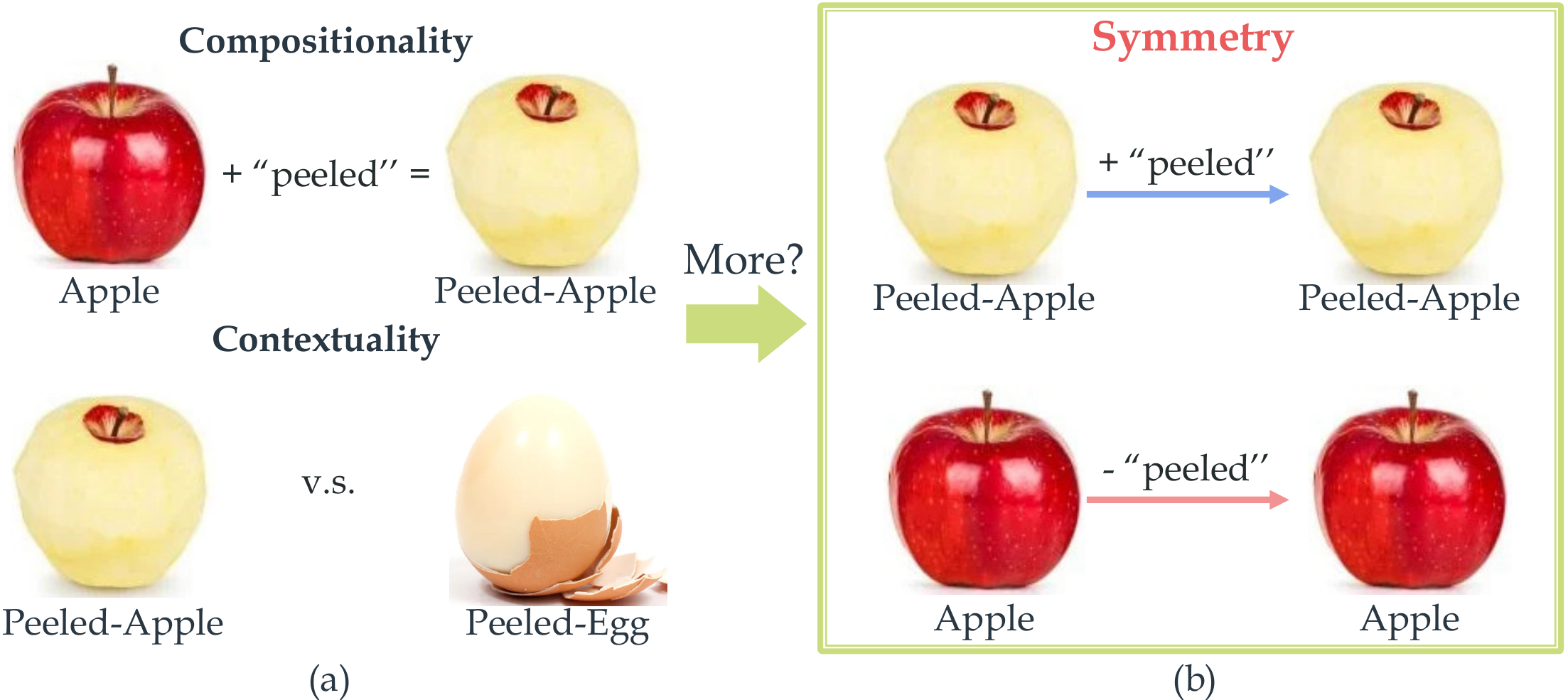}
	\end{center}
	\vspace{-0.3cm}
	\caption{Except for the compositionality and contextuality, attribute-object compositions also have the \textit{symmetry} property. For instance, a \texttt{peeled-apple} should not change after ``adding'' the \texttt{peeled} attribute. Similarly, an \texttt{apple} should keep the same after ``removing'' the \texttt{peeled} attribute because it does not have it.}
	\label{Figure:first-page}
	\vspace{-0.5cm}
\end{figure}

Going along with the road of conventional classification, some works~\cite{awa1,relativeattr,sun,cocoattr} address attribute recognition with discriminative models for objects but achieve poor performance. This is because attributes cannot be well expressed independently of the context~\cite{redwine,operator} (Fig.~\ref{Figure:first-page}(a)). 
Subsequently, researchers rethink the nature of attributes and treat them as linear operations~\cite{operator} to operate these general concepts: ``adding'' attribute to objects (coupling) or ``removing'' attribute from objects (decoupling). Though such insight has promoted this field, the ``add-remove'' system is not complete and lacks an axiomatics foundation to satisfy the specific principles of nature.

In this paper, we rethink the \textit{physical} and \textit{linguistic} properties of attribute-object, and propose a previously ignored but important principle of attribute-object transformations: \textbf{symmetry}, which would promote attribute-object learning.
Symmetry depicts the invariance under transformations, e.g., a circle has rotational symmetry under the rotation without changing its appearance. The transformation that ``adding'' or ``removing'' attributes should also satisfy the symmetry: an object should remain unchanged if we ``add'' an attribute it already has, or ``remove'' an attribute it does not have. For instance, a \texttt{peeled-apple} keeps invariant if we ``add'' attribute \texttt{peeled} upon it. Similarly, ``removing'' attribute \texttt{peeled} from \texttt{apple} would still result in \texttt{apple}.

As shown in Fig.~\ref{Figure:first-page}(b), except for the compositionality and contextuality, the symmetry should also be satisfied to guarantee rationality.
Given this, we first introduce the symmetry and propose \textbf{SymNet} to depict it.
In this work, we aim to bridge attribute-object learning and group theory. The elegant properties of groups would largely help in a more principled way, given its theoretical potential. Thus, to cover the principles existing in transformations theoretically, we borrow the principles from group theory to model symmetry. In detail, we define three transformations \{``keep'', ``add'', ``remove''\} and an operation to perform three transformations upon objects, in other words, to construct a ``group''.
To implement these, SymNet adopts Coupling Network (CoN) and Decoupling Network (DecoN) to perform coupling/adding and decoupling/removing.
To meet the fundamental requirements of group theory, \textit{symmetry} and the group axioms \textit{closure, associativity, identity element, invertibility element} are all implemented as the learning objectives of SymNet.
Naturally, SymNet considers the compositionality and contextuality during the coupling and decoupling of various attributes and objects.
All the above principles will be learned under a unified model in an end-to-end paradigm.

\begin{figure*}[!ht]
	\begin{center}
		\includegraphics[width=0.9\textwidth]{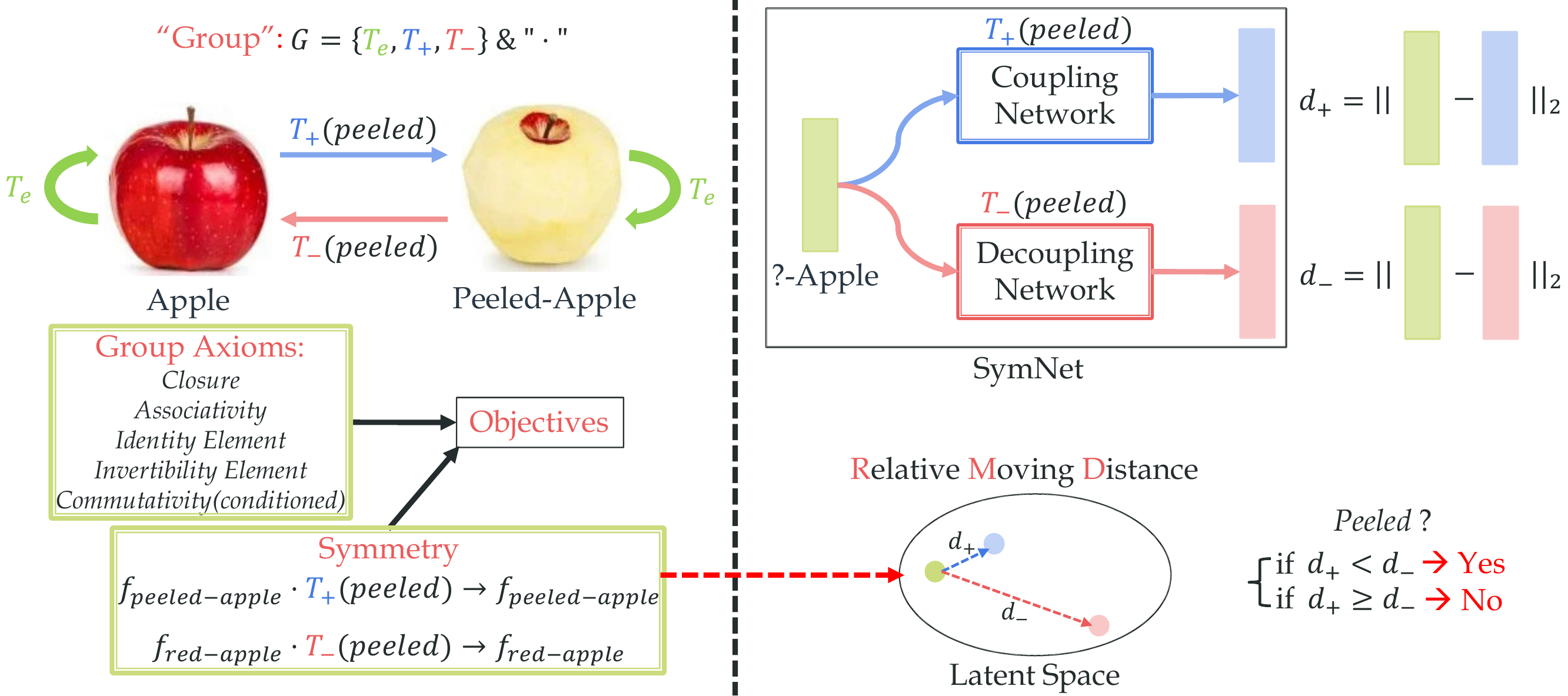}
	\end{center}
	\vspace{-0.3cm}
	\caption{Overview of our proposed method. We construct a ``group'' to learn the symmetry and operate the composition learning. The attribute transformations are implemented as coupling and decoupling networks and constrained by symmetry and group axiom objectives. Then relative moving distance based paradigm is applied in attribute classification.
	}
	\label{Figure:overview}
	\vspace{-0.3cm}
\end{figure*}

With symmetry learning, we can apply SymNet to address the Compositional Zero-Shot Learning (CZSL) task, whose target is to classify unseen compositions composed of seen attributes and objects.
We adopt a novel attribute recognition paradigm, \textbf{R}elative \textbf{M}oving \textbf{D}istance (\textbf{RMD}) (Fig.~\ref{Figure:overview}).
That said, given a specific attribute, an object would be manipulated by the ``add'' and ``remove'' transformations parallelly in \textit{latent} space.
Then, we can discriminate the existence of an attribute when transformations meet the symmetry principle (Sec.~\ref{sec:single-attr}): 
if the input object has this attribute, the output after addition should be close to the original input object, and the object after removal should be far from the input.
Contrarily, if the object does not have this attribute, the object after removal should be closer to the input than the object after addition.
So attribute classification can be accomplished concurrently by comparing the relative \textit{moving} distances between the input and two outputs.

In CZSL, the composition consists of a single attribute and an object~\cite{mit,ut}. However, in practical application, an object usually has multiple attributes simultaneously~\cite{apy,sun}. Thus, multi-attribute recognition has greater practical significance.
However, under the multi-attribute setting, the attribute \textbf{correlation} would complicate the RMD principles.
For example, when removing attribute \texttt{fresh} from objects, a \{\texttt{green}, \texttt{juicy}\} object is more likely to \textbf{change more} than a \{\texttt{black}, \texttt{hard}\} object in latent space. The reason is: though neither object has attribute \texttt{fresh}, attributes \{\texttt{green}, \texttt{juicy}\} are more closely related to \texttt{fresh} than \{\texttt{black}, \texttt{hard}\}.
Further experiments also show that vanilla RMD for single-attribute scenarios fails to model multi-attribute correlation.
Therefore, we further incorporate attribute correlation into RMD principles to adapt to the multi-attribute setting (Sec.~\ref{sec:att-corre}).

With RMD, we can utilize the robust \textit{attribute change} to classify attributes, instead of only relying on the dramatically unstable \textit{visual attribute patterns}. Extensive experiments show that our method achieves significant improvements on both single- and multi-attribute learning benchmarks~\cite{mit,ut,apy,sun}.
The main contributions of this work are: 
1) We propose a novel property of attribute-object composition transformation: symmetry, and design a framework inspired by group theory to learn it under the supervision of group axioms.
2) Based on symmetry learning, we propose a novel method to infer attributes based on relative moving distance (RMD).
3) We propose the corresponding RMD constraints to guide the learning for both single- and multi-attribute settings.
4) Substantial improvements are achieved in attribute recognition and CZSL tasks.

\section{Related Work}
\noindent{\bf Visual Attribute.}
The visual attribute was introduced into computer vision to reduce the gap between visual patterns and object concepts, such as reducing the difficulty in object recognition~\cite{apy} or acting as an intermediate representation for zero-shot learning~\cite{awa1,ALE}. 
After that, attribute has been widely applied in recognition of face~\cite{celebA}, people~\cite{poselet}, pedestrian~\cite{PETA} or action~\cite{ucf101}, zero-shot learning~\cite{cub,sun} and so on. Therefore, attribute recognition is a fundamental problem to promote visual concept understanding.

The typical approach for attribute recognition is to train a multi-attribute discriminative model same as object classification~\cite{awa1,relativeattr,sun,cocoattr}. It ignores the intrinsic properties of attributes, such as compositionality and contextuality. 
Farhadi~et~al.~\cite{apy} propose a visual feature selection method to recognize the attributes under the consideration of cross-category generalization. Gan et al.~\cite{UDICA} further enhance the generalization by integrating kernel alignment with distributional variance.
Liang et al.~\cite{UMF} think visual attributes are class-sensitive and utilize category information to predict attributes in a unified manner.
Attribute correlation is essential information and gets explicitly considered in some works~\cite{HAP,AMT,FMT,GALM}.
Choi et al.~\cite{HAP} propose a hyper-graph framework to learn the semantic attributes correlation and apply it to scene recognition.
Hand and Chellappa~\cite{AMT} designs a multi-task deep neural network with an auxiliary relation network for attribute prediction. 
Following these works, our SymNet explores attribute correlation from a novel perspective of attribute transformation.
Some works~\cite{FMT,GALM} apply automatic learning techniques to design deep neural networks for multi-task attribute learning automatically.
Tang~et~al.~\cite{tang2019improving} improve attribute recognition with weakly supervised attribute-specific localization.  
Later, some works start to consider the intrinsic properties by exploiting the attribute-object correlation~\cite{hwang2011sharing, analogous, mahajan2011joint}.
Considering the contextuality of attributes, Nagarajan~et~al.~\cite{operator} regard attributes as linear transformations operated upon object embeddings, and Misra~et~al.~\cite{redwine} map the attributes into model weight space to attain better representations. Adversarial learning is employed to model the discrepancy and correlations among attributes and objects.
Yang~et~al.~\cite{HMF} propose a hierarchical feature embedding framework with inter-class and intra-class relations. 
Li~et~al.~\cite{LI2020DOMAIN} propose a structural attribute learning framework to extract domain-invariant attribute features.
Moreover, multi-attribute compositions can also be used to describe objects in few-shot recognition~\cite{comp}.
Different from our attribute correlation constraint (Sec.~\ref{sec:att-corre}), Tokmakov~et~al.~\cite{comp} adopts an orthogonal constraint to deal with attribute correlation. In Suppl Sec.~4.3, we compare two constraints and illustrate their respective advantages.

\noindent{\bf Compositional Zero-Shot Learning.}
CZSL is a crossing field of compositional learning and zero-shot learning. In the CZSL setting, test compositions are unseen during training, while each component is seen in both the train and test sets. Chen~et~al.~\cite{analogous} construct linear classifiers for unseen compositions with tensor completion of weight vectors. 
Misra~et~al.~\cite{redwine} consider that the model space is more smooth, thus project the attributes or objects into it by training binary linear SVMs for the corresponding components. For CZSL, it composes attribute and object embeddings in model space as composition representation. 
Wang~et~al.~\cite{tafe} address the attribute-object compositional problem via conditional embedding modification, which relies on attribute word embedding~\cite{word2vec} transformation.
Nan~et~al.~\cite{genmodel} map the image features and word vectors~\cite{glove} into embedding space with the reconstruction constraint.
Nagarajan~et~al.~\cite{operator} regard attributes as linear operations for object embedding and map the image features and transformed object embeddings into a shared latent space. However, linear and explicit matrix transformation may be insufficient to represent various attribute concepts of different complexity, e.g., representing ``red'' and ``broken'' as matrices with the same capacity.
Very recently, Naeem~et~al.~\cite{2021learningGraphEmbeddings} use a graph to learn the dependency and relevance between attributes, objects, and compositions.
Previous methods usually ignored or incompletely considered the natural principles within the coupling and decoupling of attributes and objects. Hence, we propose a unified framework inspired by group theory to learn these essential principles such as symmetry.

\section{Approach}
Fig.~\ref{Figure:overview} gives an overview of our approach. Our goal is to learn the symmetry within attribute-object compositions. Thus we can utilize it to obtain a deeper understanding of attribute-object, e.g., to address the CZSL task~\cite{mit,ut}. 
To learn the symmetry in transformations, we need a comprehensive framework to cover all principles. 
Inspired by the group theory, we define a unified model named SymNet.

We define $G = \{T_e, T_+, T_-\}$ containing identity (``keep''), coupling (``add''), and decoupling (``remove'') transformations (Sec.~\ref{sec:define}) for each attribute and utilize Deep Neural Networks to implement them (Sec.~\ref{sec:implement}).
It is natural to adopt group theory as the close associations between symmetry and group to depict symmetry theoretically.
As a group should satisfy the axioms, i.e., \textit{closure, associativity, identity element, invertibility element}, we construct the learning objectives based on these axioms to train the transformations (Sec.~\ref{sec:constraints}).
In addition, SymNet satisfies the \textit{commutativity} under conditions.
With these constraints, we can naturally guarantee compositionality and contextuality.
\textit{Symmetry} allows us to use a novel method, relative moving distance, to identify whether an object has a certain attribute with $T_+$ and $T_-$ (Sec.~\ref{sec:att-classification}) for CZSL (Sec.~\ref{sec:czsl}).
 
\subsection{Group Definition}
\label{sec:define}
To depict the symmetry, we need first to define the transformations. Naturally, we need two reciprocal transformations to ``add'' and ``remove'' the attributes. 
Further, we need an axiomatic system to restrain the transformations and keep the rationality. 
Thus, we define three transformations $G = \{T_e, T_+, T_-\}$ and an operation ``$\cdot$''.
In practice, it is difficult to strictly follow the theory considering the physical and linguistic truth. 
For example, the operation between attribute transformations ``\texttt{peeled} $\cdot$ \texttt{broken}'' is odd.
Thus, our ``operation'' is defined to be operated upon object only.
\begin{definition}
\textbf{Identity transformation} $T_{e}$ keep the attributes of object.  \textbf{Coupling} transformation $T_+$ couples a specific attribute with an object. \textbf{Decoupling} transformation $T_{-}$ decouples a specific attribute from an object.
\end{definition}

\begin{definition}
Operation ``$\cdot$'' performs transformations $\{T_e,~T_+,~T_-\}$ upon an object. Notably, operation ``$\cdot$'' is not the dot product and we use this notation to maintain consistency with group theory. 
\end{definition}

More formally, for object $o \in \mathcal{O}$ and attribute $a^i, a^j \in \mathcal{A},~a^i\neq a^j$, where $\mathcal{O}$ denotes object set and $\mathcal{A}$ denotes attribute set, operation ``$\cdot$'' performs transformations in $G$ upon an object/image embedding:
\begin{eqnarray}
    \begin{aligned}
        & f_o^{i} \cdot T_+(a^j) = f_o^{ij},\\
        & f_o^{ij} \cdot T_-(a^j) = f_o^{i},\\
        & f_o^{i} \cdot T_e = f_o^{i},
    \end{aligned}    
\label{eq:3-T}
\end{eqnarray}
where $f_o^{i}$ means $o$ has one attribute $a^i$ and $f_o^{ij}$ means $o$ has two attributes $a^i, a^j$.
Here we do not sign a specific object category and use $o$ for simplicity. 

\begin{definition}
$G$ has the \textbf{symmetry} property if and only if $\forall a^i,~a^j \in \mathcal{A}, a^i\neq a^j$:
\begin{eqnarray}
    f_o^{i} = f_o^{i} \cdot T_+(a^i),~f_o^{i} = f_o^{i} \cdot T_-(a^j).
\label{eq:symmetry}
\end{eqnarray}
\end{definition}

\subsection{Group Implementation}
\label{sec:implement}
\begin{figure}[!t]
	\begin{center}
		\includegraphics[width=0.45\textwidth]{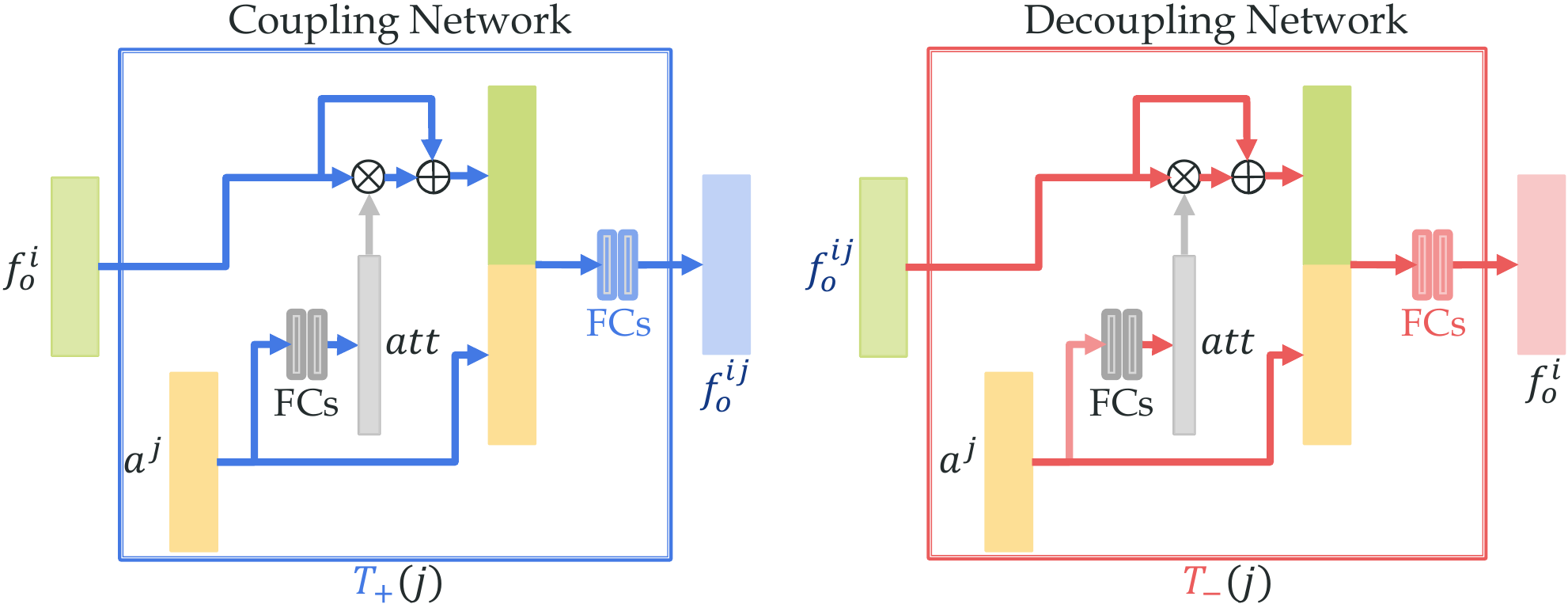}
	\end{center}
	\vspace{-0.3cm}
	\caption{The structure of CoN and DecoN. They take the attribute embedding to assign a specific attribute $a^j$. $f_o^i, f_o^{ij}$ are the object embeddings extracted from ResNet~\cite{resnet}. The attribute embeddings are converted to attentions and applied on object embeddings, then compressed by MLPs to output transformed representations.}
	\label{Figure:con-decon-structure} 
\vspace{-0.5cm}
\end{figure}

In practice, when performing $T_e$ upon $f_o^{i}$, we directly use $f_o^{i}$ as the $f_o^{i} \cdot T_e$ to implement the identity transformation for simplicity.
For $T_+,~T_-$, we propose \textbf{SymNet} which consists of two modules: Coupling Network (\textbf{CoN}) and Decoupling Network (\textbf{DecoN}).
CoN and DecoN have the same structure but independent weights and are trained with different tasks.
As seen in Fig.~\ref{Figure:con-decon-structure}, CoN and DecoN both take the object embedding $f_o^{i}$ and attribute embedding $a^j$ as inputs, and output the transformed object embedding.
We use attribute category word vectors such as GloVe~\cite{glove} or one-hot vectors to represent attributes.
$f_o^{i}$ is extracted by an ImageNet~\cite{imagenet} pre-trained ResNet~\cite{resnet} from image $I$, i.e., $f_o^{i} = F_{res}(I)$.

Intuitively, attributes affect objects differently, e.g., ``red'' changes the color, ``wooden'' changes the texture. 
In CoN and DecoN, we use an \textit{attribute-as-attention} strategy, i.e., using $att=g(a^j)$ as attention, where $g(\cdot)$ means two fully-connected (FC) and a Sigmoid layer.
We concatenate $f_o^{i} \circ att + f_o^{i}$ with original $a^j$ as the input and use two FC layers to perform the transformation.

\subsection{Group Axioms as Objectives}
\label{sec:constraints}
According to group theory, SymNet should satisfy four group axioms: \textit{closure, associativity, identity element,} and \textit{invertibility}. Under certain conditions, attribute-object also satisfy \textit{commutativity}.
Besides, SymNet must obey the symmetry property of the attribute transformations. 

In practice, we use Deep Neural Networks to implement transformations.  
Thus, we can construct training objectives to approach the theoretic transformations following the axioms.
Considering the \textit{actual characteristics} of attribute-object compositions, we slightly adjust the axioms to construct the objectives.
Besides, there are two situations with different forms of axioms: 1) coupling or decoupling an attribute $a^i$ that the object $f_o^i$ already has, or 2) coupling or decoupling an attribute $a^j$ that object $f_o^i$ does not have. 

\noindent {\bf Symmetry.}
First of all, SymNet should satisfy the symmetry property as depicted in Eq.~\ref{eq:symmetry}, i.e., $f_o^{i} = f_o^{i} \cdot T_+(a^i), f_o^{i} = f_o^{i} \cdot T_-(a^j)$.
The symmetry is essential to keep the semantic meaning during coupling and decoupling. For example, given a \texttt{peeled-egg}, adding the attribute \texttt{peeled} again should not change the object state. Similarly, a \texttt{cup} without attribute \texttt{broken} should remain unchanged after removing \texttt{broken}.
Thus, we construct the \textbf{symmetry loss}:
\begin{eqnarray}
    \begin{aligned}
        \mathcal{L}_{sym} = \|f_o^{i} - f_o^{i} \cdot T_+(a^i)\|_2 +
        \|f_o^{i} - f_o^{i} \cdot T_-(a^j)\|_2,
    \end{aligned}
\label{eq:symmetry-loss}
\end{eqnarray}
where $a^i, a^j \in \mathcal{A}, i \neq j$. We use $L_2$ norm loss to measure the distance between two embeddings.

\noindent {\bf Closure.} 
For all elements in set $G$, their operation results should also be in $G$. In SymNet, for the attribute $a^i$ that $f_o^{i}$ has, $f_o^{i} \cdot T_+(a^i) \cdot T_-(a^i)$ should be equal to $f_o^{i} \cdot T_-(a^i)$. 
For the attribute $a^j$ that $f_o^{i}$ does not have, $f_o^{i} \cdot T_-(a^j) \cdot T_+(a^j)$ should be equal to $f_o^{i} \cdot T_+(a^j)$.
Thus, we construct:
\begin{eqnarray}
    \begin{aligned}
        \mathcal{L}_{clo} = & \|f_o^{i} \cdot T_+(a^i) \cdot T_-(a^i) -  f_o^{i} \cdot T_-(a^i)\|_2 + 
        \\ & \|f_o^{i} \cdot T_-(a^j) \cdot T_+(a^j) -  f_o^{i} \cdot T_+(a^j)\|_2.
    \end{aligned}
\label{eq:closure}
\end{eqnarray}

\noindent {\bf Identity Element.}
The properties of identity element $T_e$ are automatically satisfied since we implement $T_e$ as a skip connection, i.e., $f_o^i\cdot T_*(a^i)\cdot T_e=f_o^i\cdot T_e\cdot T_*(a^i)=f_o^i\cdot T_*(a^i)$ where $T_*$ denotes any element in $G$.

\noindent {\bf Invertibility Element.}
According to the definition, $T_+$ is the invertibility element of $T_-$, vice versa. 
For the attribute $a^i$ that $f_o^{i}$ has, $f_o^{i} \cdot T_-(a^i) \cdot T_+(a^i)$ should be equal to $f_o^{i} \cdot T_e = f_o^{i}$.
For the attribute $a^j$ that $f_o^{i}$ does not have, $f_o^{i} \cdot T_+(a^j) \cdot T_-(a^j)$ should be equal to $f_o^{i} \cdot T_e = f_o^{i}$.
Therefore, we have:
\begin{eqnarray}
    \begin{aligned}
        \mathcal{L}_{inv} = & \|f_o^{i} \cdot T_+(a^j) \cdot T_-(a^j) - f_o^{i} \cdot T_e\|_2 + 
        \\ & \|f_o^{i} \cdot T_-(a^i) \cdot T_+(a^i) - f_o^{i} \cdot T_e\|_2.
    \end{aligned}
\label{eq:invertibility}
\end{eqnarray}

\noindent {\bf Associativity.} 
Because of the practical physical meaning of attribute-object compositions, we only define the operation ``$\cdot$'' that operates a transformation upon an object embedding in Sec.~\ref{sec:define} and do not define the operation between transformations. 
Thus, we \textit{relax} the constraint and do not construct an objective according to associativity in practice.

\noindent {\bf Commutativity.} Because of the specialty of attribute-object, SymNet satisfies the \textit{commutativity} when coupling and decoupling \textit{multiple} attributes. Thus, $f_o^{i} \cdot T_+(a^i) \cdot T_-(a^j)$ should be equal to $f_o^{i} \cdot T_-(a^j) \cdot T_+(a^i)$:
\begin{eqnarray}
    \begin{aligned}
    \mathcal{L}_{com} = \|&f_o^{i} \cdot T_+(a^i) \cdot T_-(a^j) - \\
    &f_o^{i} \cdot T_-(a^j) \cdot T_+(a^i)\|_2.
    \end{aligned}
\label{eq:commutativity}
\end{eqnarray}
Although the above definitions do not strictly follow the theory, the \textit{loosely} conducted axiom objectives still contribute to the robustness and effectiveness a lot (ablation study in Sec.~\ref{sec:ablation}) and open the door to a more theoretical way.

Last but not least, CoN and DecoN need to keep the \textbf{semantic consistency}, i.e., before and after the transformation, the \textit{object category} should not change. 
Hence, we use a cross-entropy loss $\mathcal{L}_{cls}^o$ for the object recognition of the input and output embeddings of CoN and DecoN.
In the same way, before and after coupling and decoupling, the \textit{attribute changes} provide the attribute classification loss $\mathcal{L}_{cls}^a$.
We use typical visual pattern-based classifiers consisting of FC layers for the object and attribute classification.

\subsection{Relative Moving Distance} 
\label{sec:att-classification}
\begin{figure}[!t]
	\begin{center}
		\includegraphics[width=0.45\textwidth]{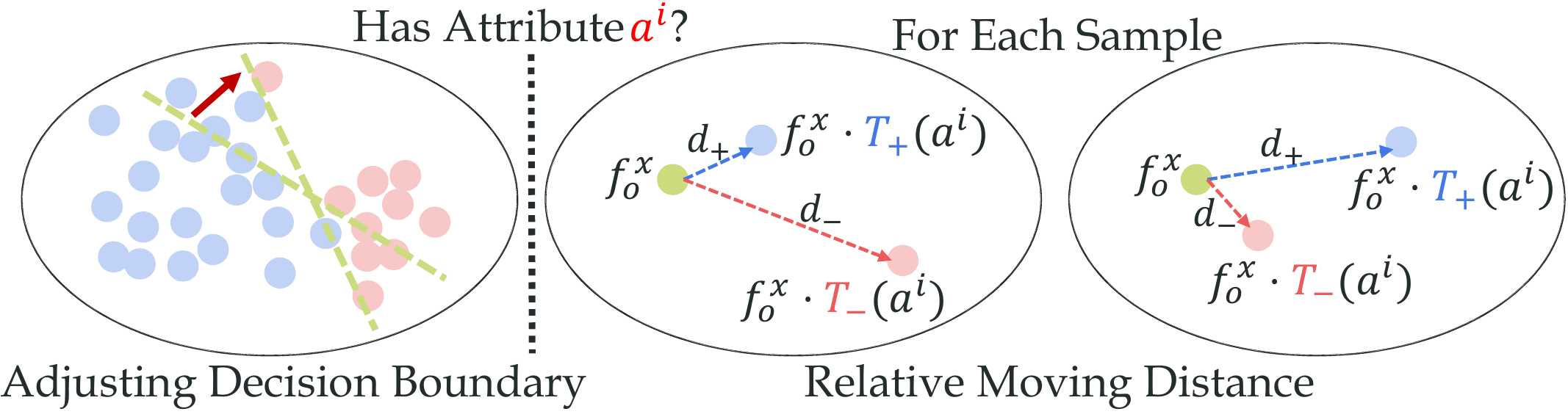}
	\end{center}
	\vspace{-0.3cm}
	\caption{Comparison between typical method and relative moving distance (RMD) based recognition. Previous methods mainly try to adjust the decision boundary. RMD based approach moves the embedding with $T_+$ and $T_-$ and classifies by comparing their moving distances.}
	\label{Figure:rmd}
\vspace{-0.5cm}
\end{figure}

As shown in Fig.~\ref{Figure:rmd}, we utilize the relative moving distance (RMD) based on the symmetry property to operate the attribute recognition. The implementations of RMD in single- and multi-attribute scenarios have minor differences.

\subsubsection{Single-attribute RMD}
\label{sec:single-attr}
Given an image embedding $f_o^{x}$ of an object with an unknown attribute $a^x$, we input it to both CoN and DecoN with all kinds of attribute word embeddings $\{a^1, a^2, ... a^n\}$ where $n$ is the number of attributes.
Two transformers would take attribute embeddings as conditions and operate coupling and decoupling \textit{in parallel}, then output $2n$ transformed embeddings $\{f_o^{x} \cdot T_+(a^1), f_o^{x} \cdot T_+(a^2),..., f_o^{x} \cdot T_+(a^n)\}$ and $\{f_o^{x} \cdot T_-(a^1), f_o^{x} \cdot T_-(a^2), ..., f_o^{x} \cdot T_-(a^n)\}$.
We compute the distances between $f_o^{x}$ and the transformed embeddings:
\begin{eqnarray}
    \begin{aligned}
        & d^{i}_+ = \|f_o^{x} - f_o^{x} \cdot T_+(a^i)\|_2,\\
        & d^{i}_- = \|f_o^{x} - f_o^{x} \cdot T_-(a^i)\|_2.
    \end{aligned}
\label{eq:distances}
\end{eqnarray}
To compare two distances, we define \textit{relative moving distance} as $d^{i}=d^{i}_--d^{i}_+$ and perform binary classification for each attribute (Fig.~\ref{Figure:rmd}): 
1) If $d^{i} \geq 0$, i.e., $f_o^{x} \cdot T_+(a^i)$ is closer to $f_o^i$ than $f_o^{x} \cdot T_-(a^i)$, we tend to believe $f_o^x$ has attribute $a^i$.
2) If $d^{i} < 0$, i.e., $f_o^{x} \cdot T_-(a^i)$ is closer, we tend to predict that $f_o^x$ does not have attribute $a^i$.
Previous zero/few-shot learning methods usually classify the instances via measuring the distance between the embedded instances and \textbf{fixed} points like prototype/label/centroid embeddings.
Differently, relative {\bf moving} distance compares the distance before and after applying the coupling and decoupling operations. 

To enhance the RMD-based classification performance, we further use a triplet loss function. 
Let $\mathcal{X}$ denote the attribute that $f_o^x$ has, the loss can be described as:
\vspace{-3mm}
\begin{eqnarray}
    \mathcal{L}_{tri} = \sum_{i}^{\mathcal{X}} [d^{i}_+ - d^{i}_- + \alpha]_{+} + \sum_{j}^{\mathcal{A}-\mathcal{X}} [d^{j}_- - d^{j}_+ + \alpha]_{+},
\label{eq:triplet-loss}
\end{eqnarray}
where $\alpha$=0.5 is triplet margin, $[\cdot]_+=max(\cdot, 0)$. 
$d^{i}_+$ should be less than $d^{i}_-$ for the attribute that $f_o^x$ has and greater than $d^{i}_-$ for the attribute $f_o^x$ does not have.
The total loss is
\begin{eqnarray}
\begin{aligned}
    \mathcal{L}_{total} = &\lambda_{1}\mathcal{L}_{sym} + \lambda_{2}\mathcal{L}_{axiom}+\lambda_{3}\mathcal{L}_{cls}^a \\
    &+ \lambda_{4}\mathcal{L}_{cls}^o + \lambda_{5}\mathcal{L}_{tri},
\end{aligned}
\label{eq:total-loss}
\end{eqnarray}
where $\mathcal{L}_{axiom} = \mathcal{L}_{clo} + \mathcal{L}_{inv} + \mathcal{L}_{com}$.

\subsubsection{Multi-attribute RMD}
\label{sec:att-corre}
\begin{figure}[h]
	\begin{center}
		\includegraphics[width=0.45\textwidth]{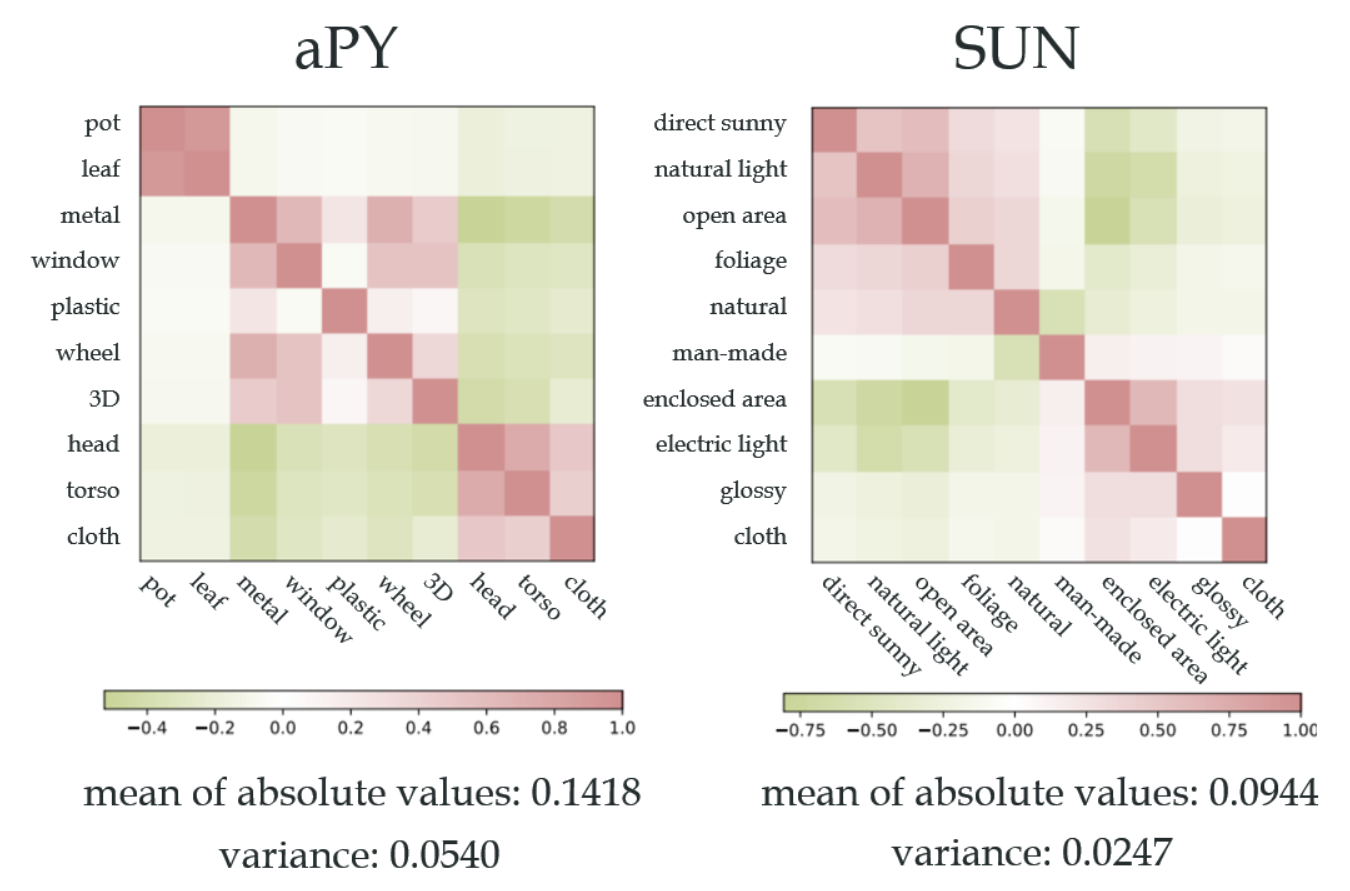}
	\end{center}
	\vspace{-0.3cm}
	\caption{Attribute correlation matrices of aPY~\cite{apy} and SUN~\cite{sun}. The correlation can be positive or negative ([-1,1], green to red). We calculate the mean and variance of the \textit{absolute} correlation values, which show that aPY~\cite{apy} contains \textit{stronger} but more \textit{variable} correlations.}
	\label{Fig:attr-corr}
	\vspace{-0.5cm}
\end{figure}

\begin{figure*}[!ht]
	\begin{center}
		\includegraphics[width=0.8\textwidth]{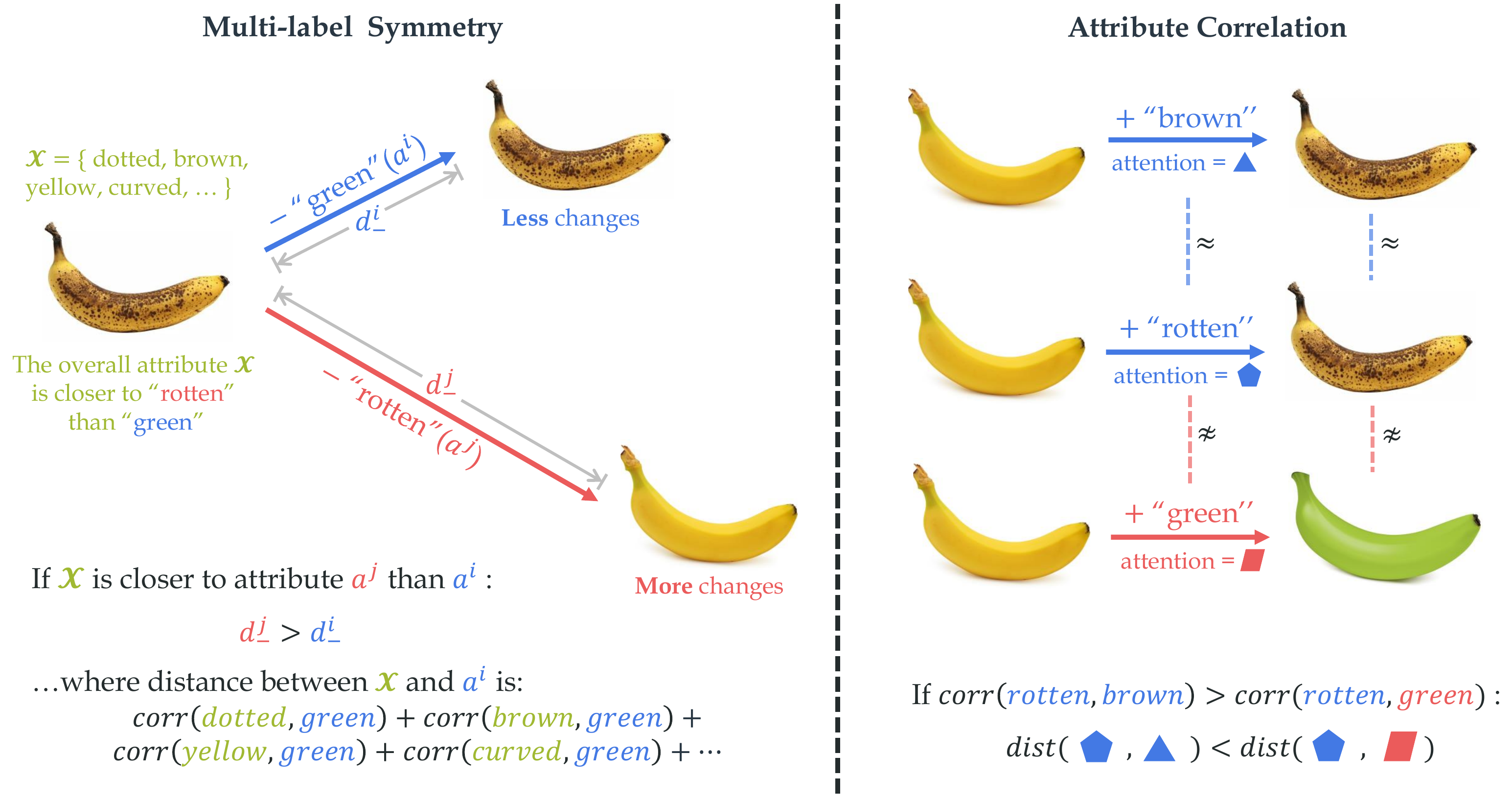}
	\end{center}
	\vspace{-0.3cm}
	\caption{Under multi-attribute/label setting, RMD need to consider more constraints since the existence of attribute correlation. 
	(Left) For example, for a banana with existing attributes $\mathcal{X}$, since $\mathcal{X}$ are closely related to \texttt{rotten} but have weak connection with \texttt{green}, the results differ when ``removing'' attribute \texttt{rotten} and \texttt{green} respectively. 
	In single-attribute RMD, the results should be comparably close to the original banana, since this banana has neither attribute \texttt{rotten} nor \texttt{green}.
	But in multi-attribute RMD, the moving distance depends on the similarity of $\mathcal{X}$ and the operated attribute. 
	In general, removing an attribute which is more correlated to $\mathcal{X}$ will make a larger difference, and the less correlated removing would make a minor difference.
	(Right) Meanwhile, for operated attributes, correlation would also affect the generated attentions: more correlated attributes should generate more similar attentions.
	}
	\label{Figure:multi-overview}
	\vspace{-0.3cm}
\end{figure*}

When an object has multiple attributes, the RMD-based paradigm should be amended due to the attribute correlation. Fig.~\ref{Fig:attr-corr} illustrates the correlation matrices of attributes in aPY~\cite{apy} and SUN~\cite{sun}, we can find that some highly correlated attributes exist. 
In transformations, the correlations between the removed attributes and existing attributes of an object should be considered.
We modify the RMD triplet loss $\mathcal{L}_{tri}$ from the following two aspects, as shown in Fig.~\ref{Figure:multi-overview}.

\begin{figure}[h]
    \begin{center}
        \begin{minipage}{0.8\linewidth}
            \centerline{\includegraphics[width=\linewidth]{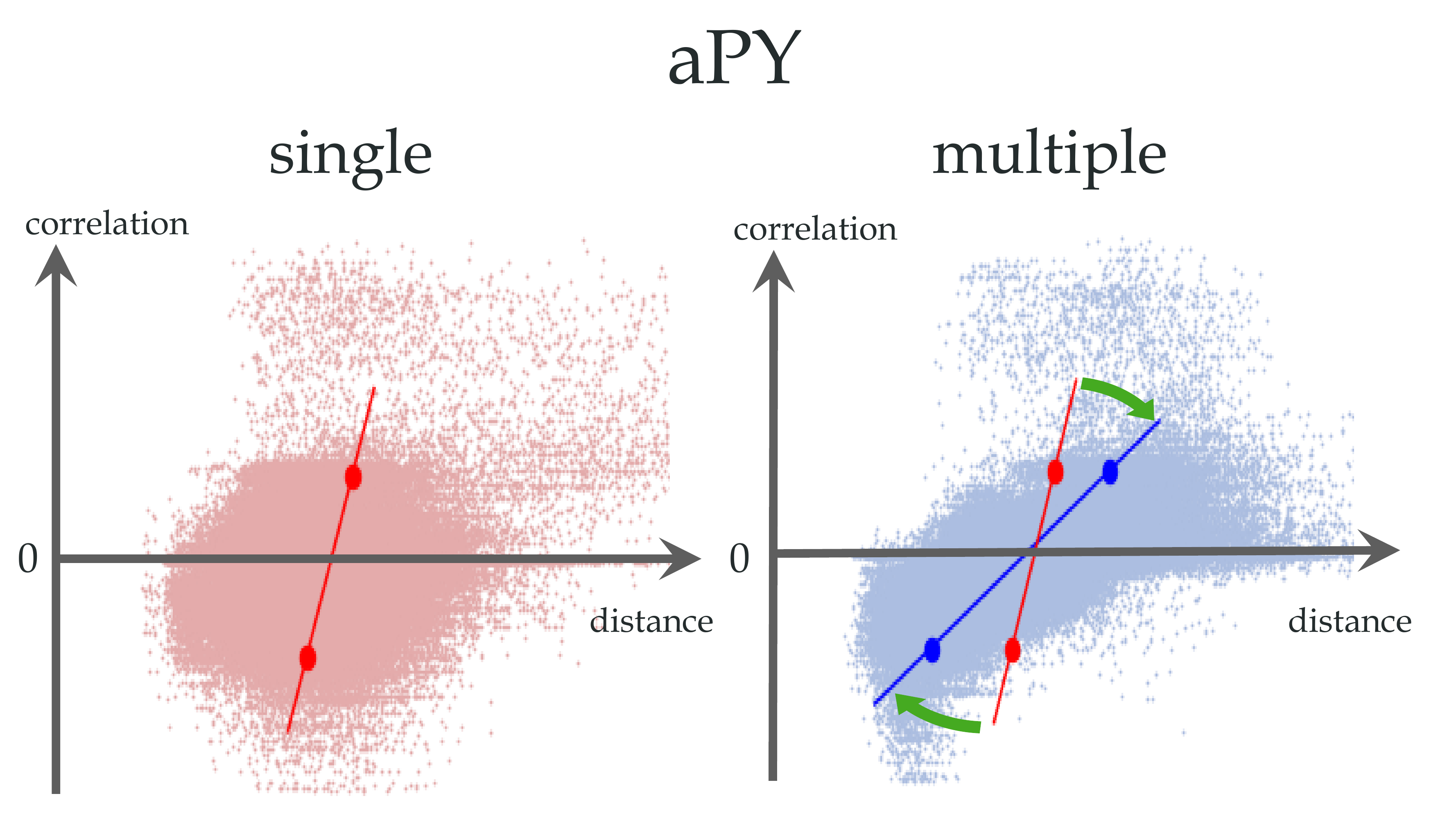}}
        \end{minipage}
        \vfill
        \begin{minipage}{0.8\linewidth}
            \centerline{\includegraphics[width=\linewidth]{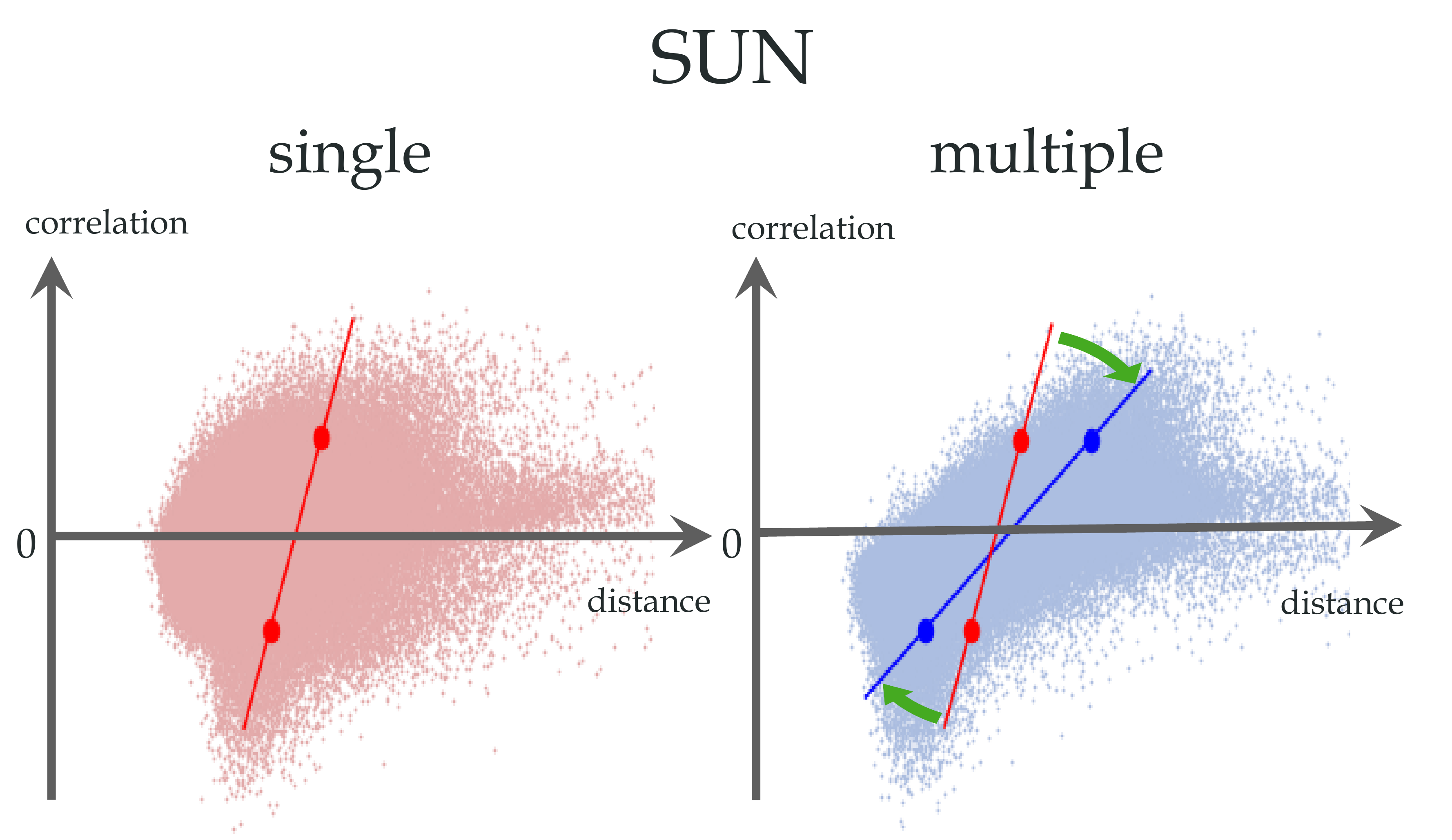}}
        \end{minipage}
    \end{center}
    \vspace{-0.3cm}
	\caption{The moving distances after attribute \textbf{removal} on multi-attribute benchmarks. The distance is always positive, and the correlation can be positive ($>0$) or negative ($<0$).
	According to our setting, \textbf{larger} positive correlation would generate \textbf{larger} moving distance in removal, e.g., removing attribute \texttt{rotten} in Fig.~\ref{Figure:multi-overview} (left). 
	Oppositely, the \textbf{smaller} negative correlation could make the moving distance \textbf{smaller}, e.g., removing attribute \texttt{green} in Fig.~\ref{Figure:multi-overview} (left). 
	The bold dots indicate the centroids of the dots with top-10\% and down-10\% correlation values. Thus, the line linked to two bold points should have a smaller slope ideally.
	The model is trained with single (left, red) or multiple (right, blue) RMD settings.
	We can find that the multi-attribute RMD generate more reasonable distances, i.e., has a smaller slope (blue lines).} 
	\label{Figure:corr-dist-scatter} 
	\vspace{-0.3cm}
\end{figure}

\noindent\textbf{Multi-attribute Symmetry Constraint}.
Under the single-attribute setting, symmetry is strictly satisfied, i.e., the moving distance after adding an existing attribute (an object has) or removing a non-existing attribute (an object does not have) is \textit{zero}. When it comes to the multi-attribute setting, this distance can be a small \textit{non-zero} value due to the attribute correlation. 
For instance, if we \textbf{remove} attribute \texttt{metallic} from a \{\texttt{small, lustrous}\} object, it may lose part of attribute \texttt{lustrous}, leading to a small but non-zero moving distance. 
A similar example is shown in Fig.~\ref{Figure:multi-overview} (left).
In the multi-attribute setting, for an object, we denote $\mathcal{X}$ as its \textbf{all existing attributes}.
Then, if attribute $a^j$ (marked red in the figure) is more positively related to $\mathcal{X}$ than $a^i$ (marked blue), removing $a^j$ would lead to a larger embedding moving distance.
The above insight can be concluded into one simple sentence: for an object, removing an attribute that is more correlated to its existing attributes $\mathcal{X}$ will make a more significant difference, and a less correlated attribute removing would make a minor difference.

Precisely, the correlations ([-1,1]) are measured via directly calculating the co-occurrence of attributes in the train set of multi-attribute benchmarks~\cite{apy,sun}. 
For attributes $a^i$ and $a^j$, let $Y^i,Y^j$ denote their $n$-dimensional label vector among $n$ total object instances in the train set, the \textbf{correlation coefficient} $corr(a^i,a^j)$ of $a^i$ and $a^j$ is measured as:
\begin{eqnarray}
    \begin{aligned}
        corr(a^i,a^j)=\frac{cov(Y^i,Y^j)}{\sqrt{cov(Y^i,Y^i)}\sqrt{cov(Y^j,Y^j)}},
    \end{aligned}
\label{eq:calculate-correlation}
\end{eqnarray}
where $cov(Y^i,Y^j)=\frac1n \left(Y^i-\overline{Y^i}\right)^T\left(Y^j-\overline{Y^j}\right)$ and $\overline{Y}$ denotes the mean value of vector $Y$. 
The definition of correlation can be generalized to that between attribute $a^i$ and $\mathcal{X}$ of an object without much effort:
\begin{eqnarray}
    \begin{aligned}
        corr(a^i,\mathcal{X}) = 
        \sum_{a^j \in \mathcal{X}}corr(a^i,a^j).
    \end{aligned}
\label{eq:calculate-correlation-X}
\end{eqnarray}

We plot the moving distance after removing an attribute w.r.t. the correlation (between the removed attribute and $\mathcal{X}$) in Fig.~\ref{Figure:corr-dist-scatter}.
The moving distance and correlation of an ideal model should have a monotonically increasing relation (larger correlation causes a larger distance/change), while Fig.~\ref{Figure:corr-dist-scatter} shows that this relation is not well-learned by the SymNet with vanilla single-attribute symmetry since attributes are regarded as independent. 
Thus, we design an extra constraint in multi-attribute scenario. 
For an object with existing attributes $\mathcal{X}$, we randomly sample two \textbf{non-existing} attributes $a^i$ and $a^j$ (Fig.~\ref{Figure:multi-overview}) and incorporate this property as a weighted triplet loss term:
\begin{eqnarray}
\begin{aligned}
    \mathcal{L}_{tri}^{sym} = \left[ (
        corr(a^i,\mathcal{X})
        - corr(a^j,\mathcal{X})
    )(d_-^{j} - d_-^{i}) + \alpha \right]_{+},
\end{aligned}
\label{eq:relative-loss}
\end{eqnarray}
where $d_-^{i}$ is the moving distance as Eq.~\ref{eq:distances}, $\alpha$ is the triplet margin, and $[\cdot]_+=max(\cdot, 0)$. Thus:

{\bf 1)} 
If $corr(a^i,\mathcal{X}) > corr(a^j,\mathcal{X})$, i.e., $a^i$ is more correlated (larger correlation) with $\mathcal{X}$ than $a^j$, then the triplet loss will reduce $d_-^{j}$ to make $d_-^{j}<d_-^{i}$.

{\bf 2)} 
If $corr(a^i,\mathcal{X}) < corr(a^j,\mathcal{X})$, i.e., $a^j$ is more correlated with $\mathcal{X}$, the triplet loss will reduce $d_-^{i}$ to achieve $d_-^{j}>d_-^{i}$.

\noindent\textbf{Attribute Correlation Constraint}.
The correlation among attributes can also help attribute recognition since similar attributes similarly transform the object. 
For example, a \{\textit{muddy}, \textit{dusty}\} object is very likely to be \textit{dirty} since their high correlations. 
If a sub-optimal classifier gives high confidence to \textit{muddy} and \textit{dusty} but low confidence to \textit{dirty}, incorporating the attribute correlation will help the classifier give consistent predictions on three similar attributes and therefore boost the performance.

The attention vectors from \textit{attribute-as-attention} strategy imply how attributes manipulate object embeddings, so we implement the above correlation constraint as an auxiliary loss on generated attention. 
For each training object, we randomly sample three attributes $a^i,~a^j,~a^k$ and compute their correlations $corr(a^i, a^j),~corr(a^i,a^k)$ via Eq.~\ref{eq:calculate-correlation}. Then we construct an auxiliary triplet loss:
\begin{eqnarray}
\begin{aligned}
    & \mathcal{L}_{tri}^{corr}  \\
    & =  \left[ \left(corr(a^i,a^j)-corr(a^i,a^k)\right)(d_+^{ij}-d_+^{ik}) + \alpha \right]_{+} \\
    &+  \left[ \left(corr(a^i,a^j)-corr(a^i,a^k)\right)(d_-^{ij}-d_-^{ik}) + \alpha \right]_{+},
\end{aligned}
\label{eq:attr-corr-loss}
\end{eqnarray}
where $\alpha$ is triplet margin, $[\cdot]_+=max(\cdot, 0)$. 
$d_+^{ij}=dist(att_+^i,att_+^j)$ is the distance between attribute attention vectors in $T_+(a^i)$ and $T_+(a^j)$, and $d_-^{ij}=dist(att_-^i,att_-^j)$ is that in $T_-(a^i)$ and $T_-(a^j)$. 
The triplet loss is weighted by correlation difference $corr(a^i,a^j)-corr(a^i,a^k)$:

{\bf 1)} 
If $corr(a^i,a^j)>corr(a^i,a^k)$, i.e., $a^i$ and $a^j$ are more correlated, the loss will reduce $d_+^{ij}$ and $d_-^{ij}$ to make the attentions of $a^i$ and $a^j$ more similar, so $d_+^{ij}<d_+^{ik}, d_-^{ij}<d_-^{ik}$.

{\bf 2)}
If $corr(a^i,a^j)<corr(a^i,a^k)$, i.e., $a^i$ and $a^k$ are more correlated, the loss will reduce $d_+^{ik}$ and $d_-^{ik}$ to make the attentions of $a^i$ and $a^k$ more similar, so $d_+^{ij}>d_+^{ik}, d_-^{ij}>d_-^{ik}$.

The complete triplet loss in multi-attribute setting is
\begin{eqnarray}
\begin{aligned}
    \mathcal{L}_{tri} = & \sum_{i}^{\mathcal{X}} [d^{i}_+ - d^{i}_- + \alpha]_{+} + \sum_{j}^{\mathcal{A}-\mathcal{X}} [d^{j}_- - d^{j}_+ + \alpha]_{+} \\
    & + \lambda_6 \mathcal{L}_{tri}^{sym} + \lambda_7 \mathcal{L}_{tri}^{corr}.
\end{aligned}
\label{eq:triplet-loss-multi}
\end{eqnarray}
Moreover, the total loss format is the same as Eq.~\ref{eq:total-loss}.

\subsubsection{Inference}
In practice, for $n$ attribute categories, we use RMDs $d=\{d^{i}\}^n_{i=1}$ as the attribute scores, i.e., $\mathcal{S}_a = \{\mathcal{S}_a^i\}^n_{i=1} = \{d^{i}\}^n_{i=1}$ and obtain attribute probability with Sigmoid function: $p_a^i = Sigmoid(\mathcal{S}_a^i)$.
Notably, we also consider the scale and use a factor $\gamma$ to adjust the score before Sigmoid.
Our method can be operated in parallel, i.e., simultaneously computing the RMD values of $n$ attributes.
We input $[B, n, m]$ sized tensor where $B$ is the mini-batch size and $m$ is the object embedding size. CoN and DecoN would output two $[B, n, m]$ sized embeddings after transformation. Then we can compute RMDs $\{d^{i}\}^n_{i=1}$ simultaneously.
Our method has approximately the same speed as a typical FC classifier. The inference speed from features to RMD is 41.0 FPS, and the FC classifier speed is 45.8 FPS. The gap can be further omitted if considering the overhead of the feature extractor.

\subsection{Discussion: Composition Zero-Shot Learning}
\label{sec:czsl}
With robust and effective symmetry learning for attribute-object, we can further apply SymNet to CZSL~\cite{mit,ut}.
The goal of CZSL is to infer the unseen attribute-object pairs in the test set, i.e., a prediction is true positive if and only if both attribute and object classifications are accurate. The pair candidates are available during testing. Thus, the predictions of impossible pairs can be masked. 

We propose a novel method to address this task based on RMD. With relative moving distance $d^{i}=d^{i}_--d^{i}_+$, the attribute probability is computed as $p_a^i = Sigmoid(d^{i})$.
For the object category, we input the object embedding to 2-layer FC with Softmax to obtain the object scores $\mathcal{S}_o = \{\mathcal{S}_o^j\}^m_{j=1}$, where $m$ is the number of object categories. The object probability $p_j=Softmax(\mathcal{S}_o^j)$ and $p_o = \{p_o^i\}^m_{j=1}$.
We then use $p_{ao}^{ij}$ to represent the probability of an attribute-object pair in the test set, which is composed of the $i$-th attribute category and $j$-th object.
The pair probabilities are given by $p_{ao}^{ij} = p_a^i \times p_o^j$. The impossible compositions would be masked according to the benchmarks~\cite{mit,ut}.

\section{Experiment}
\label{sec:experiment}
\subsection{Data and Metrics}
Our experiments are conducted on the following datasets (Suppl Sec.~2): MIT-States~\cite{mit} and UT-Zappos50K~\cite{ut} (single-attribute), aPY~\cite{apy} and SUN~\cite{sun} (multi-attribute).

In attribute recognition, for aPY and SUN, we report mAUC following previous methods such as \cite{UDICA,GALM}. For MIT-States and UT-Zappos, we report Top-1 accuracy following \cite{genmodel}.
The CZSL experiments are conducted on MIT-States and UT-Zappos, where the training and testing pairs are non-overlapping, i.e., the test set contains unseen attribute-object pairs composed of seen attributes and objects. 
We report the Top-1, 2, 3 accuracies on the unseen test set. 
We also evaluate our model under the generalized CZSL setting of TMN~\cite{tmn}, since the "open world" setting from \cite{operator} brings biases towards unseen pairs~\cite{chao2016empirical}.

\subsection{Baselines}
\label{sec:baseline}
We compare SymNet with previous state-of-the-arts (detailed in Suppl Sec.~2).
For single-attribute learning and CZSL, we adopt the Visual Product, LabelEmbed (LE)~\cite{redwine}, LabelEmbed Only Regression (LEOR)~\cite{redwine}, LabelEmbed With Regression (LE+R)~\cite{redwine}, LabelEmbed+~\cite{operator}, AnalogousAttr~\cite{analogous}, Red Wine~\cite{redwine}, AttrOperator~\cite{operator}, TAFE-Net~\cite{tafe}, GenModel~\cite{genmodel}, f-CLSWGAN~\cite{featuregen}, TMN~\cite{tmn}, and Causal~\cite{causal-czsl}.
For multi-attribute learning, the ALE~\cite{ALE}, HAP~\cite{HAP}, UDICA/KDICA~\cite{UDICA}, UMF~\cite{UMF}, AMF~\cite{AMT}, FMT~\cite{FMT}, GALM~\cite{GALM} are adopted.

\subsection{Implementation Details}
We use ImageNet pre-trained ResNet-18~\cite{resnet} as backbone to extract features for MIT-States and UT-Zappos, ResNet-50~\cite{resnet} for aPY and SUN. Especially, since images in aPY may have several instances, we use the feature after RoI-pooling~\cite{faster}. We do not fine-tune it following previous methods. The 300-dimensional pre-trained GloVe~\cite{glove} vectors are used as the word embeddings. Moreover, SymNet is trained with an SGD optimizer on a single NVIDIA GPU. 

On MIT-States and UT-Zappos, the 512-dimensional ResNet-18 feature is first transformed to 300-dimensional by a single FC. On aPY and SUN, the 2048-dimensional ResNet-50 feature is compressed to 128-dimensional via an FC too.
The main modules of our SymNet, CoN, and DecoN, have the same structures but independent weights as depicted in Fig.~\ref{Figure:con-decon-structure}: two FC layers (512/300 on MIT-States UT-Zappos, 512/128 on aPY SUN) with Sigmoid convert the attribute embedding to attention with the same dimension for the object embedding. The attention is multiplied to the input object embedding and then gets summed with a shortcut of original object embedding. Next, the object embedding after the attention operation is concatenated to the attribute embedding and then compressed to the original dimension by the other two FC layers (256/128 on aPY, 1536/128 on SUN, 768/300 on MIT-States UT-Zappos). Each hidden FC in CoN and DecoN is followed by BatchNorm and ReLU.

On single-attribute benchmarks, for each training image, we randomly sample another image with the same object but a different attribute as the negative sample to compute $L_{total}$. (Sec.~\ref{sec:constraints}). 
On multi-attribute benchmarks, we compute $L_{sym},L_{axiom},L_{tri},L_{cls} (\mathcal{L}_{cls}^a, \mathcal{L}_{cls}^o)$ with all attributes and randomly sample three different attributes to compute $L_{tri}^{corr}$ for each object. Besides, we define the correlation of \textit{an attribute} to \textit{an object} as the sum of correlations between this attribute and $\mathcal{X}$ of this object (Eq.~\ref{eq:calculate-correlation-X}). 
According to the order of correlation, we regard top-10\% and last-10\% attributes as the strongly related ones, middle-10\% attributes as the neutral ones. One strongly related attribute and one neutral attribute would form a pair, and all such pairs are used to compute $L_{tri}^{sym}$. 
In the multi-attribute setting, coupling and decoupling already involve all attributes to implement the commutativity loss.

We use cross-validation to determine the hyper-parameters (Suppl Tab.~1).
The weights on datasets are different as their different domains/ranges/scales, leading to distinct embedding spaces and different parameters.

\subsection{Single-Attribute Learning}
We first compare the attribute learning alone on two single-attribute benchmarks in Tab.~\ref{tab:att}.
We reproduce the results of AttrOperator~\cite{operator} with its open-sourced code. For all methods involved, the individual attribute and object accuracy do not consider the relations between attributes and objects. 
The object recognition module of our method is a simple 3-layer MLP classifier with the visual image features from the ResNet-18 backbone.
SymNet outperforms previous methods by a large margin, i.e., 3.8\% on MIT-States and 8.3\% on UT-Zappos, which strongly verifies that our RMD-based attribute learning is particularly effective. 

\subsection{Multi-Attribute Learning} 
Next, we evaluate SymNet on multi-attribute learning benchmarks aPY\cite{apy} and SUN\cite{sun} in Tab.~\ref{tab:att-multi}.
SymNet still outperforms the state-of-the-art by 1.4\% on aPY, 1.9\% on SUN.
The reason is the advantages of SymNet as a transformation-style framework considering the attribute correlation explicitly.
Previous methods ignore the compositional nature of attributes and objects, e.g., UMF~\cite{UMF} directly uses the image and object representations in latent space to predict attributes. Relatively, SymNet captures how attribute interacts with object and model attribute-object transformation with \textbf{symmetry} and \textbf{group} principles. Thus, SymNet in the single-attribute setting directly outperforms the state-of-the-art on SUN. Besides, as seen in Fig.~\ref{Fig:attr-corr}, attribute labels on aPY~\cite{apy} are strongly correlated, but the single-attribute setting does not work well.
Thus, the multi-attribute setting \textbf{explicitly} mines attribute correlation and uses it to regularize the attribute-attribute relationship in attribute-object transformation. Other methods like FMT~\cite{FMT}, GALM~\cite{GALM} apply multi-tasking techniques to force the model to learn correlation automatically. Moreover, such an implicit mining approach leads to complex training, which is challenging to learn the correlation well.

Besides, in the single-attribute setting without both correlation-based losses (SymNet w/o $\mathcal{L}_{tri}^{sym}$ \& $\mathcal{L}_{tri}^{corr}$), the score is 82.1\% on aPY and 88.1\% on SUN, with a drop of \textbf{4.0}\% and \textbf{0.3}\% respectively. 
The reason is that, as revealed in Fig.~\ref{Fig:attr-corr}, attribute correlation in aPY is much stronger. Thus, its performance gap between single- and multi-attribute settings is more significant. However, the attribute correlation in SUN is much softer, so the impact of multi-attribute constraint is slight, and the ablations without correlation losses result in comparable performances in two settings.
These results show the effectiveness of our correlation-based loss and multi-attribute RMD and prove their general capability.

\subsection{Compositional Zero-Shot Learning}

\begin{table}[ht]
	\begin{center}
		\resizebox{0.34\textwidth}{!}{
		\begin{tabular}{lccccc}
		\hline  
	    Method & MIT-States & UT-Zappos\\
	    \hline 
	    Visual Product~\cite{redwine} & 14.7 & 24.9 \\
	    AttrOperator~\cite{operator} & 14.6 & 29.7 \\
	    GenModel~\cite{genmodel} & 15.1 & 18.4 \\
	    \hline
	    SymNet  &  \textbf{18.9} & \textbf{38.0}\\
	    \hline
		\end{tabular}}
	\end{center}
	\vspace{-0.3cm}
	\caption{Attribute learning results (accuracy, \%) on single-attribute benchmarks.}
	\label{tab:att}
	\vspace{-0.3cm}
\end{table}

\begin{table}[ht]
	\begin{center}
		\resizebox{0.21\textwidth}{!}
		{
		\begin{tabular}{lcc}
		\hline  
	    Method & aPY & SUN \\
	    \hline 
	    ALE\cite{ALE} & 69.2 & 74.5  \\
	    HAP\cite{HAP} & 58.2 & 76.7  \\
	    UDICA\cite{UDICA} & 82.3 & 85.8\\
	    KDICA\cite{UDICA}  & 84.7 & / \\
	    UMF\cite{UMF} & 79.7 & 80.5 \\
	    AMT\cite{AMT} & 84.5 & 82.5 \\
	    FMT\cite{FMT} & 70.5 & 75.5 \\
	    GALM\cite{GALM} & 84.2 & 86.5 \\
	    \hline
	    SymNet (single) & 82.1 & 88.1 \\
	    SymNet  &  \textbf{86.1} & \textbf{88.4} \\
	    \hline
		\end{tabular}}
	\end{center}
	\vspace{-0.3cm}
	\caption{Attribute learning results (mAU, \%) on multi-attribute benchmarks. SymNet (single) is the model without $L_{tri}^{sym}$ and $L_{tri}^{corr}$.}
	\label{tab:att-multi}
	\vspace{-0.5cm}
\end{table}

\begin{table}[t]
	\centering
	\small
	\adjustbox{width=0.95\linewidth}{
		\begin{tabular}{lccc|ccc}
			\toprule
			\multirow{2}{*}{Method}&  \multicolumn{3}{c}{MIT-States}& \multicolumn{3}{c}{UT-Zappos}\\
			& Top-1 & Top-2 & Top-3 & Top-1 & Top-2 & Top-3 \\ 
			\midrule
			Visual Product~\cite{redwine}  & 9.8/13.9$^*$ & 16.1 & 20.6 & 49.9$^*$ & / & / \\
			LabelEmbed (LE)~\cite{redwine} & 11.2/13.4$^*$& 17.6 & 22.4 & 25.8$^*$ & / & / \\
			~- LEOR~\cite{redwine}            & 4.5          & 6.2  & 11.8 &  /       & / & / \\
			~- LE + R~\cite{redwine}          & 9.3          & 16.3 & 20.8 &  /       & / & / \\
			~- LabelEmbed+~\cite{operator}    & 14.8*         &  /   &  /   & 37.4*& / & / \\
			\midrule
			AnalogousAttr~\cite{analogous}& 1.4          &  /   &  /   & 18.3  &  /  &  /  \\
			Red Wine~\cite{redwine}        & 13.1         & 21.2 & 27.6 & 40.3  &  /  &  /  \\ 
			AttrOperator~\cite{operator}    & 14.2         & 19.6 & 25.1 & 46.2  & 56.6 & 69.2 \\
			TAFE-Net~\cite{tafe}           & 16.4         & 26.4 & 33.0 & 33.2  &  /  &  /  \\
			GenModel~\cite{genmodel}       & 17.8         &  /   &  /   & 48.3  &  /  &  /  \\
			\midrule
			SymNet (Ours) & \textbf{19.9} & \textbf{28.2} & \textbf{33.8} & \textbf{52.1}  &\textbf{67.8} &  \textbf{76.0}\\ 
			\bottomrule
	\end{tabular}}
	\vspace{-0.2cm}
	\caption{\small CZSL results (top-k accuracy, \%) on MIT-States~\cite{mit} and UT-Zappos~\cite{ut}. The scores with $*$ mark are reproduced by~\cite{operator} and the rest are reported in the original papers.}
	\label{tab:mit-ut}
	\vspace{-0.2cm}
\end{table}

To evaluate the symmetry learning in the compositional zero-shot task, we conduct experiments on widely-used benchmarks: MIT-States~\cite{mit} and UT-Zappos~\cite{ut}.

The results of CZSL are shown in Tab.~\ref{tab:mit-ut}, where the first five rows are baselines from \cite{redwine, operator} (the scores with $*$ are reproduced by \cite{operator}, the others are from \cite{redwine}). 
SymNet outperforms all baselines on two benchmarks. 
Although we use a simple product to compose the attribute and object scores, we still achieve 2.1\% and 3.8\% improvements over the state-of-the-art~\cite{genmodel} on two benchmarks, respectively.
Most previous approaches do not surpass the Visual Product baseline on UT-Zappos, while ours outperforms by 2.2\%.

In addition, our object classification performance on the two datasets are 28.8\% and 65.4\% respectively, which is comparable to AttrOperator~\cite{operator} (20.5\%, 67.5\%) and GenModel~\cite{genmodel} (27.7\%, 68.1\%).
Accordingly, the main contribution of the CZSL improvement of SymNet comes from attribute learning rather than object recognition.

To further evaluate our SymNet, we additionally conduct the comparison on the \textit{generalized} CZSL setting from recent state-of-the-art TMN~\cite{tmn} on the larger MIT-States~\cite{mit} in Tab.~\ref{tab:gczsl-result-add}.
SymNet also outperforms previous methods significantly on all metrics.
We notice that novel metrics on UT-Zappos~\cite{ut} are proposed by a very recent work Causal~\cite{causal-czsl}, so we also conduct a test following its metrics. The results are shown in Tab.~\ref{tab:gczsl-result-ut}. Compared with TMN~\cite{tmn}, SymNet shows its superiority on Seen, Harmonic, and AUC metrics.
Furthermore, even compared with the very recent Causal~\cite{causal-czsl} method, our SymNet still improves the Seen, Closed, and AUC metrics. 
All these results further strongly prove the efficacy of our method. 

\begin{table}[t] 
	\centering
	\small
	\resizebox{0.48\textwidth}{!}{
		\begin{tabular}{l|ccc|ccc|ccc}
			\toprule
			\multirow{2}{*}{Method} & \multicolumn{3}{|c|}{Val AUC}& \multicolumn{3}{|c|}{Test AUC} & \multirow{2}{*}{Seen} & \multirow{2}{*}{Unseen} & \multirow{2}{*}{HM} \\
			& 1 & 2 & 3 & 1 & 2 & 3 & & & \\
			\midrule
			AttrOperator~\cite{operator}  & 2.5 & 6.2 & 10.1 & 1.6 & 4.7 & 7.6 & 14.3    & 17.4 & 9.9  \\
			Red Wine~\cite{redwine}      & 2.9 & 7.3 & 11.8 & 2.4 & 5.7 & 9.3 & 20.7    & 17.9 & 11.6 \\
			LabelEmbed+~\cite{operator}  & 3.0 & 7.6 & 12.2 & 2.0 & 5.6 & 9.4 & 15.0    & 20.1 & 10.7 \\
			f-CLSWGAN~\cite{featuregen} & 3.1 & 6.9 & 10.5 & 2.3 & 5.7 & 8.8 & 24.8    & 13.4 & 11.2 \\
			TMN~\cite{tmn}               & 3.5 & 8.1 & 12.4 & 2.9 & 7.1 & 11.5& 20.2    & 20.1 & 13.0 \\
			\midrule
			SymNet & \textbf{5.4} & \textbf{11.6} & \textbf{16.6} & \textbf{4.5} & \textbf{10.1} & \textbf{15.0} & \bf 26.2 & \bf 26.3 & \bf 16.8 \\
			\bottomrule
	\end{tabular}}
	\vspace{-0.2cm}
	\caption{\small Results of generalized CZSL on MIT-States~\cite{mit} following~\cite{tmn}. All methods use ResNet-18~\cite{resnet} as backbone.}
	\label{tab:gczsl-result-add}
	\vspace{-0.5cm}
\end{table}

\begin{table}[h]
	\centering
	\small
	\resizebox{0.45\textwidth}{!}{
		\begin{tabular}{l|ccc|ccc|ccc}
			\toprule
			Model & Unseen & Seen & Harmonic & Closed & AUC \\
			\midrule
			LabelEmbed+~\cite{operator}   & 16.2 & 53.0 & 24.7 & 59.3 & 22.9 \\
			AttrOperator~\cite{operator}   & 25.5 & 37.9 & 27.9 & 54.0 & 22.1 \\
			TMN~\cite{tmn}                & 10.3 & 54.3 & 17.4 & \textbf{62.0} & 25.4 \\
			Causal~\cite{causal-czsl}     & \textbf{28.0} & 37.0 & \textbf{30.6} & 58.6 & 26.4  \\
			\midrule
			SymNet  & 10.3 & \textbf{56.3} & 24.1 & 58.7 & \textbf{26.8} \\
			\bottomrule
	\end{tabular}}
	\vspace{-0.2cm}
	\caption{\small Results of generalized CZSL on UT-Zappos~\cite{ut} following~\cite{causal-czsl}. All methods use ResNet-18~\cite{resnet} as backbone. The results of LabelEmbed+, AttrOperator, and TMN are from~\cite{causal-czsl}.}
	\label{tab:gczsl-result-ut}
	\vspace{-0.3cm}
\end{table}

\begin{table}[h]
	\centering
	\small
	\adjustbox{width=0.98\linewidth}{
		\begin{tabular}{l|ccc|ccc}
			\toprule
			\multirow{2}{*}{Method} &  \multicolumn{3}{c}{Novel}& \multicolumn{3}{c}{All} \\
	        & 1-shot & 2-shot & 5-shot & 1-shot & 2-shot & 5-shot \\
			\midrule
			COMP\cite{comp}    & 52.5 & 63.6 & 73.8 & 62.6 & 68.4 & 74.0 \\ 	
			COMP - SymNet  & \textbf{54.0} &  \textbf{63.8} & \textbf{74.3} & \textbf{63.1} & \textbf{68.9} &  \textbf{74.1} \\
            \midrule
			COMP w/ data aug ~\cite{comp}    & 53.6 & 64.8 & 74.6 & 63.1 & {\bf 69.2} & 74.5 \\ 	
			COMP w/ data aug - SymNet    & {\bf 57.3} & {\bf 66.7} & {\bf 76.0} & {\bf 64.9} & 68.8 & {\bf 74.9} \\ 		
			\bottomrule
	\end{tabular}}
	\vspace{-0.2cm}
	\caption{\small Results of few-shot recognition on CUB-200-2011\cite{cub}.}
	\label{tab:fewshot-cub}
	\vspace{-0.5cm}
\end{table}

\subsection{Application in Few-Shot Learning}
SymNet is a generic feature extractor, and its features are rich in attribute semantics which can strengthen the object representations. 
As a verification, we conduct a few-shot recognition experiment on CUB-200-2011~\cite{cub} following the protocol of COMP~\cite{comp}. We enhance the attribute representation for the downstream few-shot classification by concatenating the average CoN and DecoN transformed features over all attributes to the original ResNet feature, respectively. 
Setting details and supplementary results on SUN397~\cite{sun} are listed in Suppl Sec.~4.4.

We embed SymNet on the pre-trained ResNet-10 backbone and evaluate on both settings (whether use data augmentation during classifier training) following COMP~\cite{comp}. As shown in Table~\ref{tab:fewshot-cub}, the SymNet features bring stable performance gain to the COMP~\cite{comp} baselines on both novel and all categories in most situations, indicating that feature from SymNet attribute transformation is an effective supplementation of object representation. In addition, it also proves the generalization ability of SymNet cross object categories.

\subsection{Visualization}
\begin{figure*}[!ht]
    \begin{center}
        \begin{minipage}{.43\linewidth}
            \centerline{\includegraphics[width=\linewidth]{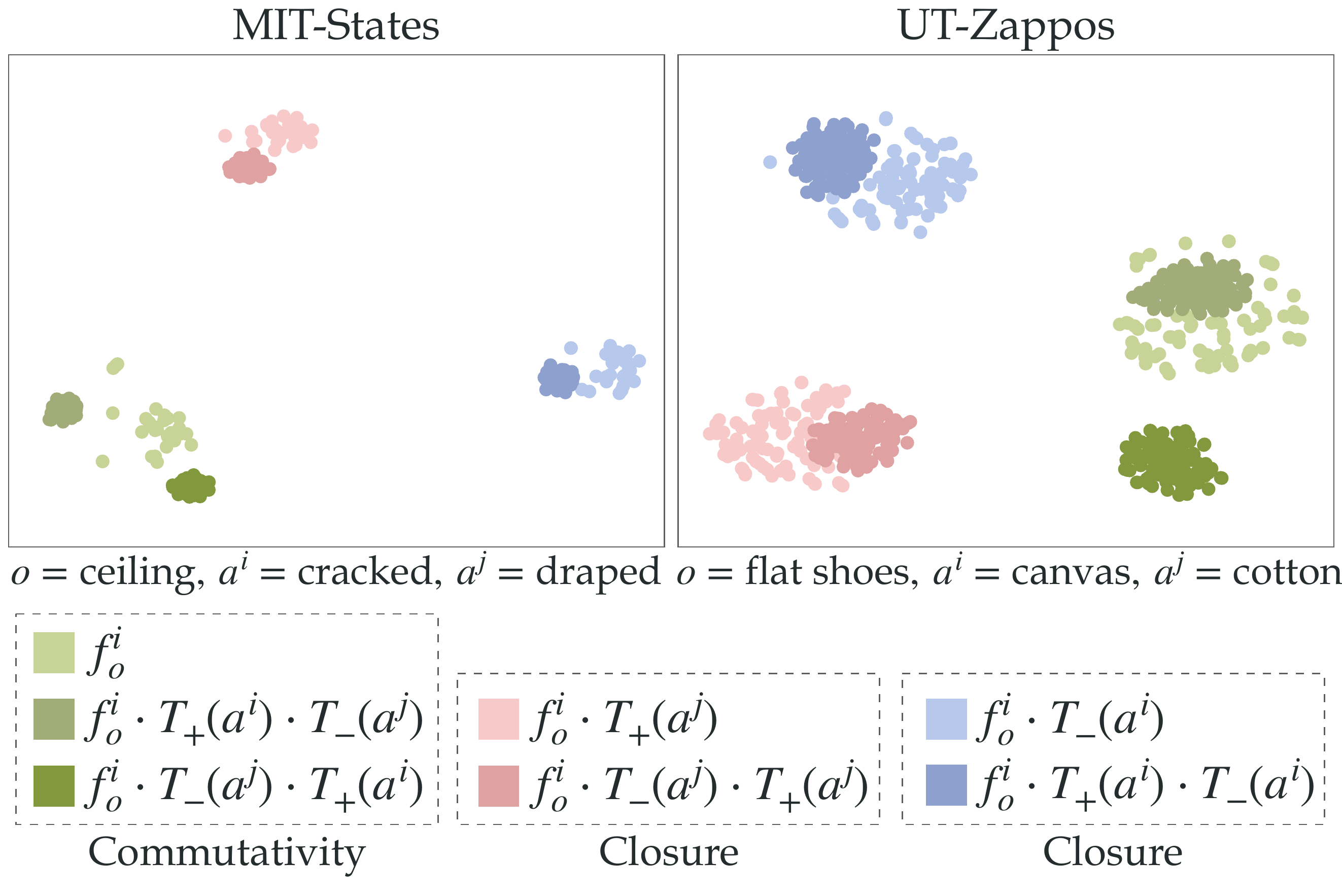}}
            \centerline{\small (a) Closure and Commutativity}
        \end{minipage}
        \hfill
        \begin{minipage}{.43\linewidth}
            \centerline{\includegraphics[width=\linewidth]{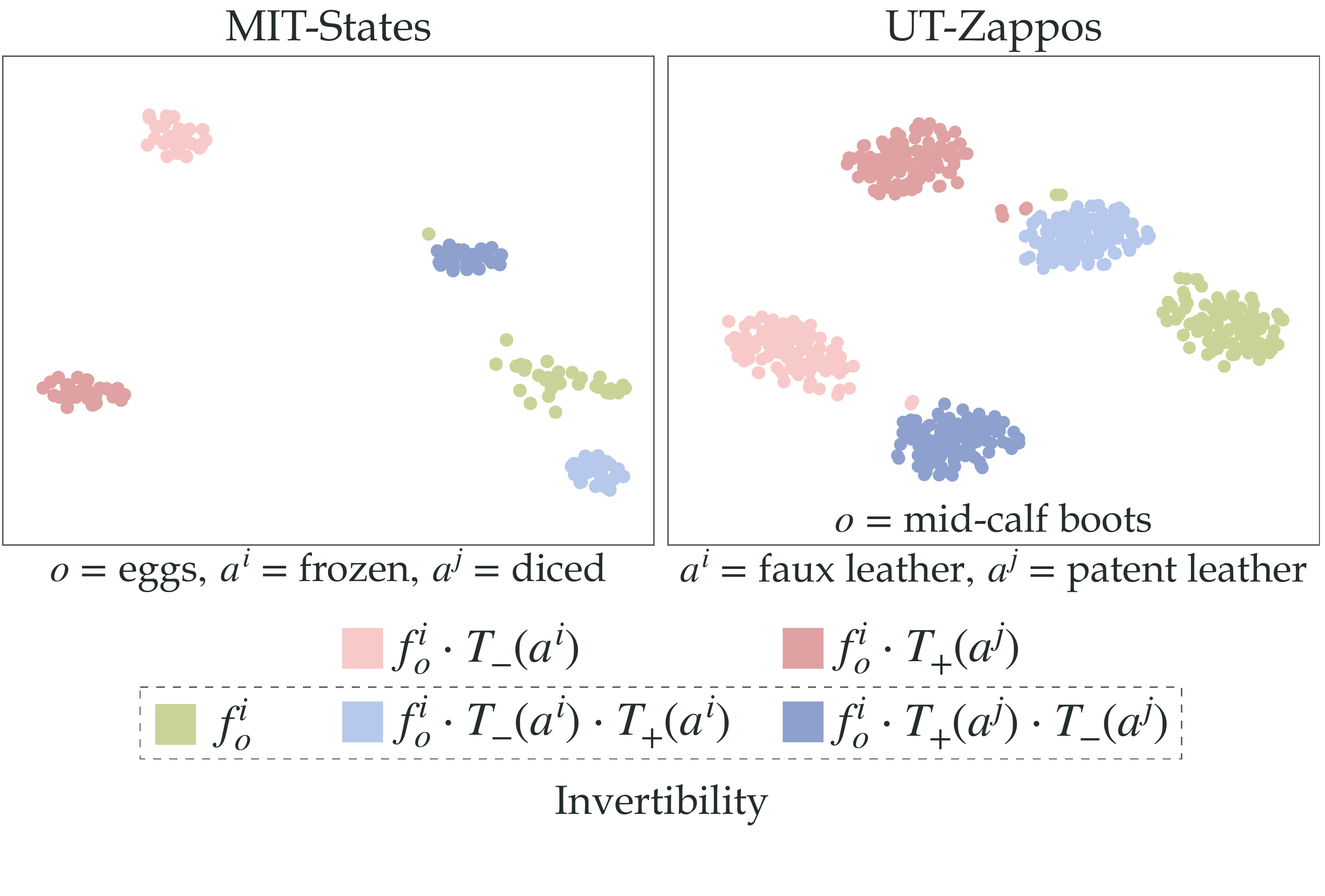}}    
            \centerline{\small (b) Invertibility}
        \end{minipage}
        \vfill
        \begin{minipage}{.43\linewidth}
            \centerline{\includegraphics[width=\linewidth]{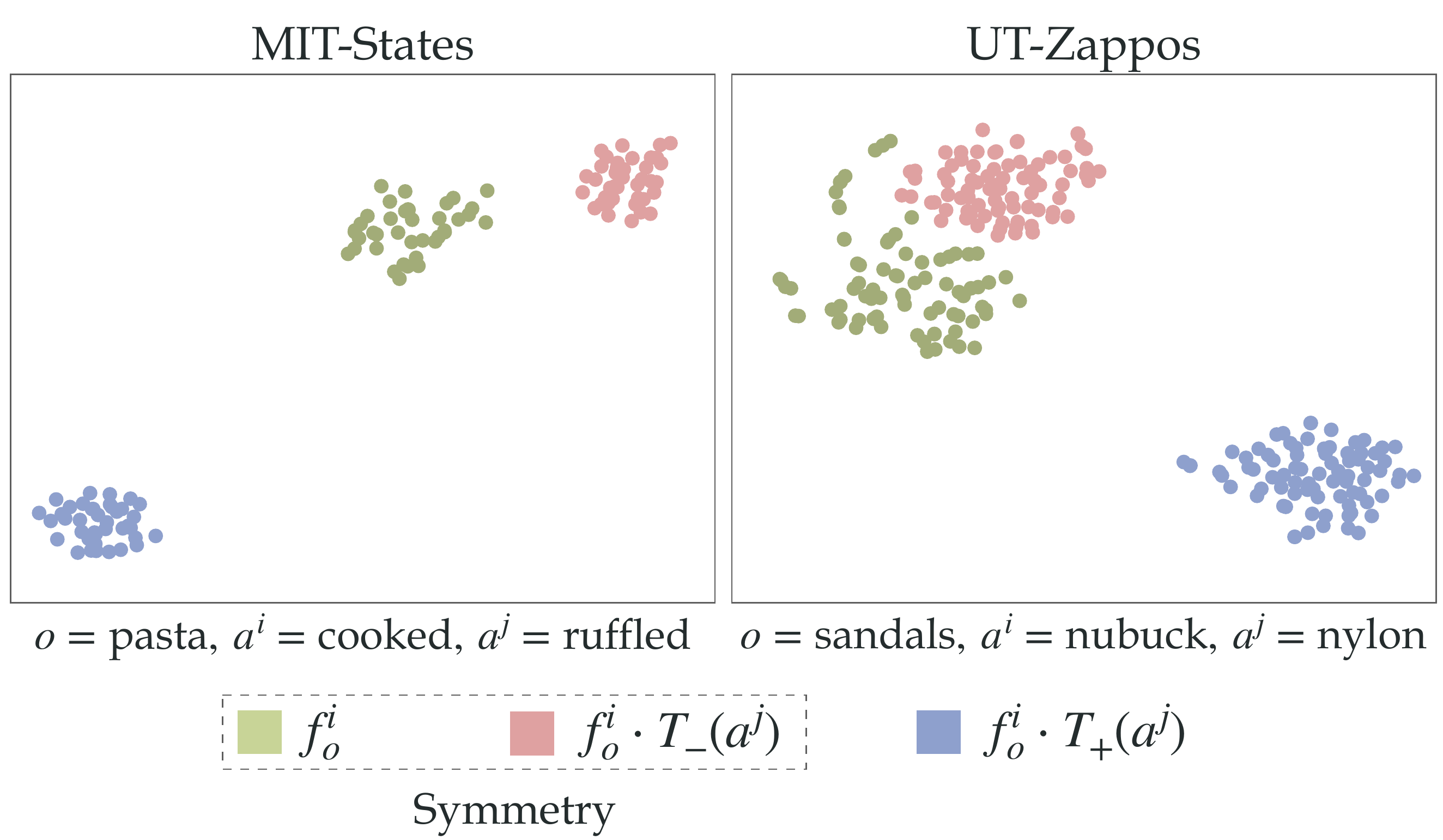}}
            \centerline{\small (c) Symmetry-1}
        \end{minipage}
        \hfill
        \begin{minipage}{.43\linewidth}
            \centerline{\includegraphics[width=\linewidth]{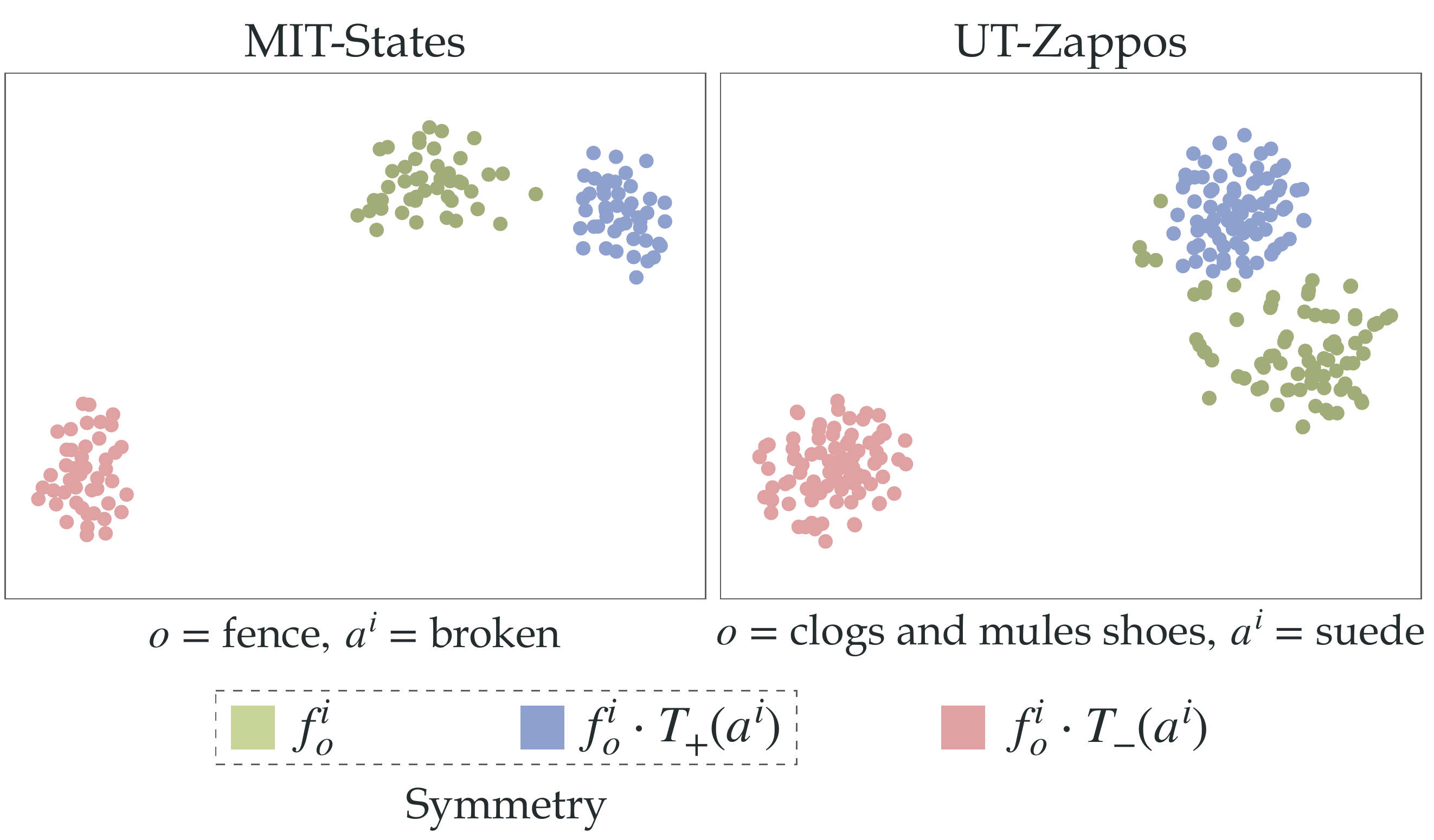}}
            \centerline{\small (d) Symmetry-2}
        \end{minipage}
        \vfill
        \begin{minipage}{.43\linewidth}
            \centerline{\includegraphics[width=\linewidth]{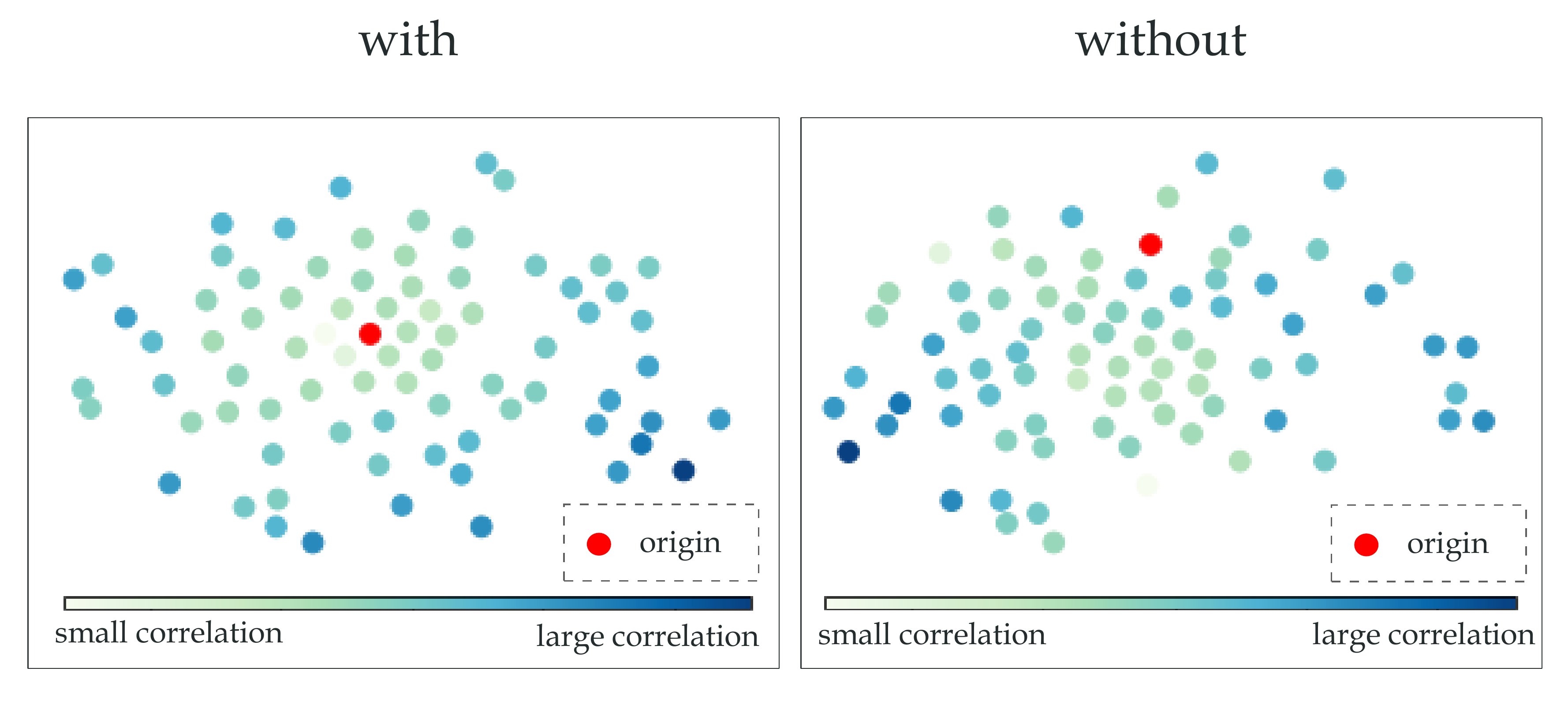}}
            \centerline{\small (e) Multi-attribute Symmetry (SUN)}
        \end{minipage}
        \hfill
        \begin{minipage}{.43\linewidth}
            \centerline{\includegraphics[width=\linewidth]{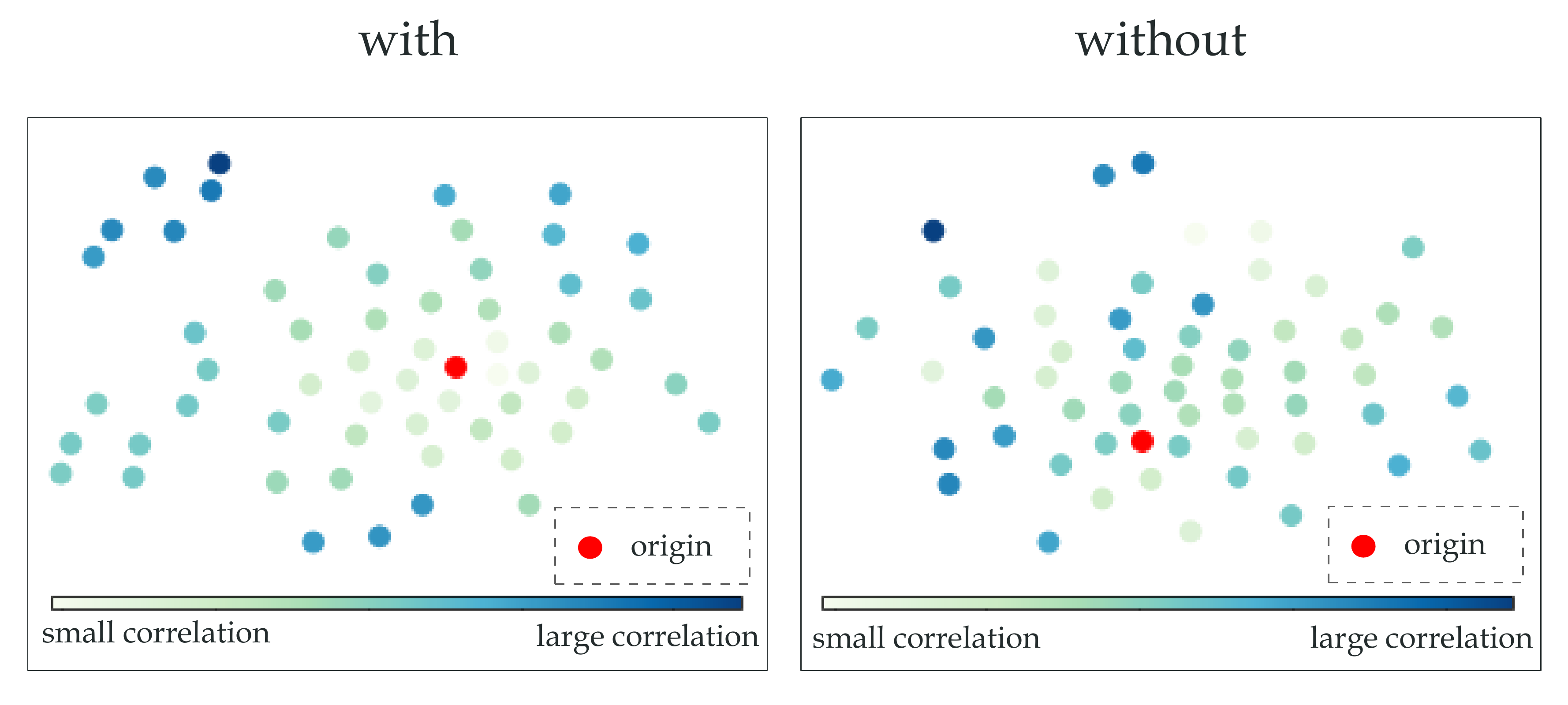}}
            \centerline{\small (f) Multi-attribute Symmetry (aPY)}
        \end{minipage}
        \vfill
        \begin{minipage}{.43\linewidth}
            \centerline{\includegraphics[width=\linewidth]{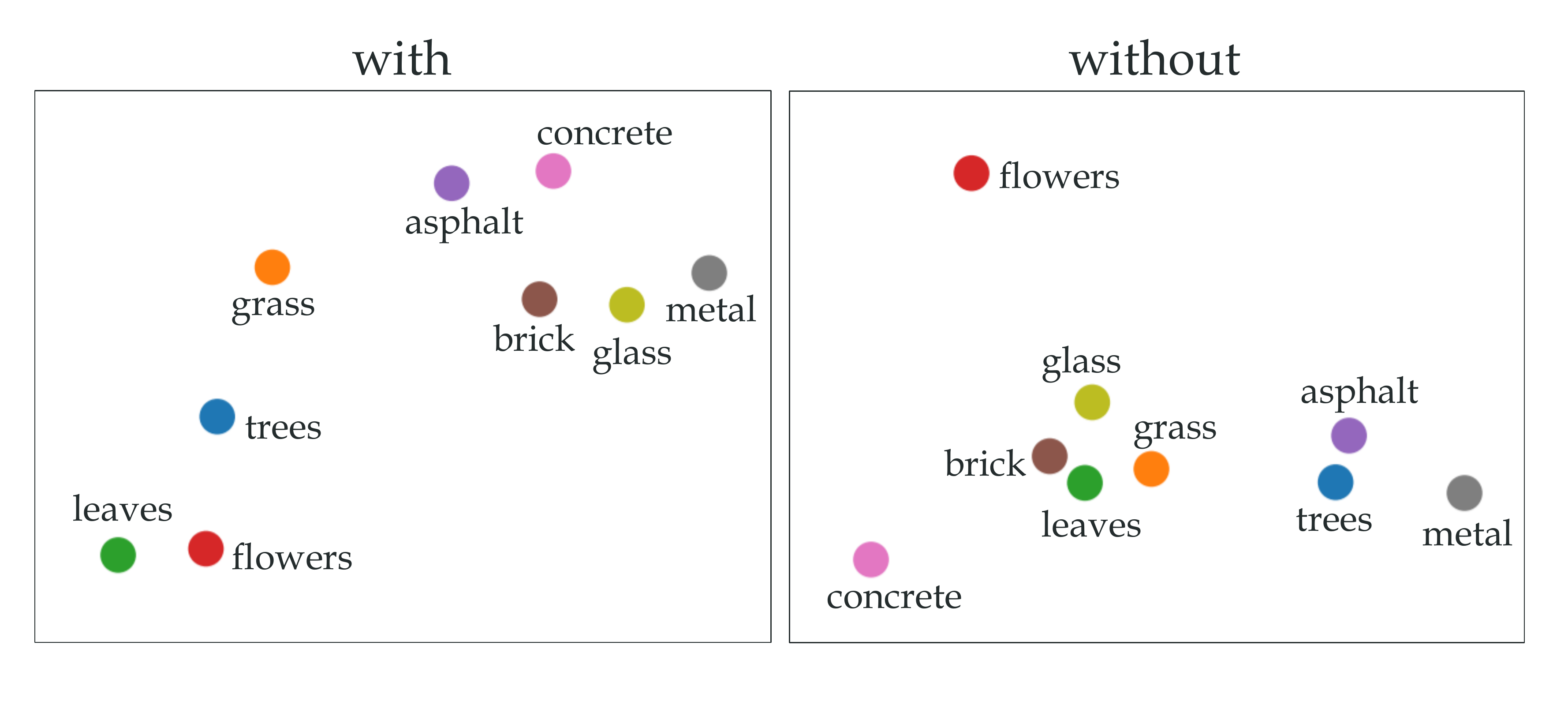}}
            \centerline{\small (g) Attention constraints (SUN)}
        \end{minipage}
        \hfill
        \begin{minipage}{.43\linewidth}
            \centerline{\includegraphics[width=\linewidth]{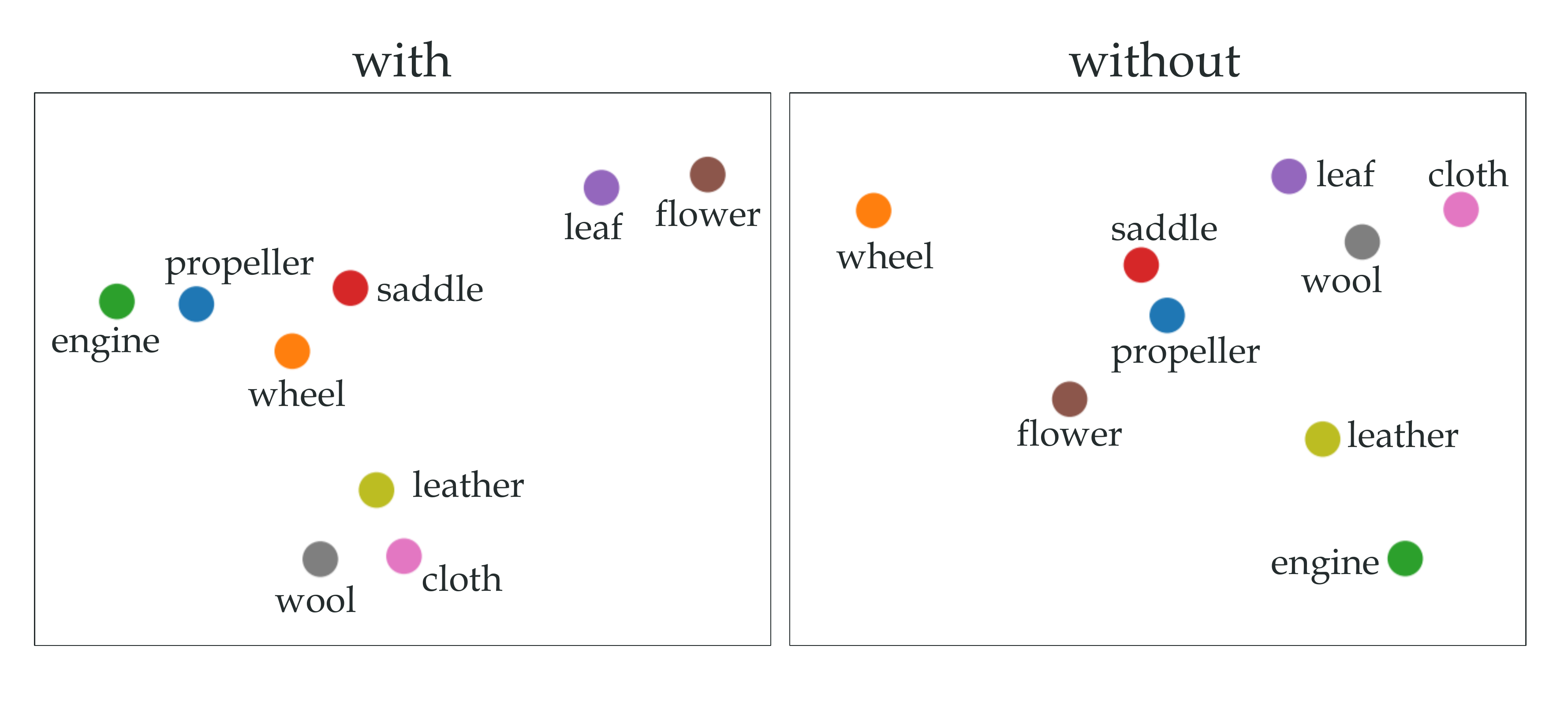}}
            \centerline{\small (h) Attention constraints (aPY)}
        \end{minipage} 
	\end{center}
	\vspace{-0.2cm}
	\caption{Visualization of symmetry, group axioms and attention by t-SNE~\cite{tsne}. In (a)-(d), points with colors in the same dotted box should be close according to the corresponding learning principles. 
	Especially, (e) and (f) shows the RMD property in the multi-attribute setting where the red dot is the original object embedding and the other dots are the embeddings after attribute removals. For example, if attribute $a^i$ has a larger correlation to the existing attributes of this object, as $corr(a^i,\mathcal{X})$ is large, then the corresponding dot is in a darker color and has a larger distance after removal. 
	On the contrary, the dot of removing attribute $a^j$ with smaller $corr(a^j,\mathcal{X})$ is represented in a lighter color and closer to the origin. However, without the $L^{sym}_{trip}$, the distribution of attribute removal is much noisier. 
	(g) and (f) are the distributions of attention representation. Attentions of more correlated attributes are closer in latent space, e.g., \texttt{leaves} and \texttt{flowers}, but this property cannot keep without $L_{tri}^{corr}$.}
	\label{Figure:tsne} 
	\vspace{-0.5cm}
\end{figure*}

To verify the robustness and principles in transformations, we use t-SNE~\cite{tsne} to visualize the attention vectors or image embeddings before or after transformations in Fig.~\ref{Figure:tsne}.

Precisely, we first visualize the group axioms related transformations:
1) \textbf{Closure} is verified by comparing \{$f_o^{i} \cdot T_+(a^i) \cdot T_-(a^i)$ v.s. $f_o^{i} \cdot T_-(a^i)$\} and \{$f_o^{i} \cdot T_-(a^j) \cdot T_+(a^j)$ v.s.  $f_o^{i} \cdot T_+(a^j)$\}.
2) \textbf{Invertibility} is verified by comparing \{$f_o^{i} \cdot T_+(a^j) \cdot T_-(a^j)$ v.s. $f_o^{i} \cdot T_e$\} and \{$f_o^{i} \cdot T_-(a^i) \cdot T_+(a^i)$ v.s. $f_o^{i} \cdot T_e$\}.
3) \textbf{Commutativity} is verified by comparing \{$f_o^{i} \cdot T_+(a^i) \cdot T_-(a^j)$ v.s. $f_o^{i} \cdot T_-(a^j) \cdot T_+(a^i)$\}. The results are shown in Fig.~\ref{Figure:tsne} (a,b).
We observe that SymNet can robustly operate the transformations and the axiom objectives are well satisfied during embedding transformations.

Then, to verify the \textbf{symmetry} property, we visualize the sample embeddings in relative moving space in Fig.~\ref{Figure:tsne}(c,d):
1) For the sample $f_o^i$ which do not have attribute $a^j$, $f_o^i \cdot T_+(a^j)$ should be far from $f_o^i$. On the contrary, $f_o^i \cdot T_-(a^j)$ are relatively close to $f_o^i$ because of the symmetry principle.
2) For the sample $f_o^i$ with attribute $a^i$, $f_o^i \cdot T_+(a^i)$  should be close to $f_o^i$ and $f_o^i \cdot T_-(a^i)$ should be far from $f_o^i$.
We can also find that the relative moving distance rules are satisfied, i.e., the symmetry is well learned by our SymNet.

We also verify the properties in multi-attribute scenario in Fig.~\ref{Figure:tsne}(e,f,g,h): 
1) \textbf{Multi-Attribute Symmetry Constraint}: after removing an attribute which has \textit{larger} positive correlation to an object, the moving distance would also be \textit{larger}.
In Fig.~\ref{Figure:tsne}(e,f), the red dots are the positions of the original object embeddings. The others are the embeddings after the attribute removals. The dot color is related to the correlation value. Darker color denotes the removal of a \textit{more correlated} attribute. We find that \textit{larger} correlations would cause \textit{larger} moving distances. This phenomenon depicts that the transformations are more principled and consistent with the practice with our multi-attribute symmetry constraint.
2) \textbf{Attention Correlation Constraint}: the attention vectors of \textit{more correlated} attributes should be \textit{closer}. 
In Fig.~\ref{Figure:tsne}(g,h), this constraint indeed leads to better clusters.
We can observe that our multi-attribute model can capture the attribute correlations and well-learn these two constraints.
Besides, we report the \textbf{image retrieval} via SymNet in Suppl Sec.~3.

\begin{table}[t]
	\centering
	\small
	\adjustbox{width=\linewidth}{
		\begin{tabular}{l|ccc|ccc|c|c}
			\toprule
			\multirow{2}{*}{Method} &  \multicolumn{3}{c}{MIT-States}& \multicolumn{3}{c|}{UT-Zappos} & \multirow{2}{*}{aPY} & \multirow{2}{*}{SUN}\\
			& Top-1 & Top-2 & Top-3 & Top-1 & Top-2 & Top-3 && \\
			\midrule
			SymNet & \textbf{19.9} & \textbf{28.2} & \textbf{33.8} & \textbf{52.1}  &\textbf{67.8} &  \textbf{76.0} & \textbf{86.1} & \textbf{88.4}\\ 
			\midrule
			SymNet w/o $\mathcal{L}_{sym}$   & 18.3 & 27.5 & 33.4 & 51.1 & 67.0 & 76.0 & 65.7 & 84.0\\
			SymNet w/o $\mathcal{L}_{axiom}$ & 16.9 & 25.5 & 30.9 & 47.6 & 65.4 & 73.6 & 83.6 & 87.9\\
			SymNet w/o $\mathcal{L}_{inv}$   & 17.9 & 26.7 & 32.5 & 50.8 & 67.4 & 76.1 & 84.9 & 88.1\\
			SymNet w/o $\mathcal{L}_{com}$   & 17.8 & 27.0 & 32.7 & 51.2 & 67.6 & 75.8 & / & /\\
			SymNet w/o $\mathcal{L}_{clo}$   & 18.0 & 27.0 & 32.8 & 51.1 & 67.2 & 76.0 & 84.1 & 88.1\\
			SymNet w/o $\mathcal{L}_{cls}$   & 10.3 & 18.9 & 25.9 & 28.7 & 51.2 & 65.2 & 81.3 & 78.0\\ 
			SymNet w/o $\mathcal{L}_{tri}$   & 17.8 & 26.8 & 32.6 & 49.2 & 65.3 & 74.2 & 83.5 & 88.0\\
			SymNet w/o $\mathcal{L}_{tri}^{sym}$   & / & / & / & / & / & / & 84.0 & 88.1\\
			SymNet w/o $\mathcal{L}_{tri}^{corr}$   & / & / & / & / & / & / & 84.5 & 88.1\\
			SymNet w/o $\mathcal{L}_{tri}^{sym}$ \& $\mathcal{L}_{tri}^{corr}$   & / & / & / & / & / & / & 82.1 & 88.1\\
			SymNet w/o $\mathcal{L}_{sym}$ \& $\mathcal{L}_{tri}$ & 17.7 & 27.0 & 33.0 & 50.1 & 66.1 & 75.6 & 63.1 & 85.3\\
			SymNet w/o $\mathcal{L}_{tri}$ \& $\mathcal{L}_{cls}$ & 10.5 & 19.4 & 26.7 & 28.6 & 51.4 & 65.6 & 80.1 & 70.9\\
			SymNet w/o $\mathcal{L}_{sym}$ \& $\mathcal{L}_{cls}$ & 9.3 & 17.0 & 22.7 & 27.4 & 48.2 & 64.1 & 66.4 & 78.0\\
			SymNet only $\mathcal{L}_{sym}$ & 9.4 & 16.9 & 22.5 & 20.4 & 38.9 & 53.5 & 73.4 & 71.0\\
			SymNet only $\mathcal{L}_{cls},\mathcal{L}_{tri}$ & 17.1 & 26.1 & 31.7 & 48.5 & 65.7 & 73.9 & 71.6 & 84.9\\
			\midrule
			SymNet w/o attention & 18.0 & 26.9 & 32.7 & 48.5 & 65.0 & 75.6 & 84.8 & 88.2\\
			SymNet $\tanh$ attention & 16.9 & 25.0 & 30.8 & 42.0 & 59.0 & 69.0 & 84.1 & 88.0 \\
			\midrule
			SymNet $L_1$ dist. & 7.1 & 11.2 & 14.3 & 37.5 & 53.3 & 62.3 & 82.0 & 87.6\\
			SymNet $Cos$ dist. & 11.3 & 20.7 & 28.5 & 18.7 & 41.1 & 60.0 & 82.6 & 86.6\\
			\bottomrule
	\end{tabular}}
	\vspace{-0.2cm}
	\caption{\small Ablation studies on CZSL and multi-attribute learning.}
	\label{tab:abl}
    \vspace{-0.6cm}
\end{table}

\subsection{Ablation Study}
\label{sec:ablation}
To evaluate the components of our model, we compare the results of different model designs in Tab.~\ref{tab:abl} following the settings of \cite{operator}. 
We also conduct experiments of the generalized CZSL setting~\cite{tmn}, COMP~\cite{comp}, and AttrOperator~\cite{operator}, which are attached in supplementary material.

(1) \textbf{Objectives}: to evaluate the objectives constructed from group axioms, the core principle symmetry, and attribute correlation in the multi-attribute scenario, we conduct experiments of removing objectives. 
In Tab.~\ref{tab:abl} and Suppl Tab.~2, SymNet shows obvious degradations without the constraints of these principles ($\mathcal{L}_{sym}, \mathcal{L}_{axiom}, \mathcal{L}_{cls}, \mathcal{L}_{tri}$). The degradations are in line with our assumption that a transformation framework that covers the essential principles can largely promote attribute learning.

\textbf{1-a)} Specifically, removing $\mathcal{L}_{cls}$ leads to a more significant performance drop than other losses. Because $\mathcal{L}_{cls}$ applied on the input/output \textbf{object embeddings} of CoN and DecoN can establish the basic distribution of object embeddings, thus keep the basic semantics of the object and attribute. While the other transformation-related losses (e.g., $\mathcal{L}_{axiom}, \mathcal{L}_{sym}, \mathcal{L}_{tri}$) ensure the attribute transformation rational and the object embedding moving to follow the theoretical guidance, thus can afford the robust classification via RMD.
Removing $\mathcal{L}_{cls}$ would destroy the basic semantic information within object embeddings (e.g., the relationship between attributes and objects, or the inter-attribute correlation), therefore, results in a larger performance drop.
Similar phenomenon also presents in AttrOperator~\cite{operator}, that removing $L_{aux}$ (counterpart of our $\mathcal{L}_{cls}$, ensuring that the identity of the attribute/object is not lost during composition), leads to about relative \textbf{55}\% drop of h-mean.

\textbf{1-b)} We conduct an ablation of \textbf{only keeping $\mathcal{L}_{cls}$ and $\mathcal{L}_{tri}$}. The top-1 accuracy drops 2.8\%, 3.6\%, 14.5\%, and 3.5\% on four datasets in CZSL and the top-1 AUC drops 1.7 on MIT-States in generalized CZSL. 
This performance gap can then be filled by our proposed $\mathcal{L}_{axiom}$ and $\mathcal{L}_{sym}$. 

\textbf{1-c)} Though the overall score drop of ``w/o  $\mathcal{L}_{cls}$'' is larger than the transformation-related losses, the situations of specific attribute classes differ.
In detail, we compare our best model to the model trained with only $\mathcal{L}_{cls}\&\mathcal{L}_{tri}$. 
From the results, we find that the symmetry and axiom constraints can facilitate the learning of \textbf{few-shot attributes}.
Our full model outperforms the one with only $\mathcal{L}_{cls}\&\mathcal{L}_{tri}$ on fewer-shot attributes, such as \texttt{tight, bent, viscous}. The accuracy of samples of the thirty least frequent attributes increases by \textbf{4.3}\% with $\mathcal{L}_{sym}\&\mathcal{L}_{axiom}$.
While on the rest attributes with more samples, the full model only has 0.9\% accuracy gain compared with the model with only $\mathcal{L}_{cls}\&\mathcal{L}_{tri}$. 
The reason may be that the RMD relying on symmetry and dynamic embedding moving ($\mathcal{L}_{sym}\&\mathcal{L}_{axiom}$) is less reliant on the training sample scale than canonical classification.

\textbf{1-d)} Noticeably, the difference of \textit{datasets and metrics} also exert an influence. That said, the effects of losses vary on different datasets depending on the data scale, quality, and distribution. 
For example, comparing the single- and multi-attribute settings of SymNet, there is a significant performance drop on aPY but a smaller one on SUN without the multi-attribute (correlation) constraint. 
The reason is the different levels of correlations in aPY and SUN, where aPY is much stronger, as revealed in Fig.~\ref{Fig:attr-corr}.
It is also noticed that training a SymNet without $\mathcal{L}_{sym}$, or $\mathcal{L}_{sym}$\&$\mathcal{L}_{tri}$ leads to a significant degradation for aPY, but a relatively minor drop for the other three datasets.

\textbf{1-e)} Besides, in training, the \textit{utilization order of losses} also makes a difference.
In generalized CZSL, if the model is first trained with $\mathcal{L}_{cls}\&\mathcal{L}_{tri}$ and then finetuned with all four losses, the performance is very close to our best model (Top-1 AUC drops from 5.40 to 5.35). 
However, if the model is first trained with $\mathcal{L}_{axiom}\&\mathcal{L}_{sym}$ and then finetuned with all losses, the Top-1 AUC considerably drops from 5.4 to 4.8 (relatively 11.1\%). 
This phenomenon accords with the above analysis that $\mathcal{L}_{cls},\mathcal{L}_{tri}$ keeps the basic semantics of embeddings and $\mathcal{L}_{axiom},\mathcal{L}_{sym}$ further guarantee the rationality of attribute transformations.
The subsequent transformations cannot be well learned without the well pre-positioned object embeddings and semantic relationships.

(2) \textbf{Attention}: we conduct an ablation study on the attention module by removing the attention and only keep the MLPs. The removal will degrade results on all benchmarks.
The model can learn the positions to modify on the object embeddings when operating attribute transformations with attention.
We also evaluate different type of attention designs, i.e., using activation function $\tanh(\cdot)$ to convert the attentions into range $(-1,1)$, but it performs worse than the Sigmoid function. The reason maybe the range of activation function (Sigmoid is $(0,1)$) and the results show Sigmoid activation is more suitable for the training.

(3) \textbf{Distance Metrics}: SymNet with other distance metrics including $L_1$ and cosine distances are evaluated, i.e., replacing all the distance computation in losses and RMD. They all perform much worse than $L_2$.
With cosine distance, the accuracy severely drops since cosine distance only measures the angle between embeddings and may not be enough for complex attribute transformations.
Moreover, training SymNet with $L_1$ is more difficult to converge, thus performing poorly. With the same hyper-parameters, the model with $L_1$ converges at 4,000 epochs, much slower than the model with $L_2$ (320 epochs). There are more spikes on the $L_1$ loss curve, indicating its training instability.

For more please refer to the supplementary: comparisons with AttrOperator~\cite{operator} (Suppl Sec.~4.2) and COMP~\cite{comp} (Suppl Sec.~4.3), application details (Suppl Sec.~4.4), the relationship between dataset and performance (Suppl Sec.~5).

\section{Conclusion}
In this work, we study the symmetry property of the attribute-object compositions, which reveals profound principles in composition transformations. 
To an object, if we add an attribute that it already has, or erase one it does not have, it would keep.
We construct a framework inspired by group theory to couple and decouple attribute-object compositions to learn symmetry and use group axioms and symmetry as the learning objectives.
Moreover, we explore the attribute correlation to improve attribute recognition with the extended learning objectives with multi-attribute constraints for a multi-attribute scenario.
On attribute learning and CZSL tasks, our method achieves state-of-the-art performances.
In the future, we consider to study the transformation with varying degrees, e.g., \texttt{not-}, \texttt{half-}, and \texttt{totally-peeled} and apply SymNet to GAN.

\section*{Acknowledgment}

This work is supported in part by the National Key R\&D Program of China, No.2017YFA0700800, National Natural Science Foundation of China under Grants 61772332, and Baidu Scholarship.

\ifCLASSOPTIONcaptionsoff
  \newpage
\fi



%


%

\bibliographystyle{IEEEtran}
\bibliography{egbib}

\begin{thebibliography}{10}\itemsep=-1pt

\bibitem{redwine}
Ishan Misra, Abhinav Gupta, and Martial Hebert.
\newblock From red wine to red tomato: Composition with context.
\newblock In {\em CVPR}, 2017.

\bibitem{operator}
Tushar Nagarajan and Kristen Grauman.
\newblock Attributes as operators: factorizing unseen attribute-object compositions.
\newblock In {\em ECCV}, 2018.

\bibitem{comp}
Pavel Tokmakov, Yu-Xiong Wang, and Martial Hebert.
\newblock Learning compositional representations for few-shot recognition.
\newblock In {\em ICCV}, 2019.

\bibitem{genmodel}
Zhixiong Nan, Yang Liu, Nanning Zheng, and Song-Chun Zhu.
\newblock Recognizing unseen attribute-object pair with generative model.
\newblock In {\em AAAI}, 2019.

\bibitem{tafe}
Xin Wang, Fisher Yu, Ruth Wang, Trevor Darrell, and Joseph~E Gonzalez.
\newblock Tafe-net: Task-aware feature embeddings for low shot learning.
\newblock In {\em CVPR}, 2019.

\bibitem{word2vec}
Tomas Mikolov, Kai Chen, Greg Corrado, and Jeffrey Dean.
\newblock Efficient estimation of word representations in vector space.
\newblock In {\em arXiv preprint arXiv:1301.3781}, 2013.

\bibitem{tsne}
Laurens van~der Maaten and Geoffrey Hinton.
\newblock Visualizing data using t-sne.
\newblock In {\em JMLR}, 2008.

\bibitem{faster}
Shaoqing Ren, Kaiming He, Ross Girshick, and Jian Sun.
\newblock Faster r-cnn: Towards real-time object detection with region proposal networks.
\newblock In {\em NIPS}, 2015.

\bibitem{li2017scene}
Yikang Li, Wanli Ouyang, Bolei Zhou, Kun Wang, and Xiaogang Wang.
\newblock Scene graph generation from objects, phrases and region captions.
\newblock In {\em ICCV}, 2017.

\bibitem{mit}
Phillip Isola, Joseph~J Lim, and Edward~H Adelson.
\newblock Discovering states and transformations in image collections.
\newblock In {\em CVPR}, 2015.

\bibitem{ut}
Aron Yu and Kristen Grauman.
\newblock Semantic jitter: Dense supervision for visual comparisons via synthetic images.
\newblock In {\em ICCV}, 2017.

\bibitem{sun}
Jianxiong Xiao, James Hays, Krista~A Ehinger, Aude Oliva, and Antonio Torralba.
\newblock Sun database: Large-scale scene recognition from abbey to zoo.
\newblock In {\em CVPR}, 2010.

\bibitem{awa1}
Christoph~H Lampert, Hannes Nickisch, and Stefan Harmeling.
\newblock Learning to detect unseen object classes by between-class attribute transfer.
\newblock In {\em CVPR}, 2009.

\bibitem{awa2}
Yongqin Xian, Christoph~H Lampert, Bernt Schiele, and Zeynep Akata.
\newblock Zero-shot learning-a comprehensive evaluation of the good, the bad and the ugly.
\newblock In {\em TPAMI}, 2018.

\bibitem{apy}
Ali Farhadi, Ian Endres, Derek Hoiem, and David Forsyth.
\newblock Describing objects by their attributes.
\newblock In {\em CVPR}, 2009.

\bibitem{glove}
Jeffrey Pennington, Richard Socher, and Christopher Manning.
\newblock Glove: Global vectors for word representation.
\newblock In {\em EMNLP}, 2014.

\bibitem{resnet}
Kaiming He, Xiangyu Zhang, Shaoqing Ren, and Jian Sun.
\newblock Deep residual learning for image recognition.
\newblock In {\em CVPR}, 2016.

\bibitem{imagenet}
Jia Deng, Wei Dong, Richard Socher, Li-Jia Li, Kai Li, and Li~Fei-Fei.
\newblock Imagenet: A large-scale hierarchical image database.
\newblock In {\em CVPR}, 2009.

\bibitem{relativeattr}
Devi Parikh and Kristen Grauman.
\newblock Relative attributes.
\newblock In {\em ICCV}, 2011.

\bibitem{PETA}
Yubin Deng, Ping Luo, Chen~Change Loy, and Xiaoou Tang.
\newblock Pedestrian attribute recognition at far distance.
\newblock In {\em ACMMM}, 2014.

\bibitem{poselet}
Lubomir Bourdev, Subhransu Maji, and Jitendra Malik.
\newblock Describing people: A poselet-based approach to attribute classification.
\newblock In {\em ICCV}, 2011.

\bibitem{celebA}
Ziwei Liu, Ping Luo, Xiaogang Wang, and Xiaoou Tang.
\newblock Deep learning face attributes in the wild.
\newblock In {\em ICCV}, 2015.

\bibitem{ucf101}
Khurram Soomro, Amir~Roshan Zamir, and Mubarak Shah.
\newblock Ucf101: A dataset of 101 human actions classes from videos in the wild.
\newblock In {\em arXiv preprint arXiv:1212.0402}, 2012.

\bibitem{cocoattr}
Genevieve Patterson and James Hays.
\newblock Coco attributes: Attributes for people, animals, and objects.
\newblock In {\em ECCV}, 2016.

\bibitem{analogous}
Chao-Yeh Chen and Kristen Grauman.
\newblock Inferring analogous attributes.
\newblock In {\em CVPR}, 2014.

\bibitem{visualgenome}
Ranjay Krishna, Yuke Zhu, Oliver Groth, Justin Johnson, Kenji Hata, Joshua Kravitz, Stephanie Chen, Yannis Kalantidis, Li-Jia Li, David~A Shamma, Michael Bernstein, and Li~Fei-Fei.
\newblock Visual genome: Connecting language and vision using crowdsourced dense image annotations.
\newblock In {\em IJCV}, 2016.

\bibitem{hwang2011sharing}
Sung~Ju Hwang, Fei Sha, and Kristen Grauman.
\newblock Sharing features between objects and their attributes.
\newblock In {\em CVPR}, 2011.

\bibitem{mahajan2011joint}
Dhruv Mahajan, Sundararajan Sellamanickam, and Vinod Nair.
\newblock A joint learning framework for attribute models and object descriptions.
\newblock In {\em ICCV}, 2011.

\bibitem{cub}
Peter Welinder, Steve Branson, Takeshi Mita, Catherine Wah, Florian Schroff, Serge Belongie, and Pietro Perona.
\newblock Caltech-ucsd birds 200.
\newblock 2010.

\bibitem{li2019transferable}
Yong-Lu Li, Siyuan Zhou, Xijie Huang, Liang Xu, Ze~Ma, Hao-Shu Fang, Yanfeng Wang, and Cewu Lu.
\newblock Transferable interactiveness knowledge for human-object interaction detection.
\newblock In {\em CVPR}, 2019.

\bibitem{li2019hake}
Yong-Lu Li, Liang Xu, Xinpeng Liu, Xijie Huang, Yue Xu, Shiyi Wang, Hao-Shu Fang, Ze Ma, Mingyang Chen, and Cewu Lu.
\newblock PaStaNet: Toward Human Activity Knowledge Engine.
\newblock In {\em CVPR}, 2020.

\bibitem{hicodet}
Yu-Wei Chao, Yunfan Liu, Xieyang Liu, Huayi Zeng, and Jia Deng.
\newblock Learning to detect human-object interactions.
\newblock In {\em WACV}, 2018.

\bibitem{tmn}
Senthil Purushwalkam, Maximilian Nickel, Abhinav Gupta, and Marc'Aurelio Ranzato.
\newblock Task-driven modular networks for zero-shot compositional learning.
\newblock In {\em ICCV}, 2019.

\bibitem{chao2016empirical}
Wei-Lun Chao, Soravit Changpinyo, Boqing Gong, and Fei Sha.
\newblock An empirical study and analysis of generalized zero-shot learning for object recognition in the wild.
\newblock In {\em ECCV}, 2016.

\bibitem{ALE}
Zeynep Akata, Florent Perronnin, Zaid Harchaoui, and Cordelia Schmid.
\newblock Label-embedding for image classification
\newblock In {\em PAMI}, 2015.

\bibitem{HAP}
Sun-Wook Choi, Chong Ho Lee, and In Kyu Park.
\newblock Scene classification via hypergraph-based semantic attributes subnetworks identification.
\newblock In {\em ECCV}, 2014.

\bibitem{UDICA}
Chuang Gan, Tianbao Yang, and Boqing Gong.
\newblock Learning attributes equals multi-source domain generalization.
\newblock In {\em CVPR}, 2016.

\bibitem{GALM}
Zhi-Qi Cheng, Xiao Wu, Siyu Huang, Jun-Xiu Li, Alexander G Hauptmann, and Qiang Peng.
\newblock Learning to transfer: Generalizable attribute learning with multitask neural model search.
\newblock In {\em ACMMM}, 2018.

\bibitem{UMF}
Kongming Liang, Hong Chang, Bingpeng Ma, Shiguang Shan, and Xilin Chen.
\newblock Unifying visual attribute learning with object recognition in a multiplicative framework.
\newblock In {\em PAMI}, 2018.

\bibitem{FMT}
Yongxi Lu, Abhishek Kumar, Shuangfei Zhai, Yu Cheng, Tara Javidi, and Rogerio Feris.
\newblock Fully-adaptive feature sharing in multi-task networks with applications in person attribute classification.
\newblock In {\em CVPR}, 2017.

\bibitem{AMT}
Emily M Hand, and Rama Chellappa.
\newblock Attributes for improved attributes: A multi-task network utilizing implicit and explicit relationships for facial attribute classification.
\newblock In {\em AAAI}, 2017.


\bibitem{causal-czsl}
Yuval Atzmon, Felix Kreuk, Uri Shalit, and Gal Chechik.
\newblock A causal view of compositional zero-shot recognition.
\newblock In {\em arXiv:2006.14610}, 2020.

\bibitem{featuregen}
Yongqin Xian, Tobias Lorenz, Bernt Schiele, and Zeynep Akata.
\newblock Feature generating networks for zero-shot learning.
\newblock In {\em CVPR}, 2018.

\bibitem{HMF}
Jie Yang, Jiarou Fan, Yiru Wang, Yige Wang, Weihao Gan, Lin Liu, and Wei Wu.
\newblock Hierarchical Feature Embedding for Attribute Recognition.
\newblock In {\em arXiv:2005.11576}, 2020.

\bibitem{tang2019improving}
Chufeng Tang, Lu Sheng, Zhaoxiang Zhang, and Xiaolin Hu.
\newblock Improving Pedestrian Attribute Recognition With Weakly-Supervised Multi-Scale Attribute-Specific Localization.
\newblock In {\em CVPR}, 2018.

\bibitem{LI2020DOMAIN}
Yuze Li and Chunling Yang and Yu Chen and Yan Zhang.
\newblock Unsupervised domain adaptation with structural attribute learning networks.
\newblock In {\em Neurocomputing}, 2020.

\bibitem{2021learningGraphEmbeddings}
Muhammad Ferjad Naeem, Yongqin Xian, Federico Tombari, and Zeynep Akata.
\newblock Learning Graph Embeddings for Compositional Zero-shot Learning.
\newblock In {\em CVarXiv:2102.01987PR}, 2021.

\bibitem{idn}
Yong-Lu Li, Xinpeng Liu, Xiaoqian Wu, Yizhuo Li, and Cewu Lu.
\newblock HOI analysis: Integrating and decomposing human-object interaction.
\newblock In {\em NeurIPS}, 2020.

\bibitem{djrn}
Yong-Lu Li, Xinpeng Liu, Han Lu, Shiyi Wang, Junqi Liu, Jiefeng Li, and Cewu Lu.
\newblock Detailed 2D-3D Joint Representation for Human-Object Interaction.
\newblock In {\em CVPR}, 2020.

\bibitem{symnet}
Yong-Lu Li, Yue Xu, Xiaohan Mao, and Cewu Lu.
\newblock Symmetry and Group in Attribute-Object Compositions.
\newblock In {\em CVPR}, 2020.

\end{thebibliography}

%

\begin{IEEEbiography}[{\includegraphics[width=1in,height=1.25in,clip,keepaspectratio]{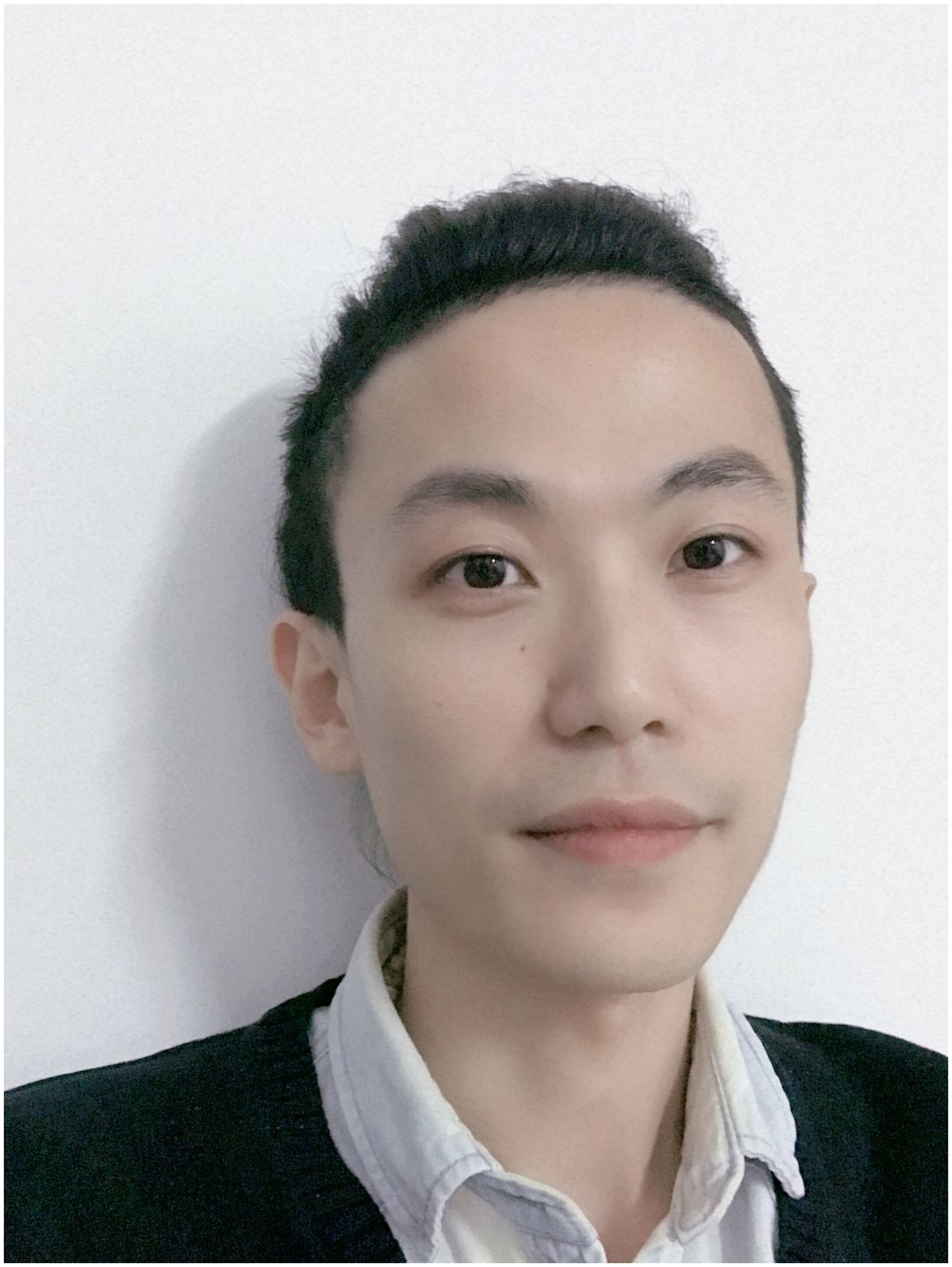}}]{Yong-Lu Li}
received a Ph.D. degree in computer science and technology from Shanghai Jiao Tong University under the supervision of Prof. Cewu Lu. He has received the Baidu Scholarship, YunFan award, Shanghai Outstanding Graduate, etc. His research interests include computer vision, reasoning, and embodied AI. He has developed HAKE, which is a knowledge-driven perception and reasoning system for human-scene-robot interaction.
\end{IEEEbiography}

\begin{IEEEbiography}[{\includegraphics[width=1in,height=1.25in,clip,keepaspectratio]{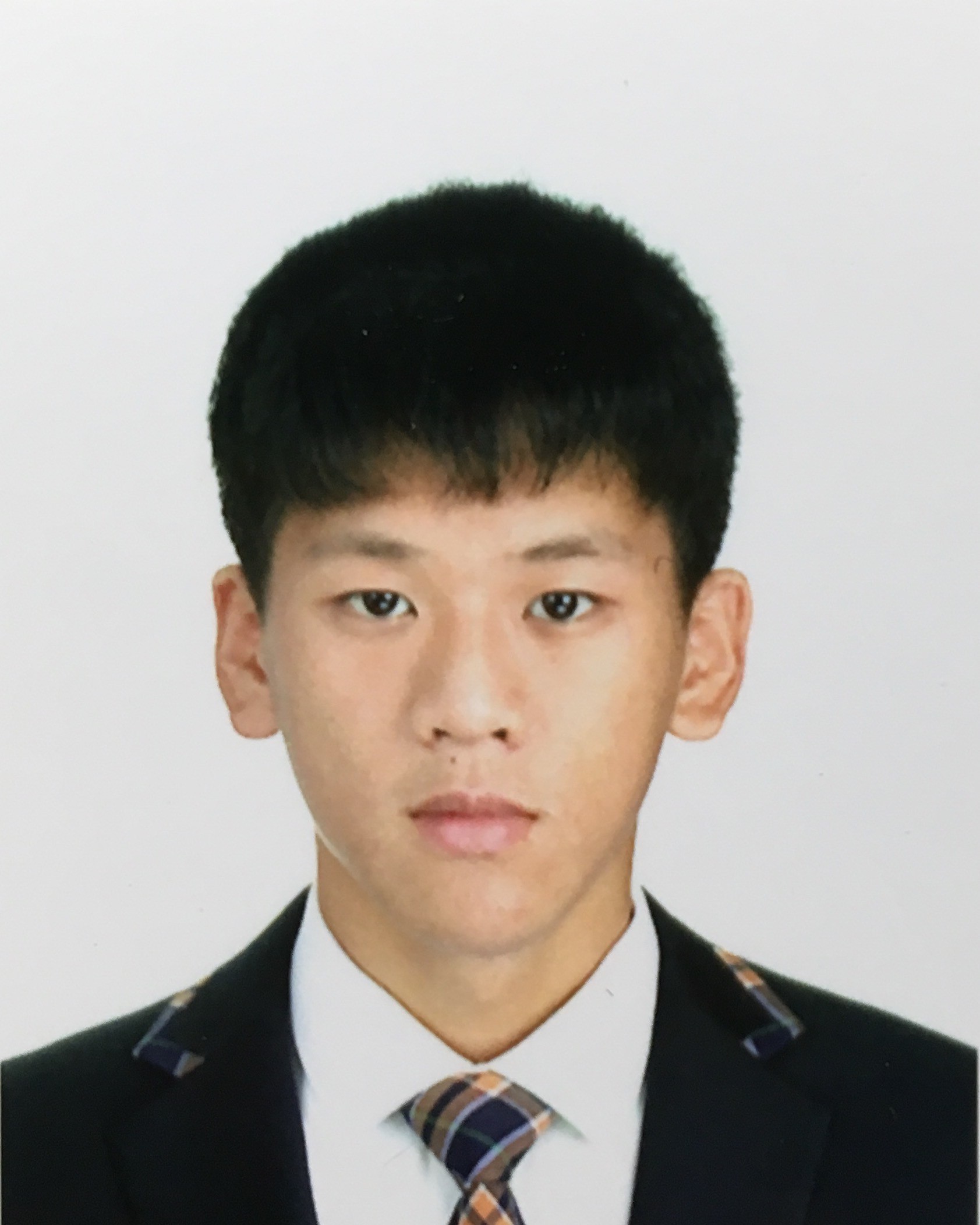}}]{Yue Xu}
received a B.E. degree, and is working toward a master's degree in computer science from Shanghai Jiao Tong University, Shanghai, China. His research interests mainly include computer vision.
\end{IEEEbiography}

\begin{IEEEbiography}[{\includegraphics[width=1in,height=1.25in,clip,keepaspectratio]{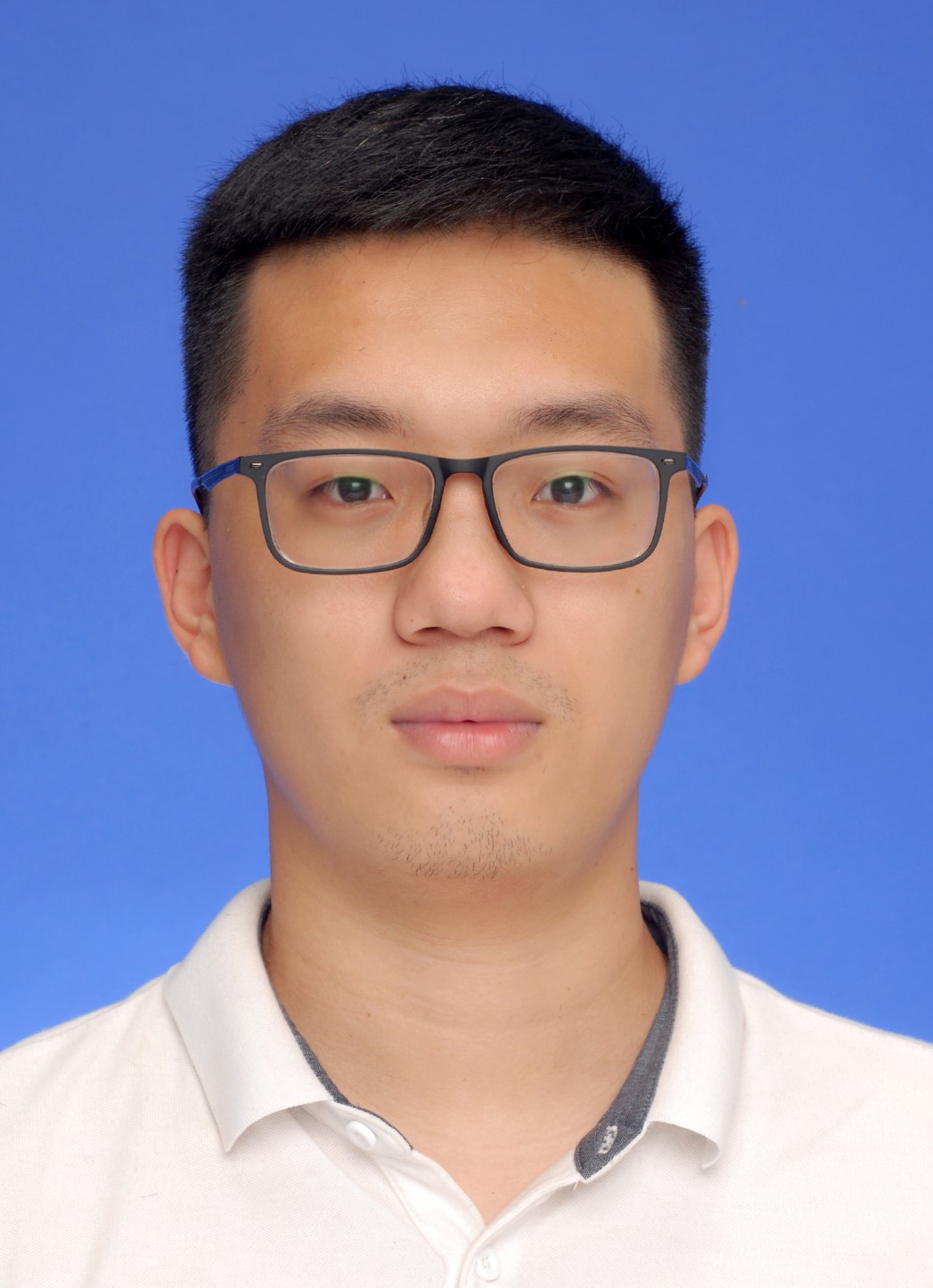}}]
{Xinyu Xu}
is an undergraduate student majoring in Computer Science and Engineering, Shanghai Jiao Tong University. His research interests include computer vision and robotics.
\end{IEEEbiography}

\begin{IEEEbiography}[{\includegraphics[width=1in,height=1.25in,clip,keepaspectratio]{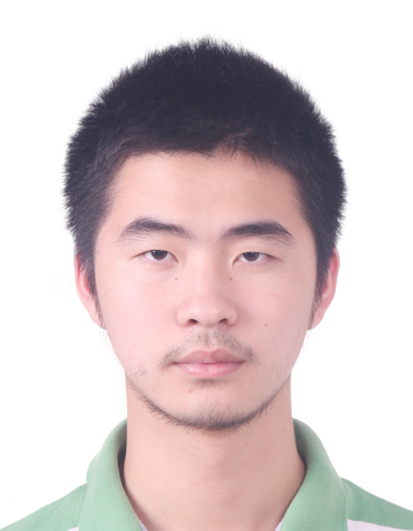}}]
{Xiaohan Mao}
is an undergraduate student majoring in Computer Science and Engineering, ACM class, Shanghai Jiao Tong University. His research interests include computer vision.
\end{IEEEbiography}

\begin{IEEEbiography}[{\includegraphics[width=1in,height=1.25in,clip,keepaspectratio]{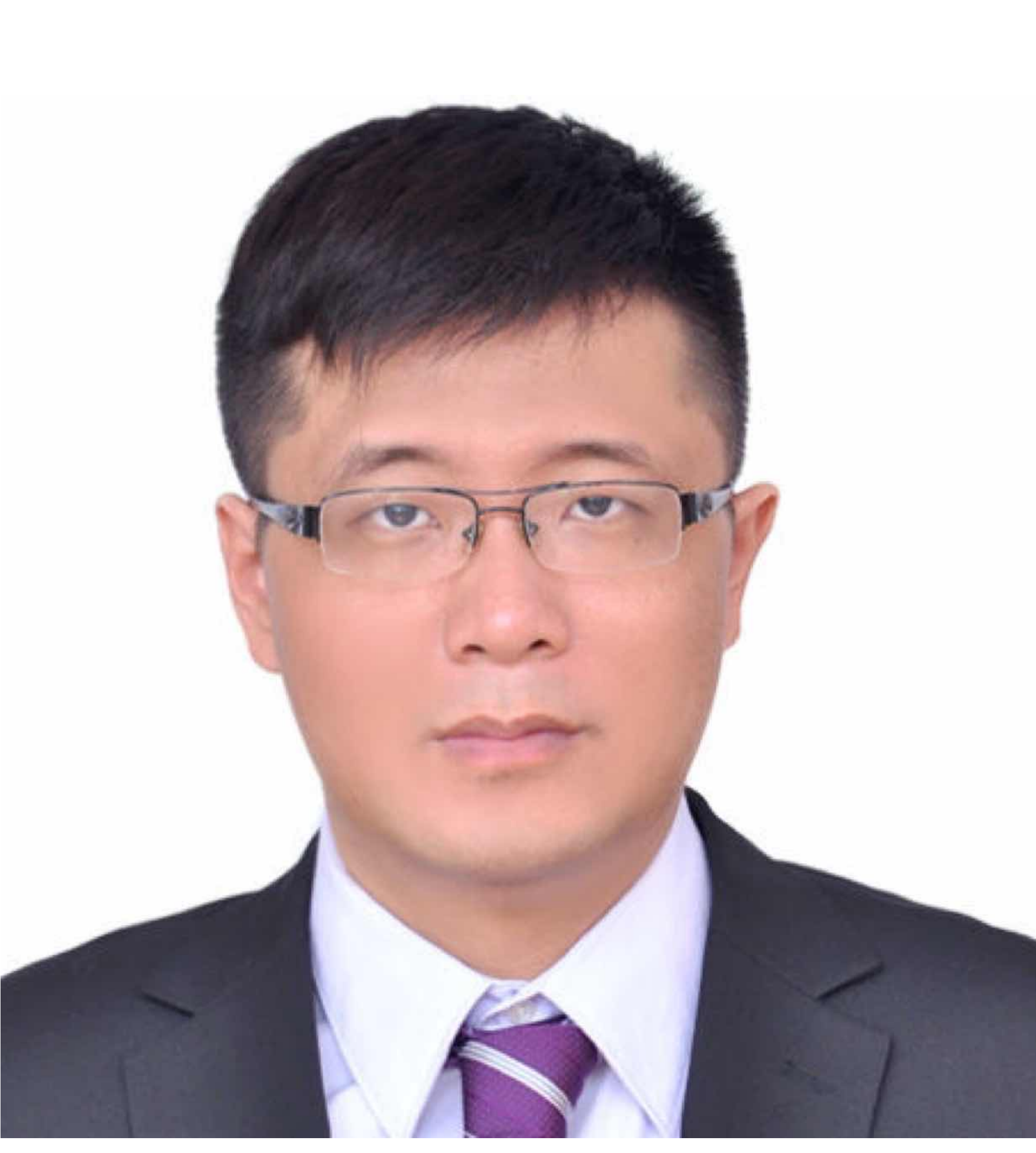}}]{Cewu Lu}
is an Associate Professor at Shanghai Jiao Tong University (SJTU). 
Before joining SJTU, he was a research fellow at Stanford University, working under Prof. Fei-Fei Li and Prof. Leonidas J. Guibas. He was a Research Assistant Professor at Hong Kong University of Science and Technology with Prof. Chi Keung Tang. He got his Ph.D. degree from The Chinese University of Hong Kong, supervised by Prof. Jiaya Jia. His research interests fall mainly in computer vision, deep learning, and robotics.
\end{IEEEbiography}





\appendix
This is a PAMI version of our CVPR'20 work SymNet~\cite{symnet}.

\section{Specific Hyper-parameters}
The specific hyper-parameters in experiments are shown in Tab.~\ref{tab:paramter}.
\begin{table*}[ht]
	\begin{center}
		\resizebox{0.95\textwidth}{!}{
		\begin{tabular}{lcccccc}
		\hline  
	   Dataset  & MIT-States~\cite{mit} & MIT-States (generalized)~\cite{mit} & UT-Zappos~\cite{ut} & UT-Zappos (generalized)~\cite{ut} & aPY~\cite{apy} & SUN~\cite{sun}    \\
	    \hline 
	    Learning rate   & 5e-4 & 3e-4 & 1e-4 & 1e-3 & 3e-3 & 5e-3   \\
	    Batch size      & 512  & 512  & 256  & 512  & 128  & 128    \\
	    Epoch           & 320  & 1000 & 600  & 290  & 177  & 95    \\
	    $\lambda_1$     & 5e-2 & 2e-2 & 1e-2 & 2e-2 & 5e-2 & 8e-3 \\
	    $\lambda_2$     & 1e-2 & 2e-2 & 3e-2 & 1e-2 & s1e-3 & 1e-3 \\
	    $\lambda_3$     & 1    & 1    & 1    & 1    & 1    & 1    \\
	    $\lambda_4$     & 1e-2 & 1e-2 & 5e-1 & 1-e2 & 5e-2 & 3e-1 \\
	    $\lambda_5$     & 3e-2 & 1    & 5e-1 & 1    & 1    & 5e-2 \\
	    $\lambda_6$     & /    & /    & /    & /    & 5e-2 & 6e-2 \\
	    $\lambda_7$     & /    & /    & /    & /    & 1    & 6e-1 \\ 
	    Triplet margin  & 0.5 & 0.3 & 0.5 & 0.5 & 0.5 & 0.5 \\
	    \hline
		\end{tabular}
		}
	\end{center}
	\caption{Hyper-parameters on four benchmarks.}
	\label{tab:paramter}
\end{table*}

\begin{figure*}[htbp]
	\begin{center}
        \begin{minipage}{.33\linewidth}
            \centerline{\includegraphics[width=\linewidth]{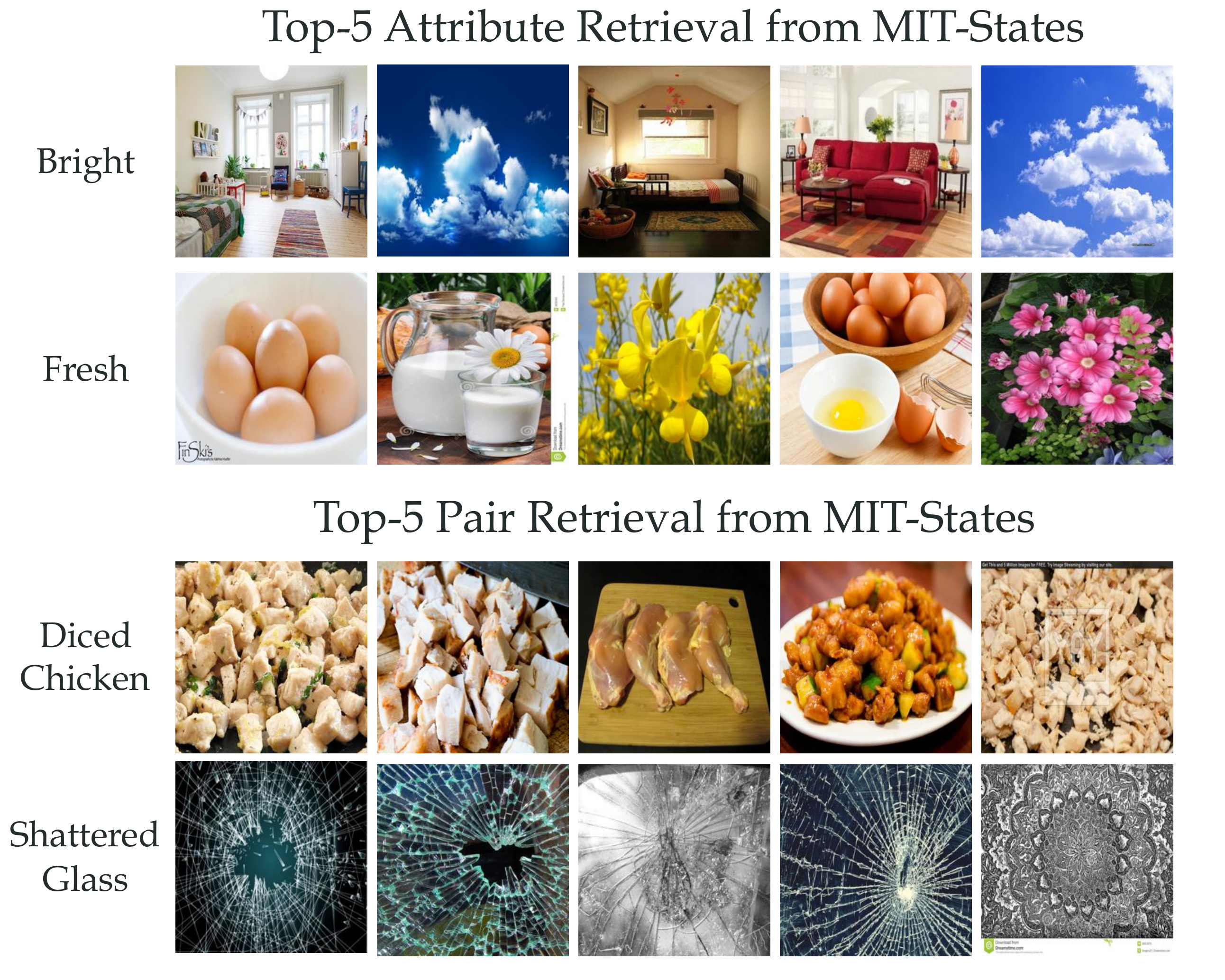}}
            \centerline{(a) MIT-States}
        \end{minipage}
        \hfill
        \begin{minipage}{.33\linewidth}
            \centerline{\includegraphics[width=\linewidth]{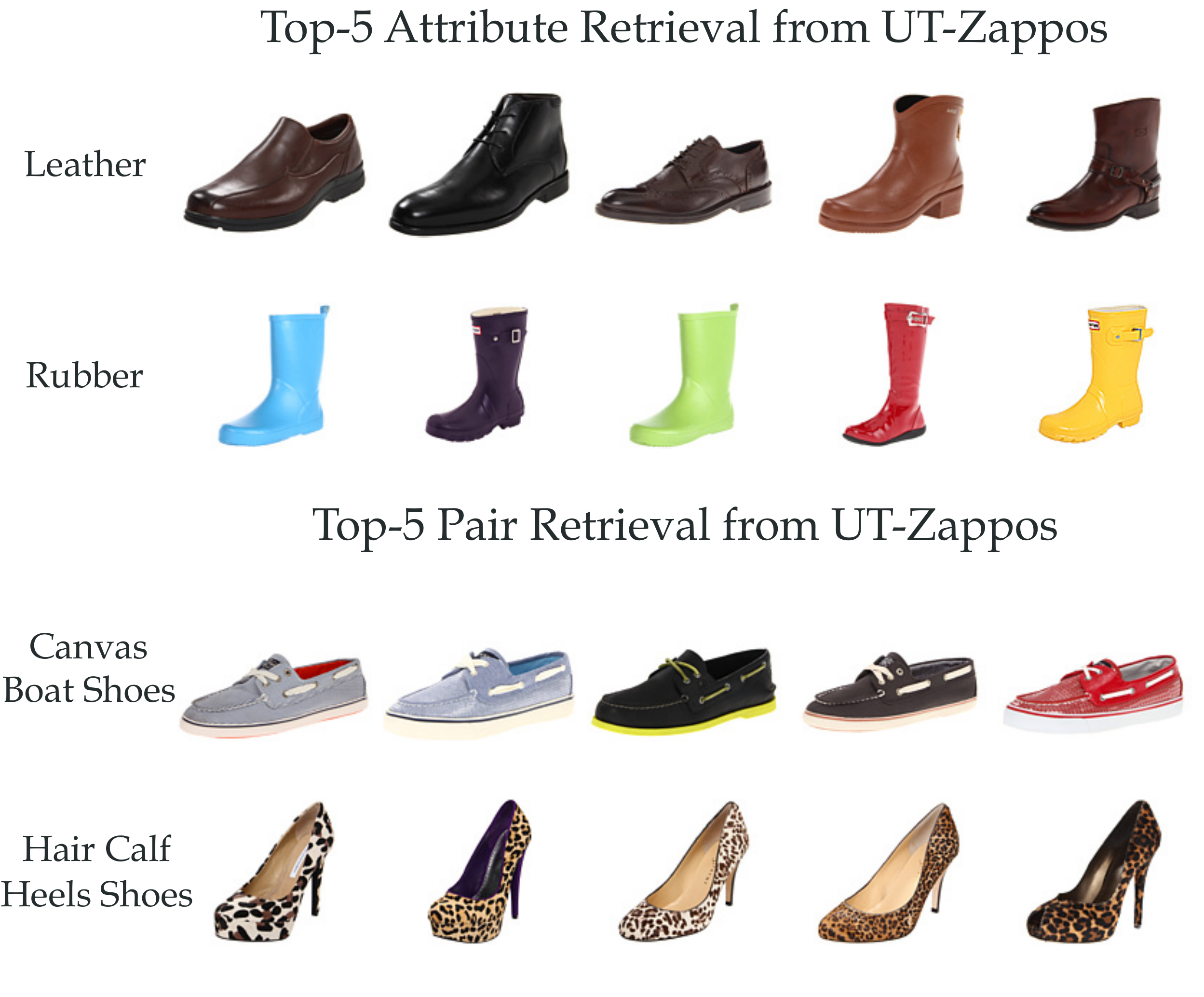}}
            \centerline{(b) UT-Zappos}
        \end{minipage}
        \hfill
        \begin{minipage}{.33\linewidth}
            \centerline{\includegraphics[width=\linewidth]{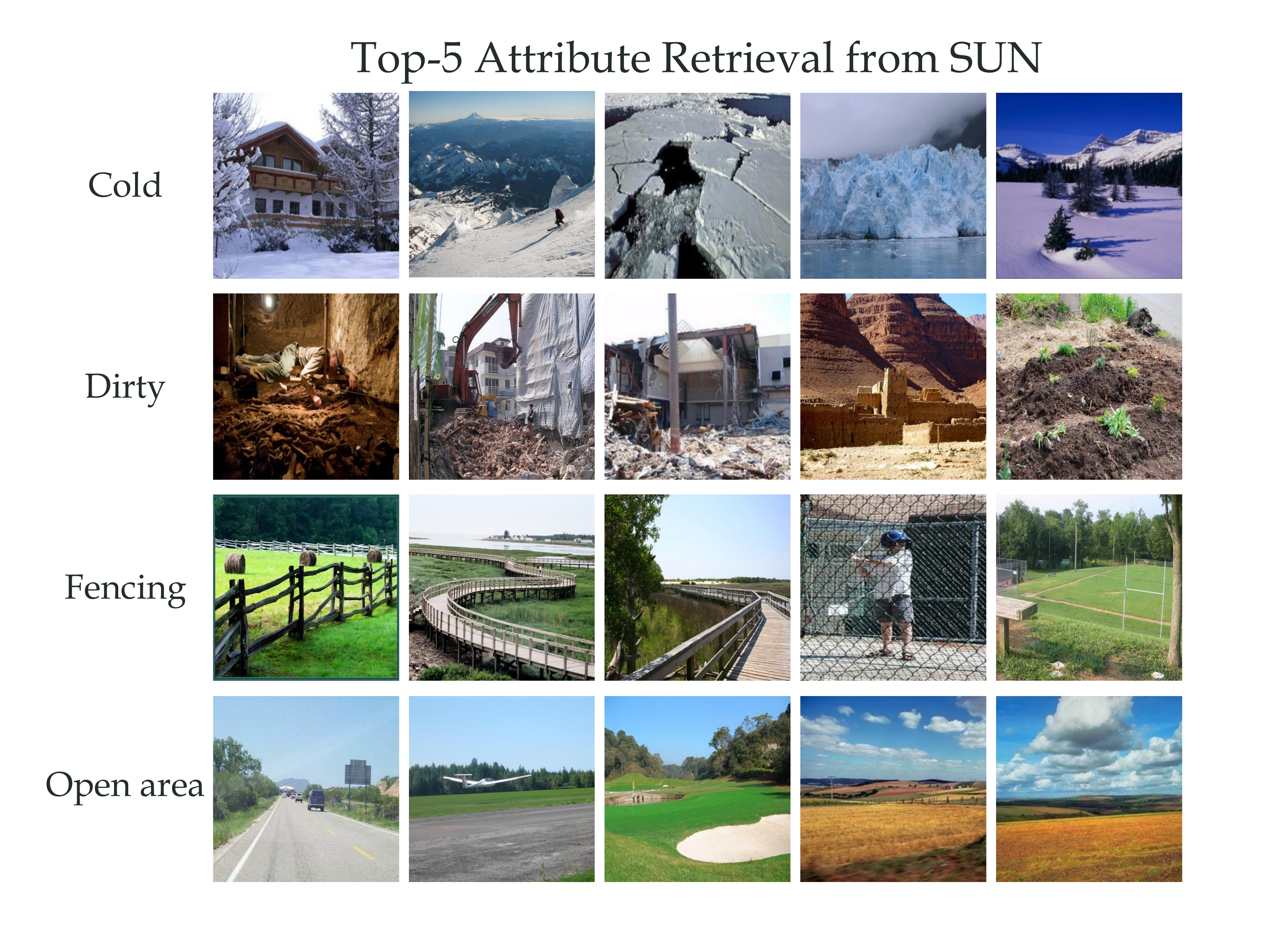}}
            \centerline{(c) SUN}
        \end{minipage}
        \vfill
        \begin{minipage}{.33\linewidth}
            \centerline{\includegraphics[width=\linewidth]{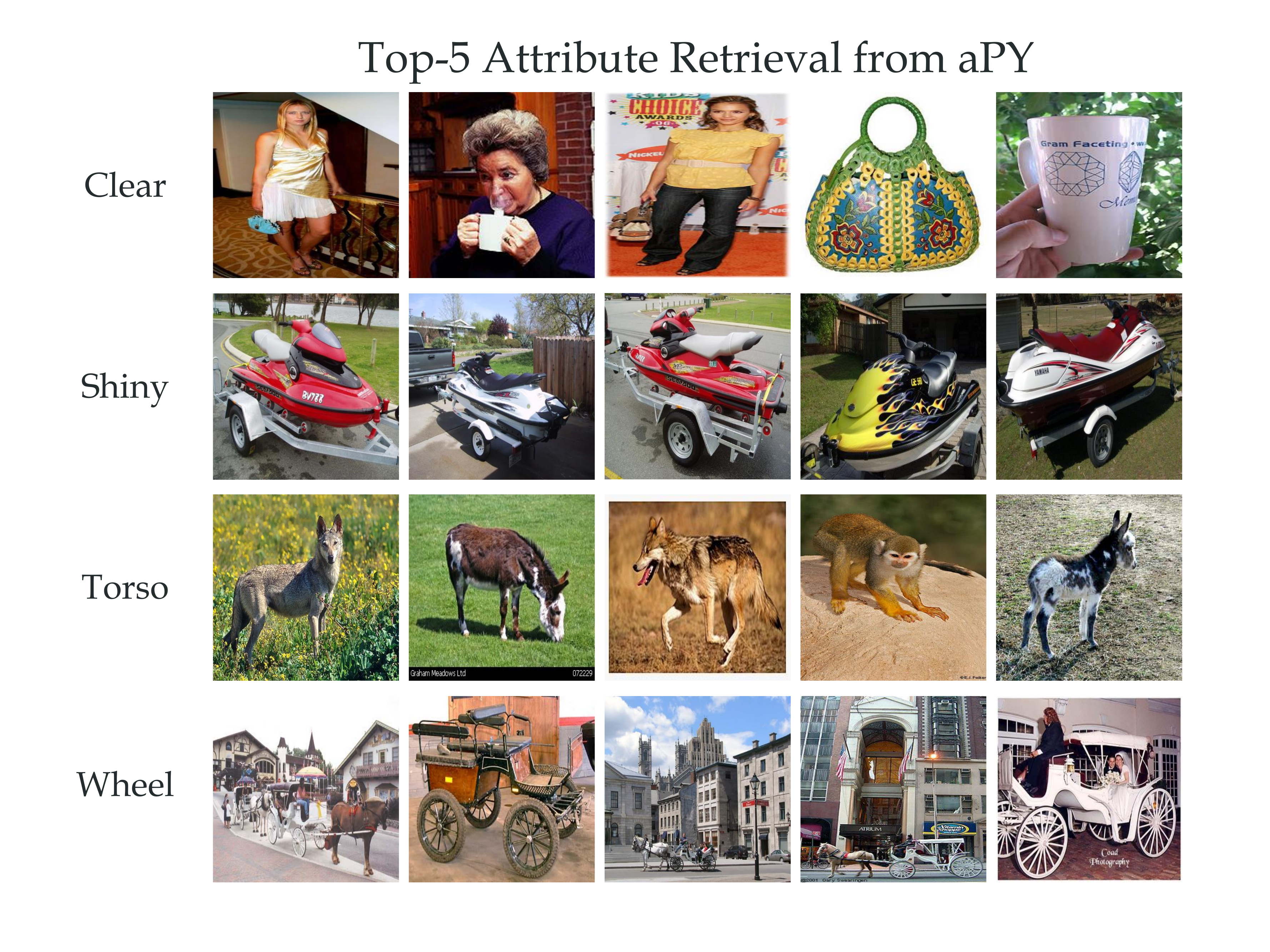}}
            \centerline{(d) aPY}
        \end{minipage}
        \hfill
        \begin{minipage}{.33\linewidth}
            \centerline{\includegraphics[width=\linewidth]{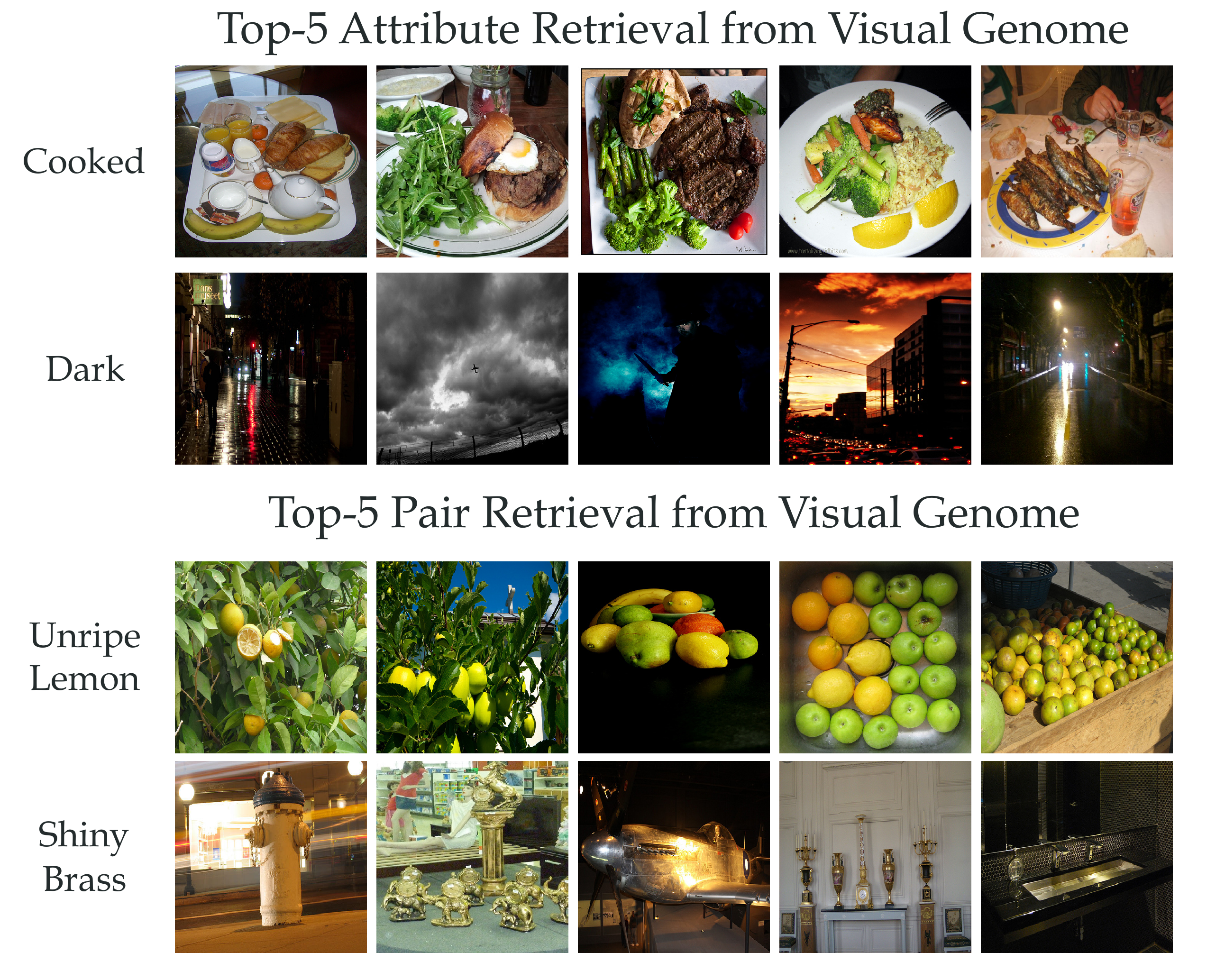}}
            \centerline{(e) Visual Genome (out-of-domain)}
        \end{minipage}
        \hfill
        \begin{minipage}{.33\linewidth}
            \centerline{\includegraphics[width=\linewidth]{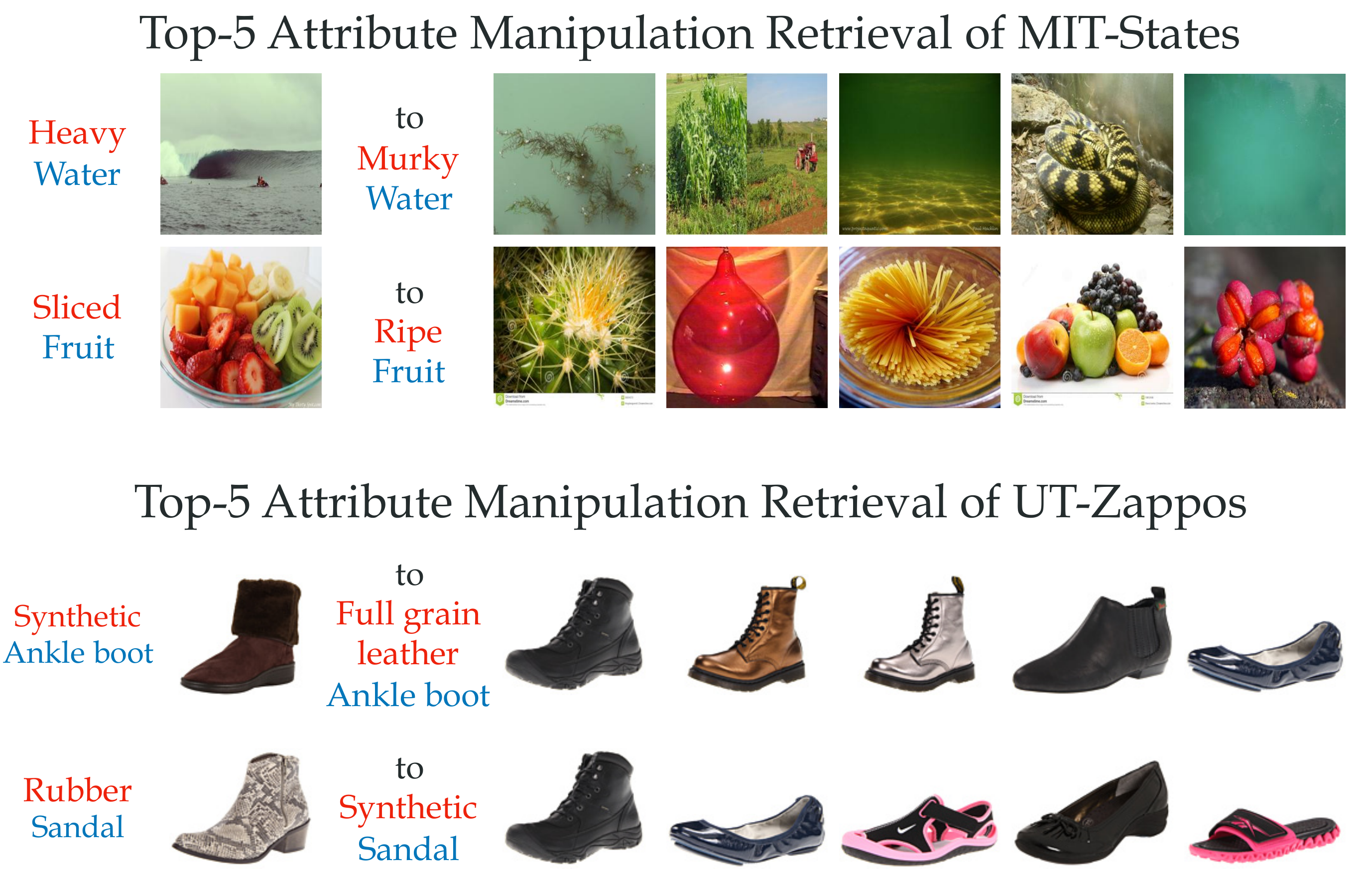}}
            \centerline{(f) Attribute Manipulation}
        \end{minipage}
	\end{center}
	\caption{Image retrieval results. On MIT-States~\cite{mit}, UT-Zappos~\cite{ut}, SUN~\cite{sun}, aPY~\cite{apy}, and Visual Genome~\cite{visualgenome}, the first row shows retrievals of attributes and the second is retrievals of unseen attribute-object pairs. Specially, the retrieval model in out-of-domain~\cite{operator} mode is not trained on Visual Genome. At last, we also show the retrievals after the attribute manipulation via SymNet.}
	\label{Figure:retrieval}
\end{figure*}

\section{Dataset and Baseline Details}
Here, we detail the datasets and baselines adopted in the experiments.

\subsection{Datasets}
\noindent\textbf{aPY}~\cite{apy} consists of two \textbf{multi-attribute} datasets, aPascal and aYahoo, both with 64 attributes. aPascal has 6,340 training samples and 6,355 testing samples covering 20 objects. aYahoo contains 2,644 images for testing, covering 12 objects disjoint from aPascal. Following \cite{UDICA,GALM}, we train our model on aPascal and test on aYahoo.

\noindent\textbf{SUN}~\cite{sun} is a \textbf{multi-attribute} dataset, contains 14,340 images with 102 attributes and 717 objects. 10,320 and 4,020 images are used for training and testing respectively.

\noindent\textbf{MIT-States}~\cite{mit} contains 63,440 images covering 245 objects and 115 attributes. Each image is attached with one \textbf{single} object-attribute composition label and there are 1,262 possible pairs in total. We follow the setting of \cite{redwine} and use 1,262 pairs/34,562 images for training and 700 pairs/19,191 images as the test set.

\noindent\textbf{UT-Zappos50K}~\cite{ut} is a fine-grained and \textbf{single-attribute} dataset with 50,025 images of shoes annotated with shoe type-material pairs. We follow the setting and split from \cite{operator}, using 83 object-attribute pairs/24,898 images as the train set and 33 pairs/4,228 images for testing. 

\subsection{Baselines for Single-Attribute Learning and CZSL}
\noindent\textbf{Visual Product} trains two simple classifiers for attributes and objects independently and fuses the outputs by multiplying their margin probabilities: $P(a,o)=P(a)P(o)$. The classifiers can be either linear SVMs~\cite{redwine} or single layer softmax regression models~\cite{operator}.

\noindent\textbf{LabelEmbed (LE)}~\cite{redwine} combines the word vectors~\cite{glove} of attribute and object and uses 3-layer FCs to transform the pair embedding into a transform matrix. The classification score is the product of transform matrix and visual feature: 
\begin{enumerate}[nosep]
    \item \textbf{LabelEmbed Only Regression (LEOR)}~\cite{redwine} changes the target to minimize the Euclidean distance between $\mathcal{T}\left(e_{a}, e_{b}\right)$ and the weight of pair SVM classifier $w_{ab}$.
    \item \textbf{LabelEmbed With Regression (LE+R)}~\cite{redwine} combines the losses of LE and LEOR aforementioned.
    \item \textbf{LabelEmbed+}~\cite{operator} embeds the attribute, object vectors, and image features into a semantic space and also optimizes the input representations during training.
\end{enumerate}
\noindent\textbf{AnalogousAttr}~\cite{analogous} trains linear classifiers for seen compositions and uses tensor completion to generalize to the unseen pairs. We report the reproduced results from \cite{operator}.

\noindent\textbf{Red Wine}~\cite{redwine} uses SVM weights as the attribute or object embeddings to replace the word vectors in LabelEmbed.

\noindent\textbf{AttrOperator}~\cite{operator} regards attributes as linear transformations and object word vectors~\cite{glove} after transformation as pair embeddings. It takes the pair with the closest distance to the image feature as the recognition result. Besides the top-1 accuracy directly reported in \cite{operator}, we evaluate the top-2, three accuracies with the open-sourced code.

\noindent\textbf{TAFE-Net}~\cite{tafe} uses word vectors~\cite{word2vec} of attribute-object pair as task embeddings of its meta learner. It generates a binary classifier for each existing composition. We report the results based on VGG-16, which is \textit{better} and more complete than the result based on ResNet-18.

\noindent\textbf{GenModel}~\cite{genmodel} projects the visual features of images and semantic language embeddings of pairs into a shared latent space. The prediction is given by comparing the distance between visual features and all candidate pair embeddings.

\noindent\textbf{f-CLSWGAN}~\cite{featuregen} generates unseen class features via GAN and train a classifier jointly with real and generated features.

\noindent\textbf{TMN}~\cite{tmn} adopts a set of FC-based modules and configure them via a gating function in a task-driven way. It can be generalized to unseen pairs via re-weighting primitive modules.

\noindent\textbf{Causal}~\cite{causal-czsl} proposes a causal view of CZSL and learns better representations by disentangling the attribute and object features according to the conditional independence principle. The predictions are given according to the distance between learned features and the attribute/object centers. 

\subsection{Baselines for Multiple-Attribute Learning}
\textbf{ALE}~\cite{ALE} embeds objects with category-level attributes. It trains attribute classifiers with an objective to meet correct object embedding. 
We report the score reproduced by \cite{GALM}.

\noindent\textbf{HAP}~\cite{HAP} constructs hyper-graph to learn the correlations of semantic attributes. We report the result by \cite{UDICA}.

\noindent\textbf{UDICA/KDICA}~\cite{UDICA} regularizes the distributional variance to achieve cross-domain attribute generalization. KDICA integrates kernel alignment for a unified optimization. The score on SUN \cite{sun} is reproduced by \cite{GALM}.

\noindent\textbf{UMF}~\cite{UMF} projects both image features and category labels to a common latent space. Then it makes element-wise multiplication and predicts attributes. We report the score reproduced by \cite{GALM}.

\noindent\textbf{AMF}~\cite{AMT} designs a multi-task deep neural network for multiple attribute prediction. It uses an auxiliary network to explore attribute relations further. We report the score reproduced by \cite{GALM}.

\noindent\textbf{FMT}~\cite{FMT} automatically designs a neural network that greedily makes branch and task-grouping decisions in each layer. We report the score reproduced by \cite{GALM}.

\noindent\textbf{GALM}~\cite{GALM} applies a tree-structured model. Its root node is shared for all attributes, but leave nodes are independently and automatically searched.

\section{Image Retrieval}
To qualitatively evaluate our method, we further report the image retrieval results of SymNet. We follow the settings of \cite{operator}: 
1) \textbf{In-domain attributes or unseen compositions}: we train SymNet on MIT-States~\cite{mit}, UT-Zappos~\cite{ut}, SUN~\cite{sun} and aPY~\cite{apy} respectively, and query the attributes or unseen pairs upon the test set of each dataset. The results are displayed in Fig.~\ref{Figure:retrieval} (a, b, c, d). 
2) \textbf{Out-of-domain retrieval}: with SymNet \textit{only trained on MIT-States}, we conduct retrieval on the large-scale Visual Genome~\cite{visualgenome} with over 100K images, which is non-overlapping with the train set of MIT-States. The results are shown in Fig.~\ref{Figure:retrieval}(e).

Our model is capable of recognizing the images with queried attributes and pairs in most cases. When querying an attribute, it accurately retrieves images across various objects, e.g. for MIT-States, the top-5 retrievals of attribute \texttt{fresh} vary among \texttt{fresh-egg}, \texttt{fresh-milk} and \texttt{fresh-flower}, suggesting that our model has well exploited the contextuality and compositionality of attributes. 
In out-of-domain retrieval, SymNet also shows its robustness. Though it has never seen the images in Visual Genome~\cite{visualgenome}, SymNet generalizes well on the target domain and returns correct retrievals, e.g. \texttt{dark} objects and \texttt{unripe lemon}. 

We further report the image retrieval results after attribute manipulation. 
We first train SymNet on MIT-States~\cite{mit} or UT-Zappos~\cite{ut}, then use trained CoN and DeCoN to manipulate the image embeddings. 
For an image with pair label $(a,o)$, we remove the attribute $a$ with DeCoN and add an attribute $b$ with CoN. Then we retrieve the top-5 nearest neighbors of the manipulated embeddings. 
This task is much more difficult than the normal attribute-object retrieval~\cite{operator,redwine,tmn} because of the complex semantic manipulation and recognition.
The results are shown in Fig.~\ref{Figure:retrieval}(f), where the images on the left are original ones and the right ones are the nearest neighbors after manipulation.
SymNet is capable of retrieving a certain number of correct samples among the top-5 nearest neighbors, especially in a fine-grained dataset like UT-Zappos~\cite{ut}, suggesting that our model has well exploited the learned symmetry in attribute transformation and learned the contextuality and compositionality of attributes.

\section{Supplementary Experiments}

\begin{table}[t]
	\centering
	\small
	\adjustbox{width=\linewidth}{
		\begin{tabular}{l|ccc|ccc}
			\toprule
			& AUC Top 1 & AUC Top-2 & AUC Top-3 & Seen Acc. & Unseen Acc. & H-Mean \\
			\midrule
			SymNet & \textbf{5.4} & \textbf{11.6} & \textbf{16.6} & \textbf{30.4}  &\textbf{25.8} &  \textbf{17.6} \\ 
			\midrule
			SymNet w/o $\mathcal{L}_{sym}$   & 4.0 & 9.2 & 14.0 & 24.2 & 24.4 & 15.4 \\
			SymNet w/o $\mathcal{L}_{axiom}$ & 4.0 & 9.4 & 14.1 & 25.3 & 23.4 & 15.3 \\
			SymNet w/o $\mathcal{L}_{inv}$   & 4.3 & 9.6 & 14.4 & 26.0 & 24.6 & 15.8 \\
			SymNet w/o $\mathcal{L}_{com}$   & 4.4 & 9.6 & 14.5 & 26.5 & 24.8 & 15.8 \\
			SymNet w/o $\mathcal{L}_{clo}$   & 4.2 & 9.9 & 14.8 & 25.9 & 24.3 & 15.5 \\
			SymNet w/o $\mathcal{L}_{cls}$   & 1.6 & 4.7 & 8.4  & 13.3 & 19.9 & 9.6  \\
			SymNet w/o $\mathcal{L}_{tri}$   & 3.9 & 9.7 & 14.7 & 25.8 & 23.5 & 15.6 \\
			SymNet w/o $\mathcal{L}_{sym}$ \& $\mathcal{L}_{tri}$ & 3.9 & 9.0 & 14.1 & 24.8 & 24.2 & 15.1 \\
			SymNet w/o $\mathcal{L}_{tri}$ \& $\mathcal{L}_{cls}$ & 1.6 & 4.7 & 8.2  & 12.7 & 19.3 & 9.5  \\
			SymNet w/o $\mathcal{L}_{sym}$ \& $\mathcal{L}_{cls}$ & 1.7 & 4.7 & 8.1  & 13.1 & 19.5 & 9.9  \\
			SymNet only $\mathcal{L}_{sym}$ & 1.6 & 4.7 & 8.4 & 12.4 & 20.2 & 9.4 \\
			SymNet only $\mathcal{L}_{cls},\mathcal{L}_{tri}$ & 3.7 & 9.0 & 13.5 & 25.0 & 23.2 & 15.0 \\
			\midrule
			SymNet w/o attention     & 3.9 & 9.1 & 13.8 & 23.2 & 24.8 & 15.1 \\
			SymNet $\tanh$ attention & 3.8 & 8.7 & 13.1 & 24.5 & 23.6 & 14.7 \\
			\midrule
			SymNet $L_1$ dist. & 2.8 & 6.3 & 9.6  & 22.9 & 17.8 & 12.9  \\
			SymNet $Cos$ dist. & 1.5 & 4.4 & 8.0  & 12.0 & 20.4 & 9.1  \\
			\bottomrule
	\end{tabular}}
	\caption{\small Results of ablation studies in generalized CZSL setting~\cite{tmn} on MIT-States~\cite{mit} validation set.}
	\label{tab:abl-gczsl}
\end{table}

\subsection{Supplementary Ablation Studies}
We further conduct more ablation studies under different metrics on multi-dataset to fully evaluate our methods. 
Precisely, we follow the metric of TMN~\cite{tmn} to conduct the ablations similar to the Tab.7 of the main text.
The results are listed in supplementary Tab.~\ref{tab:abl-gczsl}, which can further verify the effectiveness of the proposed components in SymNet. For more analyses, please refer to Sec.~4.8 of the main text.

\subsection{Comparison to AttrOperator}

\begin{table}[t]
	\centering
	\small
	\adjustbox{width=0.95\linewidth}{
		\begin{tabular}{l|ccc|ccc}
			\toprule
			& \multicolumn{3}{c|}{MIT-States} & \multicolumn{3}{c}{UT-Zappos} \\
			& Top-1 & Top-2 & Top-3 & Top-1 & Top-2 & Top-3 \\
			\midrule
			SymNet + SymNet losses (ours)       & {\bf 19.9} & {\bf 28.2} & {\bf 33.8} & {\bf 52.1} & {\bf 67.8} & {\bf 76.0} \\
			SymNet + AttrOperator losses        & 17.0 & 26.0 & 31.7 & 49.8 & 66.3 & 73.8 \\
			SymNet + Linear operator            & 16.5 & 25.5 & 31.3 & 49.6 & 66.2 & 73.9 \\
			\midrule
			AttrOperator + AttrOperator losses  & 14.2 & 19.6 & 25.1 & 46.2 & 56.6 & 69.2 \\
			AttrOperator + SymNet losses        & 14.4 & 19.9 & 25.7 & 46.5 & 56.7 & 70.1 \\
			\bottomrule
	\end{tabular}}
	\caption{\small Cross comparison of SymNet and AttrOperator~\cite{operator}.}
	\label{tab:abl-comp-operator}
\end{table}

Here, we compare SymNet and AttrOperator~\cite{operator} since they all see attribute change as transformation. However, the design of SymNet is quite different from AttrOperator~\cite{operator} in many aspects.

First, the operators in AttrOperator are applied on \textit{object Glove~\cite{glove} embeddings} to compose the anchor representations of compositions for classification, while our CoN and DecoN are applied on \textit{image representations} and we use relative moving distance for classification. Though both methods regard attributes as manipulations in latent space, the usage of manipulators, the definition of the latent space, the classification paradigm, and the corresponding constraint design are quite different.

Second, as a consequence of different model designs, our losses are naturally based on group axioms and Symmetry property, while the losses of AttrOperator are designed to conform with the linguistic meaning of attributes~\cite{operator}.
Among these losses, $\mathcal{L}_{cls}, \mathcal{L}_{inv}, \mathcal{L}_{com}$ work similar and the others are different. 
To evaluate the loss difference and effectiveness, we conduct an ablation study of training SymNet model with AttrOperator losses and training AttrOperator model with SymNet losses. The results are shown in Tab.~\ref{tab:abl-comp-operator}. 
Comparing to the original models, SymNet with AttrOperator losses drops \textbf{2.9}\% top-1 accuracy, while AttrOperator with SymNet losses instead achieves a slight improvement, indicating our axiom and symmetry-based losses are more complete and especially suitable to SymNet. 

Third, we conduct an ablation study about representation power of \textit{linear} model and CoN/DecoN, as show in Tab.~\ref{tab:abl-comp-operator}. We implement CoN and DecoN in SymNet as multiple independent linear matrices following~\cite{operator}, which leads to \textbf{3.4}\% accuracy drop. The experiments show that our CoN and DecoN are more expressive.

\subsection{Orthogonal Constraint}
\begin{table}[!ht]
	\centering
	\small
	\adjustbox{width=0.7\linewidth}{
		\begin{tabular}{l|c|c}
			\toprule
			Method & aPY & SUN \\ 
			\midrule
			SymNet (multi)                   & \textbf{86.1} & \textbf{88.4} \\
			SymNet (multi) + orthogonal      & 85.9 &  \textbf{88.4} \\ 
			\midrule
			SymNet (single)                  & 82.2 & 88.1  \\
			SymNet (single) + orthogonal     & \textbf{84.4} & \textbf{88.3}  \\
			\bottomrule
	\end{tabular}}
	\caption{\small Results of ablation studies of orthogonality constraint~\cite{comp}.}
	\label{tab:ortho}
\end{table}
Different from the constraints in SymNet, COMP~\cite{comp} proposed a more straightforward method on multi-attribute learning via orthogonal constraint. To compare their effectiveness, we conduct an extra ablation study on regularizing the orthogonality of \textbf{attribute attentions} generated by CoN and DeCoN. 
Results are reported in Tab.~\ref{tab:ortho}.

First, we directly apply the orthogonal loss into the \textbf{multi-attribute} setting of SymNet (both the orthogonal loss and our multi-attribute correlation loss are used). 
It has no noticeable effect and leads to a small degrade on aPY~\cite{apy}. 
The reason is that, in the multi-attribute setting of SymNet, we treat each attribute pair concerning their prior attribute correlation in Eq.~10 while orthogonality constraint forces each attribute pair to be orthogonal fairly. 
Thus, in this scenario, it may not be appropriate to replace the customized attribute relationships with constant orthogonality, e.g., \textit{metal} and \textit{shiny} are strongly correlated attributes that affect objects similarly in transformation and should have similar attribute attentions. 
Thus, attribute correlation is sound side knowledge in multi-attribute learning.

Second, we apply orthogonal loss into the \textbf{single-attribute} setting of SymNet (without the multi-attribute correlation loss). It brings 2.2\% and 0.2\% performance improvements on aPY~\cite{apy} and SUN~\cite{sun} respectively. 
In this scenario, original attribute attentions are challenging to be distinguished without any regularization. Therefore, orthogonality constraint avoids the dense clusters of attribute attentions and helps SymNet distinguish the different attributes' roles in coupling and decoupling. 
Thus, it is a good way to classify multiple attributes without attribute correlation.
However, attribute correlation on aPY is very strong, as revealed in Fig.~5 (main text), and there is still a performance gap (1.7\%, between 86.1\% and 84.4\%) between orthogonal constraint and multi-attribute SymNet constraints.

In conclusion, without the attribute correlation, orthogonality constraint is a good alternative way to regularize multiple attributes for SymNet. However, it does not achieve the effect of attribute correlation. 

\subsection{Additional Details of Few-Shot Learning}
To further verify that whether SymNet can help few-shot recognition learn better representations, we conduct two few-shot recognition experiments on CUB-200-2011~\cite{cub} and SUN397~\cite{sun} datasets with the same experimental protocol to COMP~\cite{comp}. 

CUB-200-2011 dataset~\cite{cub} is proposed for finer-grained classification of birds. It contains 11,788 images of 200 categories and is split into train and test sets evenly. CUB-200-2011 originally collects 307 category-level attribute annotations. In COMP~\cite{comp}, 130 filtered attributes are utilized since some rare attributes hurt the knowledge transfer in few-shot recognition~\cite{comp}. Furthermore, the dataset is randomly split into 100 bases and 100 novel categories. We follow their setting here.
SUN397~\cite{sun} is a scene recognition dataset containing 108,754 images from 397 balanced categories. We use category-level labels of 89 attributes aggregated and filtered by COMP~\cite{comp}. Moreover, the dataset is randomly split into 197 bases and 200 novel categories.

SymNet is a feature extractor across object categories guided by the group theory constraints. Its attribute transformation feature is rich in attribute semantics and can be used to strengthen the original object representation. 
Thus, given SymNet trained on two datasets respectively, we can use the concatenation of $f_o$, $\widetilde{f_o^{+}}$, $\widetilde{f_o^-}$ as the enhanced attribute representation to the downstream few-shot classification, where $f_o$ is the original feature from ResNet backbone and $\widetilde{f_o^{+}}$, $\widetilde{f_o^-}$ are the average CoN and DecoN transformed features over all attributes respectively. 

In the implementation, we build SymNet on the pre-trained models from COMP~\cite{comp} as backbones. 
We run in the two settings from COMP~\cite{comp}, about whether to use data augmentation in training classifier. They are marked as {\bf COMP} and {\bf COMP w/ data aug} respectively in tables in the main text. The default augmentation (indicated with \textit{w/ + comp}) in COMP~\cite{comp} includes \textit{Random Resized Crop}, \textit{Image Jitter}, \textit{Random Horizontal Flip}.
We use the base classes in training and set $\lambda_1=5e-2,\lambda_2=1e-3,\lambda_3=1,\lambda_4=10,\lambda_5=1,\lambda_6=5e-2,\lambda_7=1$ and learning rate $1e-3$ on both two datasets. We train SymNet for 1,000 epochs with batch size 128 on CUB-200-2011~\cite{cub}, but 500 epochs with batch size 256 on SUN397~\cite{sun} since it is much larger. 
After that, based on the concatenated features ($f_o$, $\widetilde{f_o^{+}}$, $\widetilde{f_o^-}$) from SymNet, we train a cosine classifier with learning rate 1e-1 and batch size 1,000 in 1,2,5-shot settings. It runs for 100 iterations on CUB-200-2011~\cite{cub} and 200 iterations on SUN397~\cite{sun}.

The results on CUB-200-2011~\cite{cub} and SUN397~\cite{sun} are listed in Tab.~6 (main text) and Tab.~\ref{tab:fewshot-sun}.
Enhanced with SymNet, the COMP~\cite{comp} baseline gains stable performance improvements on the novel and all categories. 

\begin{table}[h]
	\centering
	\small
	\adjustbox{width=0.9\linewidth}{
		\begin{tabular}{l|ccc|ccc}
			\toprule
			\multirow{2}{*}{Method} &  \multicolumn{3}{c}{Novel}& \multicolumn{3}{c}{All} \\
            & 1-shot & 2-shot & 5-shot & 1-shot & 2-shot & 5-shot \\
			\midrule
			COMP\cite{comp} & 43.4 & 54.5 & 65.9 & 54.9 & 60.4 & 66.3 \\ 		
			COMP - SymNet & \textbf{43.6} & \textbf{54.9} & \textbf{66.1} & \textbf{55.5} & \textbf{61.3} & \textbf{67.2} \\  
			\bottomrule
	\end{tabular}}
	\caption{\small Supplementary results of few-shot recognition on SUN397\cite{sun}.}
	\label{tab:fewshot-sun}
\end{table}

\section{More Discussion}
\noindent\textbf{Analysis of Single-attribute Dataset}.
Comparatively, accuracy on MIT-States~\cite{mit} is much lower than UT-Zappos~\cite{ut} as MIT-States has many more objects and attribute categories and suffers from noisy samples and data insufficiency.
Besides, the synonyms and near-synonyms in attributes significantly affect the results. For example, SymNet recognizes $20.4\%$ samples with attribute \texttt{ancient} as \texttt{old}, while the visual properties of these two attributes can barely be distinguished. 
These results are correct from the human perspective but mistaken according to the benchmark. 
To explore this phenomenon on MIT-States, we manually select 13 sets of near-synonyms from MIT-States\footnote{\{cracked, shattered, splintered\}; \{chipped, cut\}; \{dirty, grimy\}; \{eroded, weathered\}; \{huge, large\}; \{melted, molten\}; \{ancient, old\}; \{crushed, pureed, mashed\}; \{ripped, torn\}; \{crinkled, crumpled, ruffled, wrinkled\}; \{small, tiny; damp, wet\}
}, which are chosen according to the similarity in both linguistic meanings and visual patterns. We then regard the attributes within each set as equal, i.e., predicting the near-synonym is also considered correct. On this new benchmark, our model achieves a $3.03\%$ improvement on attribute accuracy and a $0.66\%$ improvement on CZSL accuracy. We also apply this strategy to AttrOperator~\cite{operator}, obtain an improvement of $2.25\%$ on attribute recognition and $0.28\%$ on CZSL recognition. Comparing to AttrOperator, our model suffers more from the synonym problem.

\noindent\textbf{Analysis of Multi-attribute Dataset}.
As shown in Fig.~5 of the main text, aPY~\cite{apy} have stronger overall attribute correlations than SUN~\cite{sun}. Thus, the ablation study verifies that the contribution of the correlation information in learning is more significant on aPY than SUN.
Since images in aPY may have multiple objects, conventional methods that directly crop and resize each instance suffer from noisy boxes and data insufficiency. Statistically, each instance in aPY occupies 25.1\% of the image. And instances in the same image are usually of the same category. Therefore, learning from the context may contribute to a better object representation. For aPY, our model with feature extracted by ROI-pooling~\cite{faster} performs much better than the direct cropping.

\end{document}